\pdfoutput=1
\documentclass{article} % For LaTeX2e
\usepackage{preprint,times}
\usepackage[T1]{fontenc}
\usepackage{graphicx}  
\usepackage{fvextra}
\usepackage{listings}
\usepackage{tcolorbox}
\tcbuselibrary{skins,breakable,listings}

\usepackage{amsmath,amsfonts,bm}

\def\eqref#1{equation~\ref{#1}}
\def\1{\bm{1}}

\DeclareMathAlphabet{\mathsfit}{\encodingdefault}{\sfdefault}{m}{sl}
\SetMathAlphabet{\mathsfit}{bold}{\encodingdefault}{\sfdefault}{bx}{n}

\usepackage{algorithm}
\usepackage{algpseudocode}
\usepackage{amssymb}
\usepackage{booktabs}
\usepackage{multirow}
\usepackage{hyperref}
\usepackage{url}
\newcommand{\makecell}[1]{\begin{tabular}{@{}c@{}}#1\end{tabular}}

\title{VASP: Visual-anchored Agent for \\ Scientific Poster Generation}

\author{%
\makebox[\dimexpr\textwidth-2\tabcolsep\relax][l]{\hspace{-\tabcolsep}\begin{minipage}[t]{\textwidth}
{\fontsize{9.5}{11.5}\selectfont\textbf{Huihan~Wang\textsuperscript{1,2,*}, Jiatian~Chen\textsuperscript{1,3,*}, Yiming~Li\textsuperscript{1}, Siyuan~Zhang\textsuperscript{1}, Yanhui~Hong\textsuperscript{1,\textdagger}, Nan~Wang\textsuperscript{1,\textdagger}}}\\[6pt]
\normalfont\small\raggedright
\textsuperscript{1}\,DP Technology, Beijing, China \\
\textsuperscript{2}\,Department of Biomedical Engineering and Sports Technology (BEST),\\
\phantom{\textsuperscript{2}\,}The Hong Kong Polytechnic University, Hong Kong, China \\
\textsuperscript{3}\,School of Biological Science and Medical Engineering, Beihang University, Beijing, China \\[5pt]
\textsuperscript{*}Equal contribution. \quad \textsuperscript{\textdagger}Corresponding authors. \\[2pt]
\texttt{wang2003huihan@gmail.com}\textsuperscript{*}, \quad \texttt{\{hongyanhui, wangnan01\}@dp.tech}\textsuperscript{\textdagger}
\end{minipage}}%
}

\begin{document}
\raggedbottom

\maketitle
\lhead{Preprint.}

\begin{abstract}
Scientific poster generation condenses a paper into a single page, where visual evidence must remain legible and clearly support the paper's central claims. Existing approaches exhibit two mismatches: a layout mismatch, in which text arrangements or templates constrain visual evidence regardless of its scientific importance, and an execution mismatch, in which models estimate geometry that programs can compute directly. Consequently, key evidence can be compressed or incompletely extracted, while refinement focused on space utilization leaves its association with the claims it supports unchecked. We address both mismatches with VASP, a visual-first framework that combines semantic reasoning with precise execution. \emph{Vision-Centric Anchoring} makes key visual evidence the basis of layout construction, with its scientific role guiding spatial allocation and the arrangement of related claims. \emph{Semantic-Geometric Decoupling} assigns scientific interpretation and design decisions to LLMs, while programs compute figure boundaries and check spatial constraints derived from explicitly declared claim–evidence relations. This division supports intact evidence extraction and iterative refinement guided by the intended scientific organization. We further introduce two criterion-level benchmarks that assess visual presentation and author-informed key-information coverage. On the 100-paper Paper2Poster evaluation set, VASP achieves the highest visual-presentation score (94.3 vs.\ 88.1 for the strongest existing approach) and Paper2Poster VLM-judge score (3.71 vs.\ 3.52). It also achieves 94.3 mAP in figure extraction, compared with 84.1 for ResearchStudio’s agent loop, with complete crops for all 100 sampled figures, and preserves more author-informed key information than existing fixed-canvas approaches. Code is available at \url{https://github.com/WTHH031230/VASP}.
\end{abstract}

\section{Introduction}
\label{sec:introduction}
\suppressfloats[t] % keep floats off the title page top

\begin{figure}[t]
\centering
\includegraphics[width=\textwidth]{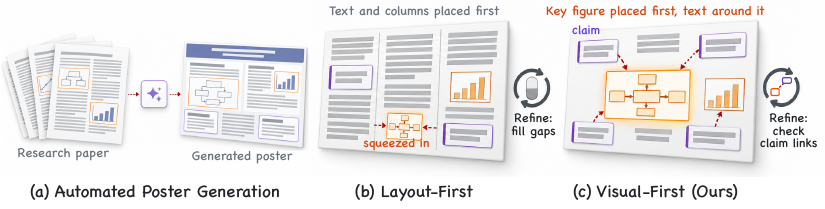}
\caption{Layout-first versus visual-first poster generation (schematic). (a)~Given a paper, the poster must present its central claims and the figures that support them. (b)~Existing systems often fix text arrangements or column structures first, so a key figure receives only the space that remains, and refinement acts on how well that space is filled. (c)~VASP uses key visual evidence (here, a figure) to guide the layout and the placement of related claims. Iterative refinement then checks and adjusts their spatial arrangement to reflect the intended claim–evidence relations.}
\label{fig:visual-first}
\end{figure}

\begin{figure*}[t]
\centering
\includegraphics[width=\textwidth]{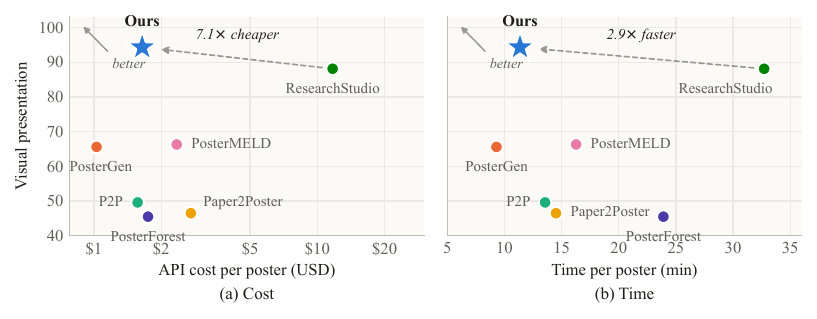}
\caption{Visual-presentation quality versus API cost and time per poster (Appendix~\ref{app:cost-time}). VASP (GPT-5.6-Sol) obtains the highest score while being 7.1$\times$ cheaper and 2.9$\times$ faster than ResearchStudio, the strongest existing system. The Paper2Poster VLM-as-Judge also ranks VASP first.}
\label{fig:teaser}
\end{figure*}

Automated scientific poster generation condenses a multimodal paper into a single page that communicates its central scientific claims and supporting evidence \citep{jaisankar2024postdoc,pang2025paper2poster,inadumi2025scipostgen}.  Figures, tables, equations, and algorithms provide visual evidence that supports or explains these claims. Effective poster generation therefore requires selecting important content and making these claim--evidence relations clear through visual composition \citep{erren2007poster}. This motivates our central premise: \textbf{key visual evidence should anchor the layout}, with its scientific importance and presentation needs guiding the allocation of space and the arrangement of related text (Figure~\ref{fig:visual-first}).

Despite advances in content--layout coordination, existing approaches face two mismatches. The first is a \textbf{layout mismatch} between the scientific importance of visual evidence and the way space is allocated. Many approaches establish text arrangements \citep{yu2026paperx,xiao2026researchstudio} or template structures \citep{jaisankar2024postdoc,sun2025p2p,zhang2025postergen,hu2026postermeld} before placing visual assets. A key figure consequently receives the remaining space or a predefined slot, which can reduce the legibility and prominence of the evidence supporting a central claim. Subsequent refinement typically addresses occupancy, overflow, and the overall appearance of the rendering \citep{pang2025paper2poster,sun2025p2p,choi-etal-2026-posterforest,hu2026postermeld,xiao2026researchstudio}. These checks do not explicitly test whether the rendered layout reflects the claim--evidence relations selected during content planning. A poster can thus use its canvas efficiently while leaving a central claim visually separated from its supporting evidence.

The second is an \textbf{execution mismatch} between decisions requiring semantic interpretation and spatial operations that admit direct computation. Reliable visual composition first requires intact evidence. However, detector-based parsing and agent-based cropping from page images can truncate figure content or include neighboring material \citep{pang2025paper2poster,sun2025p2p,xiao2026researchstudio}. Such extraction combines two distinct tasks: determining which content belongs to a figure and locating its boundaries. The former requires interpretation, whereas the latter can use the geometry already available in the PDF. A similar distinction applies to poster refinement. Deciding which evidence supports a claim requires scientific understanding, but measuring the distance between them or checking their displayed sizes can be performed directly on the rendered layout. Separating these responsibilities allows semantic decisions to guide execution while making spatial errors measurable and actionable.

Evaluation provides limited visibility into these problems. Holistic ratings depend on the judge's calibration and do not localize concrete visual failures such as compressed figures, illegible labels, or detached captions \citep{chen2024mllmjudge}, while question-answering and similarity metrics sample details across the full paper rather than testing specific visual presentation requirements or the selective takeaways authors emphasize \citep{pang2025paper2poster,vinaykumar2026any2poster}. We therefore investigate visual-first poster generation through both a generation framework and criterion-level evaluation benchmarks.

Our framework, VASP (Visual-anchored Agent for Scientific Poster generation), addresses the two mismatches through two complementary design principles. \textbf{Vision-Centric Anchoring} makes key visual evidence the starting point for composition: its scientific role guides layout construction, evidence size, and the placement of related claims (Figure~\ref{fig:visual-first}(c)). \textbf{Semantic-Geometric Decoupling} assigns scientific interpretation and design decisions to LLMs, while programs perform geometric operations during evidence extraction and poster checking \citep{gao2023pal,chen2023pot}. Together, these principles connect semantic reasoning to precise execution. The LLM determines what the poster should communicate, and programmatic measurements guide iterative revisions toward a rendering that satisfies the spatial requirements of that organization.

We instantiate these principles in two stages. \textbf{Semantics-Directed Asset Parsing (SemParser)} first uses a paper-level, text-only LLM inference step to assign semantic roles to text lines. A program then uses these roles and the PDF geometry to compute figure boundaries. This separation retains figure-associated content while excluding neighboring text, without repeated image-based cropping and verification calls. It provides the intact visual assets needed for evidence-centered composition while reducing extraction overhead.

\textbf{Vision-Centric Layout Synthesis (VisComposer)} then organizes the poster around the selected evidence and the claims it supports. An LLM-based design agent jointly determines evidence size, text content, and spatial arrangement on a template-free canvas. It also explicitly records the claim--evidence relations and narrative order in a structured design state. This declaration links scientific planning to layout verification. Specifically, a programmatic harness checks claim--evidence proximity, reading order, and minimum evidence size against the design state. The agent receives measured violations and rendered views, then iteratively revises the poster \citep{yao2023react,shinn2023reflexion,yang2024sweagent}. Refinement therefore checks and corrects how the intended scientific organization is expressed in the layout, alongside conventional checks for overflow and legibility.

We complement the Paper2Poster evaluation \citep{pang2025paper2poster} with two criterion-level benchmarks for visual presentation and author-informed key-information coverage. On the 100-paper evaluation set, VASP achieves the highest visual-presentation score (94.3 vs.\ 88.1) and VLM-as-Judge score (3.71 vs.\ 3.52), while using approximately one-seventh of ResearchStudio's API cost and achieving a 2.9$\times$ speedup (Figure~\ref{fig:teaser}). SemParser reaches 94.3 mAP in figure extraction, with complete crops for all 100 manually checked samples. VASP also preserves more author-informed key information than existing fixed-canvas approaches.

Our main contributions are:

\begin{itemize}
\item We formulate visual-first scientific poster generation around two principles: Vision-Centric Anchoring aligns spatial allocation with the scientific role of visual evidence, and Semantic-Geometric Decoupling connects LLM-based interpretation with programmatic execution.
\item We introduce VASP, a two-stage framework comprising SemParser for intact figure extraction and VisComposer for evidence-centered composition. Explicit claim--evidence relations guide iterative refinement through measurable spatial constraints.
\item We introduce two criterion-level benchmarks for visual presentation and author-informed key-information coverage, enabling assessment of how clearly posters present key evidence, organize related content, and preserve the information authors emphasize.
\end{itemize}

\section{Related Work}
\label{sec:related-work}

\paragraph{Scientific Asset Extraction and Understanding.}
Scientific visual assets are usually extracted from PDFs in one of two ways: by analyzing page structure, as PDFFigures 2.0 does when it detects captions and body text and locates figures and tables in the empty regions they leave \citep{clark2016pdffigures}, or by learning to detect figures and other layout regions \citep{zhong2019publaynet,li2020docbank,huang2022layoutlmv3}, as in DeepFigures and DocLayout-YOLO \citep{siegel2018extracting,zhao2024doclayout}. Several poster generation methods obtain their figures from such a detector or from a document parser such as Docling, MinerU, or Marker \citep{sun2025p2p,pang2025paper2poster,zhang2025postergen,choi-etal-2026-posterforest,hu2026postermeld,livathinos2025docling,wang2024mineru,paruchuri2025marker}, while ResearchStudio cleans up candidate crops deterministically, lets a vision-language model decide a tight crop, and has a separate agent verify it \citep{xiao2026researchstudio}. Our SemParser (Section~\ref{sec:asset-extraction}) instead decides membership from text roles and extent from PDF geometry.

Regardless of how figures are located, extraction quality is commonly measured by overlap with annotated boxes: PDFFigures 2.0 and DeepFigures count a predicted box as correct above an IoU of 0.8 \citep{clark2016pdffigures,siegel2018extracting}, and OmniDocBench scores layout detection, including image regions, by COCO-style mAP \citep{ouyang2025omnidocbench}; we use its evaluator for our figure crops. Overlap can nevertheless remain high when a small legend, axis label, or subfigure is cut off, so we also check by hand whether sampled crops preserve the complete figure (Section~\ref{sec:evaluating-assets}).

\paragraph{Scientific Poster Generation.}
Paper-to-poster generation links multimodal content selection to spatial composition \citep{qiang2016learning,qiang2019learning,xu2021neural,xu2022posterbot,tanaka2024scipostlayout,zhong2025scientific}. PostDoc combines text--image summarization with content-conditioned templates \citep{jaisankar2024postdoc}. Agentic workflows in Paper2Poster and P2P integrate paper understanding, planning, rendering, and iterative revision \citep{pang2025paper2poster,sun2025p2p}. More explicit coordination between content and layout takes several forms \citep{hsu2023posterlayout,lin2023layoutprompter,seol2024posterllama,hsu2025postero}. PosterGen couples narrative planning and visual anchors with programmatic layout and model-based balancing within a three-column grid \citep{zhang2025postergen}. PosterForest uses a Poster Tree to support hierarchical reasoning and joint refinement of content and layout \citep{choi-etal-2026-posterforest}. Layout guidance also comes from retrieved examples in SciPostGen \citep{inadumi2025scipostgen}, text-first refinement in PaperX \citep{yu2026paperx}, and template capacities in PosterMELD \citep{hu2026postermeld}. In several existing approaches, the space a figure can occupy is bounded by a text arrangement or template slot that is decided before the figure is placed \citep{sun2025p2p,zhang2025postergen,yu2026paperx,hu2026postermeld,xiao2026researchstudio}. Following Vision-Centric Anchoring, our VisComposer instead sizes key visual evidence by its role in the claims and organizes the rest of the poster around it on a template-free canvas (Section~\ref{sec:poster-design}).

\paragraph{Agentic Design and Refinement.}
Code-based representations and tool interfaces support agentic visual design and editing \citep{si2025design2code,cheng2024graphicdesign,ge2025autopresent}. DesignAsCode uses HTML/CSS within a plan--implement--reflect workflow \citep{liu2026designascode}, while APEX provides multi-level APIs and review-guided revision for editing existing academic posters \citep{shi2026apex}. In scientific poster production, ResearchStudio uses measured layout feedback to guide content filling \citep{xiao2026researchstudio}, and AutoDesign optimizes the harness itself using generation trajectories and evaluation feedback \citep{luo2026autodesign}. Existing refinement loops for posters act on feedback about occupancy, overflow, and geometric validity, sometimes complemented by a vision-language model's judgment of the rendered poster \citep{pang2025paper2poster,sun2025p2p,choi-etal-2026-posterforest,hu2026postermeld,xiao2026researchstudio}. Following Semantic-Geometric Decoupling, our VisComposer leaves interpretation and composition decisions to the LLM-based design agent and assigns rendering and measurement to programs, iteratively refining the poster to realize its planned semantic organization.

\paragraph{Poster Evaluation.}
Evaluation examines both information preservation and visual communication. Paper2Poster introduces PaperQuiz alongside assessments of poster content and design \citep{pang2025paper2poster}; Any2Poster extends factual and interpretive quizzes and visual judging to multiple input modalities and domains \citep{vinaykumar2026any2poster}. AutoDesign's PosterBench evaluates scientific and visual quality with criteria including faithfulness, coverage, and visual evidence \citep{luo2026autodesign}. Broad numerical ratings \citep{pang2025paper2poster,zhang2025postergen,luo2026autodesign} depend on how the VLM judge calibrates its scoring scale \citep{chen2024mllmjudge,lee2024prometheusvision}, and quiz or similarity scores \citep{pang2025paper2poster,vinaykumar2026any2poster,lin2004rouge,zhang2020bertscore} do not directly test whether a poster meets specific visual communication requirements or preserves the information that the authors emphasize. To address these limitations, we complement the established Paper2Poster evaluation with two evidence-based protocols, key information coverage and visual presentation quality (Section~\ref{sec:evaluating-posters}).

\section{Method}
\label{sec:method}

\begin{figure}[t]
  \centering
  \includegraphics[width=\textwidth]{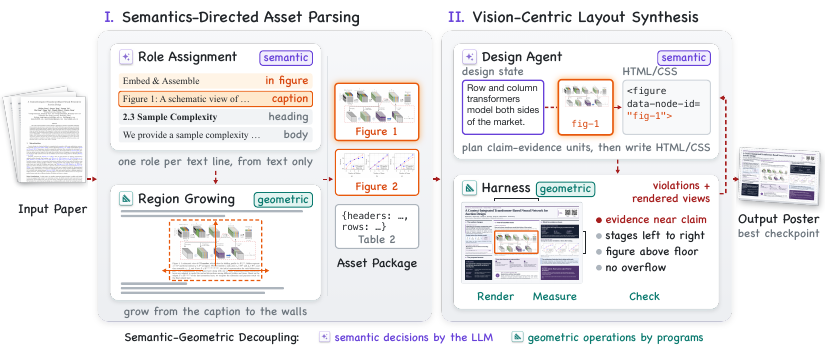}
  \caption{Overview. I.~Semantics-Directed Asset Parsing (SemParser, Section~\ref{sec:asset-extraction}): a single text-only LLM inference
  step assigns a role to every numbered text line, and the program grows each figure region from its
  caption until it meets a wall, a line that belongs elsewhere. II.~Vision-Centric Layout Synthesis
  (VisComposer, Section~\ref{sec:poster-design}): the design agent declares a design state and writes HTML/CSS that
  carries the same identifiers. The harness renders the source, checks whether the declared
  organization is realized, and returns rendered views and measured violations to the same session.
  Tags mark the semantic decisions made by the LLM and the geometric operations performed by programs
  (Semantic-Geometric Decoupling).}
  \label{fig:method}
\end{figure}

We instantiate the two principles of Section~\ref{sec:introduction} in a two-stage method (Figure~\ref{fig:method}). After fixing notation (Section~\ref{sec:setup}), we present SemParser, which applies Semantic-Geometric Decoupling to figure extraction (Section~\ref{sec:asset-extraction}), and VisComposer, which applies Vision-Centric Anchoring to composition and Semantic-Geometric Decoupling to refinement (Section~\ref{sec:poster-design}).

\subsection{Problem Setup}
\label{sec:setup}

Given a paper $\mathcal{D}$ and a canvas $\Omega=[0,W]\times[0,H]$, we generate poster source code $C$ whose rendering $\Gamma(C)$ presents the central claims of $\mathcal{D}$ together with the evidence that supports them. SemParser maps $\mathcal{D}$ to an asset package $\mathcal{A}$, and VisComposer maps $\mathcal{A}$ to $C$. The paper consists of text lines $\{\ell_i\}_{i=1}^{N}$ with texts $s_i$, and on the poster every element $v$ occupies a rendered box $B_v\subset\Omega$.

\subsection{Semantics-Directed Asset Parsing (SemParser)}
\label{sec:asset-extraction}

Following Semantic-Geometric Decoupling, SemParser decides which content belongs to each figure with an LLM that reads the text, and computes each figure's extent from the geometry of the PDF with a program. Existing pipelines localize figures on page images (Section~\ref{sec:related-work}), and their crops drop text outside the graphics (e.g., subfigure captions), absorb neighboring content, split multi-panel figures, or miss figures. These errors concern which content belongs to which figure, which the paper's captions already settle, and LLMs read such roles from text more reliably than VLMs judge geometry from images~\citep{gilardi2023chatgpt,wang2024picture,rahmanzadehgervi2024blind,guan2024hallusionbench}.

The first decision, membership, is made in a single text-only LLM call (Equation~\ref{eq:role-assignment}),
\begin{equation}
  \big(r,\ \mathcal{X}\big)=f\big(\{(i,s_i)\}_{i=1}^{N}\big),\qquad r:\{1,\dots,N\}\rightarrow\mathcal{R},
  \label{eq:role-assignment}
\end{equation}
where $r$ assigns every line a role from a closed vocabulary $\mathcal{R}$ (e.g., figure caption, subfigure caption, body text) and $\mathcal{X}$ are transcriptions of tables, equations, and algorithms (Appendix~\ref{app:asset-parsing}). A program then computes extent from the PDF geometry: each caption becomes an anchor, lines that belong elsewhere become walls, and each figure region grows from its anchor until it meets a wall. Figures are thus enumerated from their captions and bounded by text roles rather than detected in images.

The cropped figures, the transcriptions $\mathcal{X}$, and the numbered text form the asset package $\mathcal{A}$, which contains no narrative or spatial decision and can therefore serve as a shared asset bundle for any downstream design, much as ResearchStudio feeds one bundle to a poster, a video deck, and a blog~\citep{xiao2026researchstudio}.

\subsection{Vision-Centric Layout Synthesis (VisComposer)}
\label{sec:poster-design}

VisComposer lets the point each figure makes decide its size, with legibility as a condition of admission (Vision-Centric Anchoring), and has the agent declare the intended organization so that the harness can verify it on the rendering (Semantic-Geometric Decoupling). Both depart from existing approaches, which size figures by the space that text or templates leave and refine on occupancy, overflow, geometric validity, or a vision-language model's judgment of the rendering (Sections~\ref{sec:introduction} and~\ref{sec:related-work}). Such feedback can be written as $\rho(\Gamma(C))$: the planned relations, such as which evidence supports which claim, are not among its arguments, so no signal checks them against the rendering.

Composition carries out the first principle: the evidence, not the space that remains, determines how much room each figure receives. The agent composes the poster from claim--evidence units on a template-free HTML/CSS canvas (Appendix~\ref{app:layout-synthesis}). Let $\mathcal{C}$ be the candidate units, $z_v\in\{0,1\}$ indicate whether $v$ is placed, and $(\tilde{w}_v,\tilde{h}_v)$ be the size at which a figure $v$ is displayed. Composition is cast as the problem in Equation~\ref{eq:composition}, which the agent solves approximately rather than by a numerical solver:
\begin{equation}
  \max_{\mathbf{z},\,C}\ \sum_{v\in\mathcal{C}} z_v\,u(v)
  \quad\text{s.t.}\quad
  z_v=1\Rightarrow\tilde{w}_v\ge\alpha W\wedge\tilde{h}_v\ge\beta H\ \ \forall v\in\mathcal{C}_{\mathrm{fig}},
  \qquad \Gamma(C)\in\mathcal{F},
  \label{eq:composition}
\end{equation}
where $u(v)$ is the contribution of $v$ to communicating the paper, $\mathcal{C}_{\mathrm{fig}}$ are the figures, $\alpha$ and $\beta$ are fixed fractions of the canvas, and $\mathcal{F}$ contains the renderings that fit $\Omega$ with legible type sizes. Space conflicts are thus resolved by selection rather than by scaling, and the decoupling applies here as well: $u$ is judged by the agent, whereas the constraints are measured and enforced by the harness.

Verification carries out the second principle: it checks whether the rendering realizes the relations that the agent planned between claims and evidence. Alongside $C$~\citep{liu2026designascode}, the agent declares a design state $S$, a claim--evidence graph over the placed units with narrative stages and a primary anchor, whose identifiers are carried by the rendered elements. The harness evaluates a set $\Phi$ of predicates on $S$ and the measured boxes $B_v$ (Appendix~\ref{app:layout-synthesis}), and the feedback at checkpoint $t$ is the set of violated predicates,
\begin{equation}
  F_t=\big\{\phi\in\Phi:\ \phi\big(S_t,\{B_v\}_t\big)=0\big\},\qquad g_t=\mathbb{I}\big[F_t=\emptyset\big].
  \label{eq:feedback}
\end{equation}
Unlike $\rho(\Gamma(C))$, $F_t$ in Equation~\ref{eq:feedback} takes the declaration as an argument, so the loop can detect a planned relation that the rendering breaks, such as evidence placed far from the claim it supports. The predicates are necessary conditions, and emphasis, hierarchy, and aesthetics remain with the agent.

The refinement loop runs within one persistent session. At each checkpoint the agent receives $F_t$ and rendered views $I_t$ and answers with local edits $\Delta_t$ and a re-declared state, $(\Delta_t,S_{t+1})\sim\pi(\cdot\mid\mathcal{A},\mathcal{H}_t,I_t,F_t)$, where $\mathcal{H}_t$ is the session history and $C_{t+1}=\operatorname{Apply}(C_t,\Delta_t)$. The loop stops once all predicates hold or after $K$ revisions and returns the best stored checkpoint (Algorithm~\ref{alg:method} in Appendix~\ref{app:layout-synthesis}).

\section{Evaluation Framework}
\label{sec:evaluation}
\label{sec:evaluating-posters}

To complement holistic ratings and quiz-based measures (Section~\ref{sec:related-work}), we introduce two benchmarks, one for visual presentation and one for key-information coverage. Both decompose evaluation into binary judgments with explicit evidence conditions, so that each score is the proportion of requirements satisfied or reference units covered, and every decision can be checked against the poster. We report the two dimensions separately.

\textbf{Visual presentation.} Twelve criteria in three groups test whether figures are identifiable and appropriately sized (figure presentation, 3 criteria), whether hierarchy and emphasis guide reading (visual guidance and emphasis, 5), and whether spatial association, information density, overlap, and whitespace support communication (layout and spatial organization, 4). Given only the rendered poster, the evaluator judges each criterion, and only \textsc{Pass} receives credit. The overall score weights the three groups equally and the criteria within each group equally, so that no group gains weight merely by containing more criteria (Appendix~\ref{app:visual-benchmark}).

\textbf{Author-informed key-information coverage.} For each paper, a VLM first reads only the author-designed poster to identify seven distinct and independently judgeable key-information units, covering aspects such as the research problem, method, theoretical contribution, empirical findings, or limitations. These units are then audited against the source paper, revised where necessary, and frozen before any candidate poster is evaluated. A unit is \textsc{Covered} when the candidate visibly communicates its required meaning, whether in text, equations, figures, captions, or legible tables, and coverage is the fraction of units covered. Because coverage also depends on how much content a poster's canvas can hold, we rank only methods with a fixed canvas (Appendix~\ref{app:coverage-benchmark}).

\section{Experiments}
\label{sec:experiments}

\subsection{Evaluation Setup}
\label{sec:evaluation-setup}

\textbf{Evaluation design.} We use the 100-paper evaluation set released with Paper2Poster \citep{pang2025paper2poster}. For each paper, the dataset provides the source paper and an official author-designed poster. We compare the official posters, six existing paper-to-poster methods (P2P, Paper2Poster, PosterForest, PosterGen, PosterMELD, and ResearchStudio), and two instantiations of VASP driven by GPT-5.6-Sol and Opus-5 (Appendix~\ref{app:backbone-comparison} compares a third backbone). All baselines are run with GPT-5.6-Sol under their default settings. All methods are evaluated on the same paper set, yielding nine sets of outputs and 900 posters in total. All posters are evaluated from their final PNG renderings while preserving their original dimensions and aspect ratios.

\textbf{Implementation.} VASP designs every poster on a $2544\times1696$\,px canvas with a revision budget of $K=2$ (Appendices~\ref{app:implementation} and~\ref{app:effort-rounds}).

\subsection{Evaluation on Paper2Poster Benchmark}
\label{sec:existing-benchmark-results}

We first evaluate our generated posters using the VLM-as-Judge and PaperQuiz components of the Paper2Poster benchmark \citep{pang2025paper2poster}, which provide an established reference point for comparison. For VLM-as-Judge, each poster is scored by GPT-5.6-Sol using the benchmark's original judging prompts. For PaperQuiz, GPT-5.6-Sol answers the benchmark's released multiple-choice questions from the poster image alone using the original answering prompt, and we report raw accuracy.

\begin{table*}[t]
    \centering
    \caption{Results on the VLM-as-Judge and PaperQuiz components of the Paper2Poster benchmark. For VLM-as-Judge, Aesthetic and Information are the means of their three sub-scores, and Overall is the mean of all six. For PaperQuiz, Verbatim and Interpretive are the accuracies (\%) on the 50 verbatim and 50 interpretive questions per paper, and Overall is their mean. Higher is better. $^\dagger$P2P renders a fixed-width page of unbounded height and is therefore not ranked on PaperQuiz.}
    \label{tab:paper2poster-vlm-efficiency}
    \small
    \setlength{\tabcolsep}{5pt}
    \begin{tabular}{lcccccc}
        \hline
        & \multicolumn{3}{c}{VLM-as-Judge} & \multicolumn{3}{c}{PaperQuiz} \\
        \cmidrule(lr){2-4} \cmidrule(lr){5-7}
        Method & Aesthetic$\uparrow$ & Information$\uparrow$ & Overall$\uparrow$ & Verbatim$\uparrow$ & Interpretive$\uparrow$ & Overall$\uparrow$ \\
        \hline
        \multicolumn{7}{l}{\textit{Ours}} \\
        \textbf{VASP (Opus-5)}      & \underline{3.58} & \textbf{3.83} & \underline{3.70} & \textbf{66.94} & \textbf{89.22} & \textbf{78.08} \\
        \textbf{VASP (GPT-5.6-Sol)} & \textbf{3.72} & \underline{3.69} & \textbf{3.71} & 52.62 & \underline{87.30} & \underline{69.96} \\
        \hline
        \multicolumn{7}{l}{\textit{Baseline methods}} \\
        Paper2Poster       & 2.24 & 3.28 & 2.76 & 52.32 & 86.18 & 69.25 \\
        PosterMELD         & 2.55 & 3.11 & 2.83 & 41.90 & 82.82 & 62.36 \\
        PosterForest       & 2.49 & 3.20 & 2.85 & 50.52 & 87.20 & 68.86 \\
        PosterGen          & 2.80 & 3.48 & 3.14 & \underline{52.64} & 82.80 & 67.72 \\
        P2P$^\dagger$      & 2.74 & 3.66 & 3.20 & 75.72 & 94.68 & 85.20 \\
        ResearchStudio     & 3.44 & 3.59 & 3.52 & 49.68 & 85.28 & 67.48 \\
        \hline
        \multicolumn{7}{l}{\textit{Human reference}} \\
        Official posters   & 3.05 & 3.20 & 3.12 & 50.10 & 85.00 & 67.55 \\
        \hline
    \end{tabular}
\end{table*}

\textbf{VLM-as-Judge quality.} As shown in Table~\ref{tab:paper2poster-vlm-efficiency}, VASP (GPT-5.6-Sol) and VASP (Opus-5) obtain the two highest overall scores, 3.71 and 3.70, exceeding the strongest baseline, ResearchStudio (3.52), by 0.19 and 0.18 points. The largest margin is in aesthetics (3.72 vs.\ 3.44). On PaperQuiz, P2P obtains the highest accuracy (85.20\%), but it renders a page of unbounded height and, as for key-information coverage (Section~\ref{sec:evaluation}), is not ranked against fixed-canvas methods. Among these, VASP (Opus-5) is highest (78.08\%) and VASP (GPT-5.6-Sol) second (69.96\%), closely followed by Paper2Poster (69.25\%).

\subsection{Figure Extraction}
\label{sec:asset-figure-quality}
\label{sec:evaluating-assets}

\begin{table*}[t]
    \centering
    \caption{Crop-accuracy and figure-completeness results. Automatic crop-accuracy metrics are percentages computed on the 462 main-text figures from our 948-figure human-annotated benchmark (462 main-text and 486 appendix figures across the 100 papers). The human column reports the number of crops judged complete on a shared sample of 100 figures. Higher is better.}
    \label{tab:asset-auto-results}
    \small
    \setlength{\tabcolsep}{7pt}
    \begin{tabular}{lccccc}
        \hline
        & \multicolumn{4}{c}{Automatic evaluation} & \multicolumn{1}{c}{Human completeness} \\
        \cline{2-5}\cline{6-6}
        Method & mAP$\uparrow$ & AP$_{50}$$\uparrow$ & AP$_{75}$$\uparrow$ & mIoU$\uparrow$ & Complete / 100$\uparrow$ \\
        \hline
        \multicolumn{6}{l}{\textit{Ours}} \\
        \textbf{VASP (Opus-5)}               & \underline{94.06} & \underline{99.97} & \underline{98.81} & \underline{95.59} & \textbf{100} \\
        \textbf{VASP (GPT-5.6-Sol)} & \textbf{94.30} & \textbf{100.00} & \textbf{100.00} & \textbf{95.70} & \textbf{100} \\
        \hline
        \multicolumn{6}{l}{\textit{Baseline methods}} \\
        Paper2Poster (Docling)        & 68.24 & 75.08 & 70.56 & 83.26 & 73 \\
        PosterMELD (MinerU)           & 83.34 & 93.65 & 90.58 & 93.46 & 69 \\
        PosterForest (Docling)        & 73.07 & 79.39 & 75.64 & 80.00 & 76 \\
        PosterGen (Marker)            & 67.77 & 92.86 & 87.95 & 83.88 & 82 \\
        P2P (DocLayout-YOLO)            & 91.06 & 94.94 & 92.13 & 94.69 & 93 \\
        ResearchStudio              & 84.07 & 89.33 & 88.08 & 86.70 & 88 \\
        \hline
    \end{tabular}
\end{table*}

To evaluate figure extraction, we manually annotate bounding boxes for all 948 figures across the 100 papers (462 main-text and 486 appendix figures) and release these annotations, evaluating automatic crop accuracy in Table~\ref{tab:asset-auto-results} on the 462 main-text figures. We compare each method's figure crops with them, reporting bounding-box mAP, $\mathrm{AP}_{50}$, and $\mathrm{AP}_{75}$ under the COCO convention with the OmniDocBench evaluator \citep{lin2014microsoft,ouyang2025omnidocbench}, and mIoU for boundary agreement. Because IoU can remain high when a small legend, axis label, or subfigure is clipped, two annotators also check the crops of 100 figures sampled with a fixed seed for completeness. As Table~\ref{tab:asset-auto-results} shows, VASP (GPT-5.6-Sol) achieves the best result on every automatic metric (94.30 mAP and 95.70 mIoU), and all 100 of its sampled crops are complete.

\subsection{Evaluation on Our Benchmarks}
\label{sec:our-benchmark-results}
\label{sec:poster-level-quality}

\begin{figure*}[t]
\centering
\includegraphics[width=\textwidth]{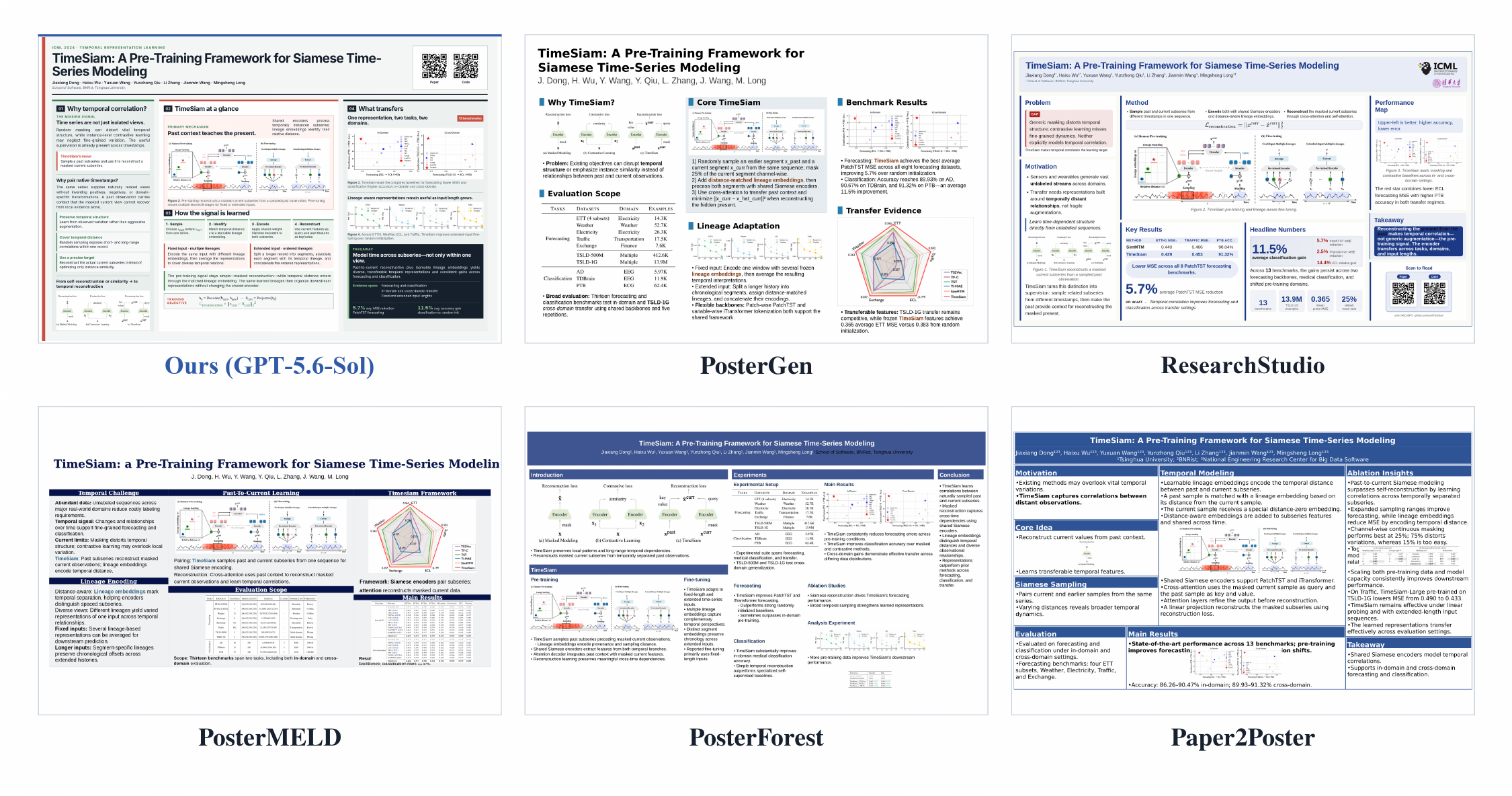}
\caption{Posters generated for the same paper (TimeSiam) by VASP (GPT-5.6-Sol) and five baseline methods. Our poster displays the key figures at a scale at which their content stays legible and places the related claims next to them, whereas in posters generated by the baselines the figures are small relative to the text and fill the space left by text blocks or template slots.}
\label{fig:qualitative}
\end{figure*}

\textbf{Visual presentation.} As shown in Table~\ref{tab:visual-presentation-results}, VASP obtains the two highest overall scores, 94.3 with GPT-5.6-Sol and 92.5 with Opus-5, compared with 88.1 for the strongest baseline, ResearchStudio, and 83.0 for the official posters. The largest gain is in figure presentation, where GPT-5.6-Sol reaches 88.0 against 73.7 for ResearchStudio. Visual guidance and layout improve by up to 3.0 and 4.3 points, respectively. A direct measurement of how large each paper's most important figure is displayed confirms this gain (Appendix~\ref{app:pilot-protocols}), and Figure~\ref{fig:qualitative} shows an example (more examples and failure cases in Appendix~\ref{app:examples}). Figure~\ref{fig:teaser} relates these scores to the API cost and time per poster: VASP (GPT-5.6-Sol) is 2.9$\times$ faster and 7.1$\times$ cheaper than ResearchStudio. Even with the lower-cost GPT-5.6-Luna, VASP still exceeds ResearchStudio on the Paper2Poster VLM-as-Judge (Appendix~\ref{app:backbone-comparison}).

\begin{table*}[t]
    \centering
    \caption{Visual-presentation results on our poster-level benchmark. Higher is better.}
    \label{tab:visual-presentation-results}
    \small
    \setlength{\tabcolsep}{7pt}
    \begin{tabular}{lcccc}
        \hline
        Method & Figure Presentation$\uparrow$ & Visual Guidance$\uparrow$ & Layout$\uparrow$ & Overall$\uparrow$ \\
        \hline
        \multicolumn{5}{l}{\textit{Ours}} \\
        \textbf{VASP (Opus-5)}      & \underline{81.3} & \underline{98.4} & \textbf{97.8} & \underline{92.5} \\
        \textbf{VASP (GPT-5.6-Sol)} & \textbf{88.0} & \textbf{100.0} & \underline{95.0} & \textbf{94.3} \\
        \hline
        \multicolumn{5}{l}{\textit{Baseline methods}} \\
        Paper2Poster       & 35.0 & 61.6 & 43.0 & 46.5 \\
        PosterMELD         & 71.7 & 80.6 & 46.5 & 66.3 \\
        PosterForest       & 38.0 & 56.4 & 42.0 & 45.5 \\
        PosterGen          & 53.0 & 75.4 & 68.3 & 65.6 \\
        P2P                & 11.0 & 76.4 & 61.3 & 49.6 \\
        ResearchStudio    & 73.7 & 97.0 & 93.5 & 88.1 \\
        \hline
        \multicolumn{5}{l}{\textit{Human reference}} \\
        Official posters   & 72.7 & 86.4 & 90.0 & 83.0 \\
        \hline
    \end{tabular}
\end{table*}

\begin{table*}[t]
    \centering
    \caption{Author-informed key-information coverage (\%) on our poster-level benchmark. Higher is better. Among fixed-canvas methods, the best result is in bold and the second best is underlined.}
    \label{tab:key-information-coverage}
    \small
    \setlength{\tabcolsep}{7pt}
    \sbox0{%
    \begin{tabular}{lccccc}
        \hline
        & \multicolumn{3}{c}{Coverage by information type} & & \\
        \cline{2-4}
        Method & Problem$\uparrow$ & Method$\uparrow$ & Results$\uparrow$ & Overall$\uparrow$ & Full$\uparrow$ \\
        \hline
        \multicolumn{6}{l}{\textit{Ours}} \\
        \textbf{VASP (Opus-5)}      & \textbf{94.0} & \textbf{89.0} & \textbf{84.8} & \textbf{88.7} & \textbf{48} \\
        \textbf{VASP (GPT-5.6-Sol)} & \underline{91.3} & \underline{86.9} & 71.7 & \underline{82.0} & 20 \\
        \hline
        \multicolumn{6}{l}{\textit{Baseline methods}} \\
        Paper2Poster       & 72.5 & 73.9 & 73.7 & 74.1 & 15 \\
        PosterMELD         & 81.9 & 73.9 & 48.5 & 66.9 & 7 \\
        PosterForest       & 73.2 & 78.7 & 70.2 & 73.9 & 12 \\
        PosterGen          & 86.6 & 81.4 & \underline{74.7} & 79.7 & \underline{22} \\
        P2P$^\dagger$      & 91.9 & 93.8 & 93.4 & 93.6 & 67 \\
        ResearchStudio     & 79.2 & 75.9 & 65.7 & 73.0 & 18 \\
        \hline
        \multicolumn{6}{l}{\textit{Human reference}} \\
        Official posters   & 99.3 & 97.3 & 94.4 & 96.4 & 80 \\
        \hline
    \end{tabular}}%
    \usebox0

    \vspace{3pt}
    \parbox{\wd0}{\footnotesize Information types are defined in Appendix~\ref{app:coverage-construction}. Full: share of posters that cover all seven units. $^\dagger$P2P renders a page of unbounded height and is not ranked.}
\end{table*}

\textbf{Author-informed key-information coverage.} As shown in Table~\ref{tab:key-information-coverage}, VASP (Opus-5) covers the most key-information units among fixed-canvas methods (88.7\%) and covers all seven units on 48\% of its posters, and VASP (GPT-5.6-Sol) ranks second (82.0\%), ahead of the strongest baseline, PosterGen (79.7\%).

\textbf{Ablations.} On a 20-paper subset (Appendix~\ref{sec:ablation}), removing SemParser, anchored composition, semantic verification, or the refinement loop lowers the visual-presentation score of VASP (GPT-5.6-Sol) from 91.4 to 83.1, 80.8, 81.1, and 73.8, respectively. Decoupled controls on the same subset further isolate planning priority (72.8 when switching from visual-first to text-first on an identical flexible grid) and separate geometric verification, which governs whitespace, from semantic verification, which governs claim--evidence grounding.

\textbf{Additional evaluations.} Appendices~\ref{app:backbone-comparison}--\ref{app:validation-robustness} report extended comparisons under alternative backbones (Claude Opus-5 and Qwen3.8-Flash-Next) and with the multi-format framework PaperX, cross-family judging with Gemini-4-Argon, blind human rankings and criterion-level annotations, visual-theme sensitivity, criterion-subset analyses, \texttt{N/A}-excluded scoring, and paper-paired confidence intervals; the main trends hold across all settings.

\section{Conclusion}
\label{sec:conclusion}

We presented VASP, a visual-first framework that composes scientific posters around their key visual evidence. Vision-Centric Anchoring lets this evidence govern the layout, and Semantic-Geometric Decoupling lets LLMs make interpretive decisions while programs compute and check geometry. Across multiple benchmarks, model backbones, and both automated and human evaluations, VASP presents key evidence more clearly than existing approaches, and ablations support both principles. Because SemParser builds a layout-agnostic asset package once per paper, the same assets and division of labor can extend to other evidence-centered document design tasks, such as slides \citep{sun2021d2s,fu2022doc2ppt,zheng2025pptagent}, short videos, and infographics. We discuss limitations in Appendix~\ref{app:limitations}.

\bibliography{references}

@article{pang2025paper2poster,
  title = {{Paper2Poster: Towards Multimodal Poster Automation from Scientific Papers}},
  author = {Pang, Wei and Lin, Kevin Qinghong and Jian, Xiangru and He, Xi and Torr, Philip},
  journal = {arXiv preprint arXiv:2505.21497},
  year = {2025},
  url = {https://arxiv.org/abs/2505.21497}
}

@article{sun2025p2p,
  title = {{P2P: Automated Paper-to-Poster Generation and Fine-Grained Benchmark}},
  author = {Sun, Tao and Pan, Enhao and Yang, Zhengkai and Sui, Kaixin and Shi, Jiajun and Cheng, Xianfu and Li, Tongliang and Huang, Wenhao and Zhang, Ge and Yang, Jian and Li, Zhoujun},
  journal = {arXiv preprint arXiv:2505.17104},
  year = {2025},
  url = {https://arxiv.org/abs/2505.17104}
}

@article{zhang2025postergen,
  title = {{PosterGen: Aesthetic-Aware Multi-Modal Paper-to-Poster Generation via Multi-Agent LLMs}},
  author = {Zhang, Zhilin and Zhang, Xiang and Wei, Jiaqi and Xu, Yiwei and You, Chenyu},
  journal = {arXiv preprint arXiv:2508.17188},
  year = {2025},
  url = {https://arxiv.org/abs/2508.17188}
}

@article{hu2026postermeld,
  title = {{PosterMELD: Multi-Agent Paper-to-Poster Generation for Controllable Design Diversity with Editable Print-Ready Outputs}},
  author = {Hu, Haojie and Dang, Chenhao and Liu, Yaojia and Kang, Hengrui and He, Conghui and Li, Weijia},
  journal = {arXiv preprint arXiv:2608.02218},
  year = {2026},
  url = {https://arxiv.org/abs/2608.02218}
}

@article{xiao2026researchstudio,
  title = {{ResearchStudio-Reel: Automate the Last Mile of Research from Paper to Poster, Video, and Blog}},
  author = {Xiao, Lingao and Dai, Yalun and Huang, Yangyu and Zhao, Qihao and Wu, Wenshan and He, Hugo and Chen, Ruishuo and Jiang, Jin and Ma, Qianli and Zhang, Jiahuan and Zhang, Xin and Xin, Ying and Ou, Yang and Xia, Yan and Li, Scarlett and Huang, Longbo and Zhang, Zhipeng and He, Yang and Hui, Yap Kim and Lu, Yan},
  journal = {arXiv preprint arXiv:2607.04438},
  year = {2026},
  url = {https://arxiv.org/abs/2607.04438}
}

@article{yu2026paperx,
  title = {{PaperX: A Unified Framework for Multimodal Academic Presentation Generation with Scholar DAG}},
  author = {Yu, Tao and Zhang, Minghui and Cui, Zhiqing and Wang, Hao and Luo, Zhongtian and Chai, Shenghua and Gong, Junhao and Peng, Yuzhao and Zhou, Yuxuan and Yang, Yujia and Zhang, Zhenghao and Jin, Haopeng and Wang, Xinming and Xiong, Yufei and Yang, Jiabing and Yuan, Jiahao and Wang, Hanqing and Yi, Hongzhu and Huang, Yan and Wang, Liang},
  journal = {arXiv preprint arXiv:2602.03866},
  year = {2026},
  url = {https://arxiv.org/abs/2602.03866}
}

@inproceedings{choi-etal-2026-posterforest,
  title = {{PosterForest}: Hierarchical Multi-Agent Collaboration for Scientific Poster Generation},
  author = {Choi, Jiho and Park, Seojeong and Song, Seongjong and Shim, Hyunjung},
  booktitle = {Proceedings of the 64th Annual Meeting of the Association for Computational Linguistics (Volume 1: Long Papers)},
  year = {2026},
  publisher = {Association for Computational Linguistics},
  pages = {379--401},
  doi = {10.18653/v1/2026.acl-long.15},
  url = {https://aclanthology.org/2026.acl-long.15/}
}

@inproceedings{clark2016pdffigures,
  title = {{PDFFigures 2.0: Mining Figures from Research Papers}},
  author = {Clark, Christopher and Divvala, Santosh},
  booktitle = {Proceedings of the ACM/IEEE Joint Conference on Digital Libraries},
  year = {2016},
  url = {https://pdffigures2.allenai.org/}
}

@article{siegel2018extracting,
  title = {{Extracting Scientific Figures with Distantly Supervised Neural Networks}},
  author = {Siegel, Noah and Lourie, Nicholas and Power, Russell and Ammar, Waleed},
  journal = {arXiv preprint arXiv:1804.02445},
  year = {2018},
  url = {https://arxiv.org/abs/1804.02445}
}

@article{zhao2024doclayout,
  title = {{DocLayout-YOLO: Enhancing Document Layout Analysis through Diverse Synthetic Data and Global-to-Local Adaptive Perception}},
  author = {Zhao, Zhiyuan and Kang, Hengrui and Wang, Bin and He, Conghui},
  journal = {arXiv preprint arXiv:2410.12628},
  year = {2024},
  url = {https://arxiv.org/abs/2410.12628}
}

@article{jaisankar2024postdoc,
  title = {{PostDoc: Generating Poster from a Long Multimodal Document Using Deep Submodular Optimization}},
  author = {Vijay Jaisankar and Sambaran Bandyopadhyay and Kalp Vyas and Varre Chaitanya and Shwetha Somasundaram},
  journal = {arXiv preprint arXiv:2405.20213},
  year = {2024},
  url = {https://arxiv.org/abs/2405.20213}
}

@article{inadumi2025scipostgen,
  title = {{SciPostGen: Bridging the Gap between Scientific Papers and Poster Layouts}},
  author = {Inadumi, Shun and Tanaka, Shohei and Hirasawa, Tosho and Hashimoto, Atsushi and Yoshino, Koichiro and Ushiku, Yoshitaka},
  journal = {arXiv preprint arXiv:2511.22490},
  year = {2025},
  url = {https://arxiv.org/abs/2511.22490}
}

@article{liu2026designascode,
  title = {{DesignAsCode: Bridging Structural Editability and Visual Fidelity in Graphic Design Generation}},
  author = {Liu, Ziyuan and Sun, Shizhao and Huang, Danqing and Shi, Yingdong and Zhang, Meisheng and Li, Ji and Yu, Jingsong and Bian, Jiang},
  journal = {arXiv preprint arXiv:2602.17690},
  year = {2026},
  url = {https://arxiv.org/abs/2602.17690}
}

@article{shi2026apex,
  title = {{APEX: Academic Poster Editing Agentic Expert}},
  author = {Shi, Chengxin and Cai, Qinnan and Chen, Zeyuan and Zeng, Long and Zhao, Yibo and Yu, Jing and Yu, Jianxiang and Li, Xiang},
  journal = {arXiv preprint arXiv:2601.04794},
  year = {2026},
  url = {https://arxiv.org/abs/2601.04794}
}

@article{luo2026autodesign,
  title = {{AutoDesign: Meta-Harness Optimization for Long-Horizon Agentic Design}},
  author = {Luo, Yaxin and Jiang, Haobin and Zou, Jialv and Huang, Xu and Yan, Wenhao and Li, Haodong and Yue, Zhengrong and Li, Jing and Chen, Xiaofu and Zhao, Xiaohan and Liu, Jiacheng and Cui, Jiacheng and Shen, Zhiqiang and Li, Xiaotong},
  journal = {arXiv preprint arXiv:2608.13560},
  year = {2026},
  url = {https://arxiv.org/abs/2608.13560}
}

@article{vinaykumar2026any2poster,
  title = {{Any2Poster: Any-Source Poster Generation Across Modalities and Domains}},
  author = {Vinaykumar, Amogh and Li, Aiden and Huang, Suozhi and Liu, Shilong},
  journal = {arXiv preprint arXiv:2606.02915},
  year = {2026},
  url = {https://arxiv.org/abs/2606.02915}
}

@article{erren2007poster,
  title = {Ten Simple Rules for a Good Poster Presentation},
  author = {Erren, Thomas C. and Bourne, Philip E.},
  journal = {PLOS Computational Biology},
  volume = {3},
  number = {5},
  pages = {e102},
  year = {2007},
  doi = {10.1371/journal.pcbi.0030102},
  url = {https://doi.org/10.1371/journal.pcbi.0030102}
}

@inproceedings{chen2024mllmjudge,
  title = {{MLLM}-as-a-Judge: Assessing Multimodal {LLM}-as-a-Judge with Vision-Language Benchmark},
  author = {Chen, Dongping and Chen, Ruoxi and Zhang, Shilin and Wang, Yaochen and Liu, Yinuo and Zhou, Huichi and Zhang, Qihui and Wan, Yao and Zhou, Pan and Sun, Lichao},
  booktitle = {Proceedings of the 41st International Conference on Machine Learning},
  series = {Proceedings of Machine Learning Research},
  volume = {235},
  pages = {6562--6595},
  publisher = {PMLR},
  year = {2024},
  url = {https://proceedings.mlr.press/v235/chen24h.html}
}

@inproceedings{lin2014microsoft,
  title     = {Microsoft {COCO}: Common Objects in Context},
  author    = {Lin, Tsung-Yi and Maire, Michael and Belongie, Serge
               and Hays, James and Perona, Pietro and Ramanan, Deva
               and Doll{\'a}r, Piotr and Zitnick, C. Lawrence},
  booktitle = {European Conference on Computer Vision},
  pages     = {740--755},
  year      = {2014},
  doi       = {10.1007/978-3-319-10602-1_48}
}

@article{gilardi2023chatgpt,
  title   = {{ChatGPT} Outperforms Crowd Workers for Text-Annotation Tasks},
  author  = {Gilardi, Fabrizio and Alizadeh, Meysam and Kubli, Ma{\"e}l},
  journal = {Proceedings of the National Academy of Sciences},
  volume  = {120},
  number  = {30},
  pages   = {e2305016120},
  year    = {2023},
  doi     = {10.1073/pnas.2305016120}
}

@inproceedings{wang2024picture,
  title     = {Is A Picture Worth A Thousand Words? {D}elving Into Spatial Reasoning for Vision Language Models},
  author    = {Wang, Jiayu and Ming, Yifei and Shi, Zhenmei and Vineet, Vibhav and Wang, Xin and Li, Yixuan and Joshi, Neel},
  booktitle = {Advances in Neural Information Processing Systems},
  volume    = {37},
  year      = {2024},
  doi       = {10.52202/079017-2400}
}

@inproceedings{rahmanzadehgervi2024blind, title={Vision Language Models are blind}, ISBN={9789819609178}, ISSN={1611-3349}, url={http://dx.doi.org/10.1007/978-981-96-0917-8_17}, DOI={10.1007/978-981-96-0917-8_17}, booktitle={Computer Vision – ACCV 2024}, publisher={Springer Nature Singapore}, author={Rahmanzadehgervi, Pooyan and Bolton, Logan and Taesiri, Mohammad Reza and Nguyen, Anh Totti}, year={2024}, month=Dec, pages={293–309} }

@inproceedings{ouyang2025omnidocbench,
  title     = {{OmniDocBench}: Benchmarking Diverse {PDF} Document Parsing with Comprehensive Annotations},
  author    = {Ouyang, Linke and Qu, Yuan and Zhou, Hongbin and Zhu, Jiawei and Zhang, Rui and Lin, Qunshu and Wang, Bin and Zhao, Zhiyuan and Jiang, Man and Zhao, Xiaomeng and Shi, Jin and Wu, Fan and Chu, Pei and Liu, Minghao and Li, Zhenxiang and Xu, Chao and Zhang, Bo and Shi, Botian and Tu, Zhongying and He, Conghui},
  booktitle = {Proceedings of the IEEE/CVF Conference on Computer Vision and Pattern Recognition (CVPR)},
  pages     = {24838--24848},
  year      = {2025}
}

@article{chen2023pot,
title={Program of Thoughts Prompting: Disentangling Computation from Reasoning for Numerical Reasoning Tasks},
author={Wenhu Chen and Xueguang Ma and Xinyi Wang and William W. Cohen},
journal={Transactions on Machine Learning Research},
issn={2835-8856},
year={2023},
url={https://openreview.net/forum?id=YfZ4ZPt8zd},
note={}
}

@misc{cheng2024graphicdesign,
      title={Graphic Design with Large Multimodal Model}, 
      author={Yutao Cheng and Zhao Zhang and Maoke Yang and Hui Nie and Chunyuan Li and Xinglong Wu and Jie Shao},
      year={2024},
      eprint={2404.14368},
      howpublished={arXiv preprint arXiv:2404.14368},
      archivePrefix={arXiv},
      primaryClass={cs.CV},
      url={https://arxiv.org/abs/2404.14368}, 
}

@article{fu2022doc2ppt, title={{DOC2PPT}: Automatic Presentation Slides Generation from Scientific Documents}, volume={36}, ISSN={2159-5399}, url={http://dx.doi.org/10.1609/aaai.v36i1.19943}, DOI={10.1609/aaai.v36i1.19943}, number={1}, journal={Proceedings of the AAAI Conference on Artificial Intelligence}, publisher={Association for the Advancement of Artificial Intelligence (AAAI)}, author={Fu, Tsu-Jui and Wang, William Yang and McDuff, Daniel and Song, Yale}, year={2022}, month=jun, pages={634–642} }

@InProceedings{gao2023pal,
  title = 	 {{PAL}: Program-aided Language Models},
  author =       {Gao, Luyu and Madaan, Aman and Zhou, Shuyan and Alon, Uri and Liu, Pengfei and Yang, Yiming and Callan, Jamie and Neubig, Graham},
  booktitle = 	 {Proceedings of the 40th International Conference on Machine Learning},
  pages = 	 {10764--10799},
  year = 	 {2023},
  editor = 	 {Krause, Andreas and Brunskill, Emma and Cho, Kyunghyun and Engelhardt, Barbara and Sabato, Sivan and Scarlett, Jonathan},
  volume = 	 {202},
  series = 	 {Proceedings of Machine Learning Research},
  month = 	 {23--29 Jul},
  publisher =    {PMLR},
  url = 	 {https://proceedings.mlr.press/v202/gao23f.html}
}

@inproceedings{ge2025autopresent, title={{AutoPresent}: Designing Structured Visuals from Scratch}, url={http://dx.doi.org/10.1109/CVPR52734.2025.00276}, DOI={10.1109/cvpr52734.2025.00276}, booktitle={2025 IEEE/CVF Conference on Computer Vision and Pattern Recognition (CVPR)}, publisher={IEEE}, author={Ge, Jiaxin and Wang, Zora Zhiruo and Zhou, Xuhui and Peng, Yi-Hao and Subramanian, Sanjay and Tan, Qinyue and Sap, Maarten and Suhr, Alane and Fried, Daniel and Neubig, Graham and Darrell, Trevor}, year={2025}, month=jun, pages={2902–2911} }

@inproceedings{guan2024hallusionbench, title={{Hallusionbench}: An Advanced Diagnostic Suite for Entangled Language Hallucination and Visual Illusion in Large Vision-Language Models}, url={http://dx.doi.org/10.1109/CVPR52733.2024.01363}, DOI={10.1109/cvpr52733.2024.01363}, booktitle={2024 IEEE/CVF Conference on Computer Vision and Pattern Recognition (CVPR)}, publisher={IEEE}, author={Guan, Tianrui and Liu, Fuxiao and Wu, Xiyang and Xian, Ruiqi and Li, Zongxia and Liu, Xiaoyu and Wang, Xijun and Chen, Lichang and Huang, Furong and Yacoob, Yaser and Manocha, Dinesh and Zhou, Tianyi}, year={2024}, month=jun, pages={14375–14385} }

@inproceedings{hsu2023posterlayout, title={{PosterLayout}: A New Benchmark and Approach for Content-Aware Visual-Textual Presentation Layout}, url={http://dx.doi.org/10.1109/CVPR52729.2023.00583}, DOI={10.1109/cvpr52729.2023.00583}, booktitle={2023 IEEE/CVF Conference on Computer Vision and Pattern Recognition (CVPR)}, publisher={IEEE}, author={Hsu, HsiaoYuan and He, Xiangteng and Peng, Yuxin and Kong, Hao and Zhang, Qing}, year={2023}, month=jun, pages={6018–6026} }

@inproceedings{hsu2025postero, title={{PosterO}: Structuring Layout Trees to Enable Language Models in Generalized Content-Aware Layout Generation}, url={http://dx.doi.org/10.1109/CVPR52734.2025.00760}, DOI={10.1109/cvpr52734.2025.00760}, booktitle={2025 IEEE/CVF Conference on Computer Vision and Pattern Recognition (CVPR)}, publisher={IEEE}, author={Hsu, HsiaoYuan and Peng, Yuxin}, year={2025}, month=jun, pages={8117–8127} }

@inproceedings{huang2022layoutlmv3, series={MM ’22}, title={{LayoutLMv3}: Pre-training for Document {AI} with Unified Text and Image Masking}, url={http://dx.doi.org/10.1145/3503161.3548112}, DOI={10.1145/3503161.3548112}, booktitle={Proceedings of the 30th ACM International Conference on Multimedia}, publisher={ACM}, author={Huang, Yupan and Lv, Tengchao and Cui, Lei and Lu, Yutong and Wei, Furu}, year={2022}, month=Oct, pages={4083–4091}, collection={MM ’22} }

@misc{lee2024prometheusvision,
      title={{Prometheus-Vision}: Vision-Language Model as a Judge for Fine-Grained Evaluation}, 
      author={Seongyun Lee and Seungone Kim and Sue Hyun Park and Geewook Kim and Minjoon Seo},
      year={2024},
      eprint={2401.06591},
      howpublished={arXiv preprint arXiv:2401.06591},
      archivePrefix={arXiv},
      primaryClass={cs.CL},
      url={https://arxiv.org/abs/2401.06591}, 
}

@inproceedings{li2020docbank,
    title = "{D}oc{B}ank: A Benchmark Dataset for Document Layout Analysis",
    author = "Li, Minghao  and
      Xu, Yiheng  and
      Cui, Lei  and
      Huang, Shaohan  and
      Wei, Furu  and
      Li, Zhoujun  and
      Zhou, Ming",
    editor = "Scott, Donia  and
      Bel, Nuria  and
      Zong, Chengqing",
    booktitle = "Proceedings of the 28th International Conference on Computational Linguistics",
    month = dec,
    year = "2020",
    address = "Barcelona, Spain (Online)",
    publisher = "International Committee on Computational Linguistics",
    url = "https://aclanthology.org/2020.coling-main.82/",
    doi = "10.18653/v1/2020.coling-main.82",
    pages = "949--960"
}

@inproceedings{lin2004rouge,
    title = "{ROUGE}: A Package for Automatic Evaluation of Summaries",
    author = "Lin, Chin-Yew",
    booktitle = "Text Summarization Branches Out",
    month = jul,
    year = "2004",
    address = "Barcelona, Spain",
    publisher = "Association for Computational Linguistics",
    url = "https://aclanthology.org/W04-1013/",
    pages = "74--81"
}

@inproceedings{lin2023layoutprompter,
 author = {Lin, Jiawei and Guo, Jiaqi and Sun, Shizhao and Yang, Zijiang and Lou, Jian-Guang and Zhang, Dongmei},
 booktitle = {Advances in Neural Information Processing Systems},
 doi = {10.52202/075280-1902},
 editor = {A. Oh and T. Naumann and A. Globerson and K. Saenko and M. Hardt and S. Levine},
 pages = {43852--43879},
 publisher = {Curran Associates, Inc.},
 title = {{LayoutPrompter}: Awaken the Design Ability of Large Language Models},
 url = {https://proceedings.neurips.cc/paper_files/paper/2023/file/88a129e44f25a571ae8b838057c46855-Paper-Conference.pdf},
 volume = {36},
 year = {2023}
}

@misc{livathinos2025docling,
      title={Docling: An Efficient Open-Source Toolkit for {AI}-driven Document Conversion}, 
      author={Nikolaos Livathinos and Christoph Auer and Maksym Lysak and Ahmed Nassar and Michele Dolfi and Panos Vagenas and Cesar Berrospi Ramis and Matteo Omenetti and Kasper Dinkla and Yusik Kim and Shubham Gupta and Rafael Teixeira de Lima and Valery Weber and Lucas Morin and Ingmar Meijer and Viktor Kuropiatnyk and Peter W. J. Staar},
      year={2025},
      eprint={2501.17887},
      howpublished={arXiv preprint arXiv:2501.17887},
      archivePrefix={arXiv},
      primaryClass={cs.CL},
      url={https://arxiv.org/abs/2501.17887}, 
}

@article{qiang2016learning, title={Learning to Generate Posters of Scientific Papers}, volume={30}, ISSN={2159-5399}, url={http://dx.doi.org/10.1609/aaai.v30i1.10000}, DOI={10.1609/aaai.v30i1.10000}, number={1}, journal={Proceedings of the AAAI Conference on Artificial Intelligence}, publisher={Association for the Advancement of Artificial Intelligence (AAAI)}, author={Qiang, Yuting and Fu, Yanwei and Guo, Yanwen and Zhou, Zhi-Hua and Sigal, Leonid}, year={2016}, month=Feb }

@article{qiang2019learning, title={Learning to Generate Posters of Scientific Papers by Probabilistic Graphical Models}, volume={34}, ISSN={1860-4749}, url={http://dx.doi.org/10.1007/s11390-019-1904-1}, DOI={10.1007/s11390-019-1904-1}, number={1}, journal={Journal of Computer Science and Technology}, publisher={Springer Science and Business Media LLC}, author={Qiang, Yu-Ting and Fu, Yan-Wei and Yu, Xiao and Guo, Yan-Wen and Zhou, Zhi-Hua and Sigal, Leonid}, year={2019}, month=Jan, pages={155–169} }

@inproceedings{seol2024posterllama, title={{PosterLlama}: Bridging Design Ability of Language Model to Content-Aware Layout Generation}, ISBN={9783031730078}, ISSN={1611-3349}, url={http://dx.doi.org/10.1007/978-3-031-73007-8_26}, DOI={10.1007/978-3-031-73007-8_26}, booktitle={Computer Vision – ECCV 2024}, publisher={Springer Nature Switzerland}, author={Seol, Jaejung and Kim, Seojun and Yoo, Jaejun}, year={2024}, month=Oct, pages={451–468} }

@inproceedings{shinn2023reflexion, series={NeurIPS 2023}, title={Reflexion: language agents with verbal reinforcement learning}, url={http://dx.doi.org/10.52202/075280-0377}, DOI={10.52202/075280-0377}, booktitle={Advances in Neural Information Processing Systems 36}, publisher={Neural Information Processing Systems Foundation, Inc. (NeurIPS)}, author={Shinn, Noah and Cassano, Federico and Gopinath, Ashwin and Narasimhan, Karthik and Yao, Shunyu}, year={2023}, pages={8634–8652}, collection={NeurIPS 2023} }

@inproceedings{si2025design2code,
    title = "{D}esign2{C}ode: Benchmarking Multimodal Code Generation for Automated Front-End Engineering",
    author = "Si, Chenglei  and
      Zhang, Yanzhe  and
      Li, Ryan  and
      Yang, Zhengyuan  and
      Liu, Ruibo  and
      Yang, Diyi",
    editor = "Chiruzzo, Luis  and
      Ritter, Alan  and
      Wang, Lu",
    booktitle = "Proceedings of the 2025 Conference of the Nations of the Americas Chapter of the Association for Computational Linguistics: Human Language Technologies (Volume 1: Long Papers)",
    month = apr,
    year = "2025",
    address = "Albuquerque, New Mexico",
    publisher = "Association for Computational Linguistics",
    url = "https://aclanthology.org/2025.naacl-long.199/",
    doi = "10.18653/v1/2025.naacl-long.199",
    pages = "3956--3974",
    ISBN = "979-8-89176-189-6"
}

@inproceedings{sun2021d2s,
    title = "{D}2{S}: Document-to-Slide Generation Via Query-Based Text Summarization",
    author = "Sun, Edward  and
      Hou, Yufang  and
      Wang, Dakuo  and
      Zhang, Yunfeng  and
      Wang, Nancy X. R.",
    editor = "Toutanova, Kristina  and
      Rumshisky, Anna  and
      Zettlemoyer, Luke  and
      Hakkani-Tur, Dilek  and
      Beltagy, Iz  and
      Bethard, Steven  and
      Cotterell, Ryan  and
      Chakraborty, Tanmoy  and
      Zhou, Yichao",
    booktitle = "Proceedings of the 2021 Conference of the North American Chapter of the Association for Computational Linguistics: Human Language Technologies",
    month = jun,
    year = "2021",
    address = "Online",
    publisher = "Association for Computational Linguistics",
    url = "https://aclanthology.org/2021.naacl-main.111/",
    doi = "10.18653/v1/2021.naacl-main.111",
    pages = "1405--1418"
}

@misc{tanaka2024scipostlayout,
      title={{SciPostLayout}: A Dataset for Layout Analysis and Layout Generation of Scientific Posters}, 
      author={Shohei Tanaka and Hao Wang and Yoshitaka Ushiku},
      year={2024},
      eprint={2407.19787},
      howpublished={arXiv preprint arXiv:2407.19787},
      archivePrefix={arXiv},
      primaryClass={cs.CV},
      url={https://arxiv.org/abs/2407.19787}, 
}

@misc{xu2021neural,
      title={Neural Content Extraction for Poster Generation of Scientific Papers}, 
      author={Sheng Xu and Xiaojun Wan},
      year={2021},
      eprint={2112.08550},
      howpublished={arXiv preprint arXiv:2112.08550},
      archivePrefix={arXiv},
      primaryClass={cs.CL},
      url={https://arxiv.org/abs/2112.08550}, 
}

@article{xu2022posterbot, title={{PosterBot}: A System for Generating Posters of Scientific Papers with Neural Models}, volume={36}, ISSN={2159-5399}, url={http://dx.doi.org/10.1609/aaai.v36i11.21738}, DOI={10.1609/aaai.v36i11.21738}, number={11}, journal={Proceedings of the AAAI Conference on Artificial Intelligence}, publisher={Association for the Advancement of Artificial Intelligence (AAAI)}, author={Xu, Sheng and Wan, Xiaojun}, year={2022}, month=jun, pages={13233–13235} }

@inproceedings{yang2024sweagent,
 author = {Yang, John and Jimenez, Carlos and Wettig, Alexander and Lieret, Kilian and Yao, Shunyu and Narasimhan, Karthik and Press, Ofir},
 booktitle = {Advances in Neural Information Processing Systems},
 doi = {10.52202/079017-1601},
 editor = {A. Globerson and L. Mackey and D. Belgrave and A. Fan and U. Paquet and J. Tomczak and C. Zhang},
 pages = {50528--50652},
 publisher = {Curran Associates, Inc.},
 title = {{SWE-agent}: Agent-Computer Interfaces Enable Automated Software Engineering},
 url = {https://proceedings.neurips.cc/paper_files/paper/2024/file/5a7c947568c1b1328ccc5230172e1e7c-Paper-Conference.pdf},
 volume = {37},
 year = {2024}
}

@inproceedings{yao2023react,
title={{ReAct}: Synergizing Reasoning and Acting in Language Models},
author={Shunyu Yao and Jeffrey Zhao and Dian Yu and Nan Du and Izhak Shafran and Karthik R Narasimhan and Yuan Cao},
booktitle={The Eleventh International Conference on Learning Representations },
year={2023},
url={https://openreview.net/forum?id=WE_vluYUL-X}
}

@inproceedings{zhang2020bertscore,
title={{BERTScore}: Evaluating Text Generation with {BERT}},
author={Tianyi Zhang and Varsha Kishore and Felix Wu and Kilian Q. Weinberger and Yoav Artzi},
booktitle={International Conference on Learning Representations},
year={2020},
url={https://openreview.net/forum?id=SkeHuCVFDr}
}

@misc{zheng2025pptagent,
      title={{PPTAgent}: Generating and Evaluating Presentations Beyond Text-to-Slides}, 
      author={Hao Zheng and Xinyan Guan and Hao Kong and Jia Zheng and Weixiang Zhou and Hongyu Lin and Yaojie Lu and Ben He and Xianpei Han and Le Sun},
      year={2025},
      eprint={2501.03936},
      howpublished={arXiv preprint arXiv:2501.03936},
      archivePrefix={arXiv},
      primaryClass={cs.AI},
      url={https://arxiv.org/abs/2501.03936}, 
}

@inproceedings{zhong2019publaynet, title={{PubLayNet}: Largest Dataset Ever for Document Layout Analysis}, url={http://dx.doi.org/10.1109/ICDAR.2019.00166}, DOI={10.1109/icdar.2019.00166}, booktitle={2019 International Conference on Document Analysis and Recognition (ICDAR)}, publisher={IEEE}, author={Zhong, Xu and Tang, Jianbin and Jimeno Yepes, Antonio}, year={2019}, month=sep, pages={1015–1022} }

@article{zhong2025scientific, title={Scientific poster generation: A new dataset and approach}, volume={164}, ISSN={0031-3203}, url={http://dx.doi.org/10.1016/j.patcog.2025.111507}, DOI={10.1016/j.patcog.2025.111507}, journal={Pattern Recognition}, publisher={Elsevier BV}, author={Zhong, Xinyi and Tan, Zusheng and Li, Jing and Gao, Shen and Ma, Jing and Feng, Shanshan and Chiu, Billy}, year={2025}, month=Aug, pages={111507} }

@misc{paruchuri2025marker,
  title = {{Marker: Convert PDF to Markdown and JSON Quickly with High Accuracy}},
  author = {Vik Paruchuri},
  howpublished = {\url{https://github.com/datalab-to/marker}},
  note = {GitHub repository, accessed September 2026},
  year = {2025}
}

@misc{wang2024mineru,
      title={{MinerU}: An Open-Source Solution for Precise Document Content Extraction}, 
      author={Bin Wang and Chao Xu and Xiaomeng Zhao and Linke Ouyang and Fan Wu and Zhiyuan Zhao and Rui Xu and Kaiwen Liu and Yuan Qu and Fukai Shang and Bo Zhang and Liqun Wei and Zhihao Sui and Wei Li and Botian Shi and Yu Qiao and Dahua Lin and Conghui He},
      year={2024},
      eprint={2409.18839},
      howpublished={arXiv preprint arXiv:2409.18839},
      archivePrefix={arXiv},
      primaryClass={cs.CV},
      url={https://arxiv.org/abs/2409.18839}, 
}
\bibliographystyle{preprint}

\clearpage
\appendix

\section{Implementation Details}
\label{app:implementation}

\subsection{Semantics-Directed Asset Parsing (SemParser)}
\label{app:asset-parsing}

\paragraph{LLM input.}
Text lines are extracted with PyMuPDF together with their pages and boxes. The boxes are used only by the program. Lines keep the native extraction order of the PDF. Headings divide the text into body, appendix, and references. References are always excluded, and appendix lines are kept in our configuration. The LLM receives the remaining lines as numbered pairs $(i,s_i)$ separated by page markers, without page images or line coordinates. Its only geometric input is a set of candidate boundary rules per page: long vector strokes (horizontal strokes of at least 18\% of the page width, vertical strokes of at least 12\% of the page height, centered within 42\% of the page size from an edge, at most 12 per side), each described by the identifiers of neighboring lines and by its repetition across pages.

\paragraph{LLM output.}
The LLM $f$ of Equation~\ref{eq:role-assignment} reads the page-delimited pairs $(i,s_i)$, excluding references. The box $b_i$ of each line on its page is used only by the program. Apart from candidate rules that may bound the page content, the LLM sees only text, without page images or line coordinates. In one call under a strict schema, the LLM assigns each contiguous range of lines one of 15 closed roles. Figure captions carry an element key (family and number) shared by their wrapped lines, subfigure captions carry their parent figure, and text inside graphics is labeled as figure content. The LLM also selects the candidate rules that bound the content, returns structured records for tables (headers, rows, footnotes, and emphasis), equations (\LaTeX{} or empty), and algorithms (steps), each with the identifiers of its source lines and a self-reported status (ok, partial, or unresolved) that the program does not verify, and extracts the paper metadata. The call is repeated only when its output is missing or fails schema validation, and the repeated call includes the validator's message. Coverage and crop quality never trigger a repetition.

\paragraph{Anchors and walls.}
On each page, caption boxes with the same element key are merged into one anchor, and a panel label serves as an anchor only when the page contains no caption for its figure. Walls are the lines labeled as headings, title and authors, body text, footnotes, page numbers, running headers and footers, references, captions of other figures, table titles, equation captions, and algorithms, together with the lines of excluded sections, the selected boundary rules, and the rules that delimit tables and algorithms. Figure content and subfigure captions are neither walls nor anchors. Each wall is placed at the extent actually drawn for its glyphs, which may exceed the nominal text box, and its edge is refined on a local $4\times$ raster without padding.

\paragraph{Region growing.}
The roles divide the lines of a page into the anchor $a_e$ (the caption of $e$), the walls $\mathcal{W}$ (lines that belong elsewhere, together with rules that bound the page content or delimit tables and algorithms), and the interior lines (text inside graphics and subfigure captions), which are neither. With $R_e=X_e\times Y_e$, the region grows from the anchor through alternating updates (Algorithm~\ref{alg:asset-parsing}),
\begin{equation}
  Y_e\leftarrow\operatorname{ext}_y\big(a_e\mid X_e;\ \mathcal{W}\big),\qquad
  X_e\leftarrow\operatorname{ext}_x\big(a_e\mid Y_e;\ \mathcal{W}\big),
  \label{eq:region-growing}
\end{equation}
where $\operatorname{ext}_y$ extends the vertical interval away from the caption up to the nearest wall whose horizontal extent overlaps $X_e$, and $\operatorname{ext}_x$ acts analogously. Because each update stops at the nearest wall that overlaps the other interval, every update preserves $R_e\cap\operatorname{ink}(\mathcal{W})=\emptyset$: text that belongs elsewhere can enter the region only if $r$ fails to label it as a wall. Conversely, interior lines never stop the growth, so subfigure captions, axis labels, and legends are retained without being detected individually, and panels that share a caption grow into a single region. Because figures are enumerated from their recognized captions rather than detected, a figure can be missed only if its caption is.

\begin{algorithm}[t]
\caption{Role-constrained figure extraction.}
\label{alg:asset-parsing}
\begin{minipage}{0.62\textwidth}
\small
\begin{algorithmic}[1]
\Require lines with roles $r$, content-boundary rules
\For{each page $p$}
  \State $\mathcal{W}\gets$ wall lines, boundary rules, table and algorithm rules
  \State place each wall at its drawn glyph extent
  \For{each figure $e$ with caption $a_e$ on $p$}
    \State $R_e\gets a_e$
    \For{$k=1,\dots,k_{\max}$} \Comment{three rounds}
      \State $Y_e\gets\operatorname{ext}_y(a_e\mid X_e;\mathcal{W})$ \Comment{Eq.~\ref{eq:region-growing}}
      \State $X_e\gets\operatorname{ext}_x(a_e\mid Y_e;\mathcal{W})$
    \EndFor
  \EndFor
  \State split adjacent regions about their captions or at gaps
  \State tighten each $R_e$ to its ink
  \State render each $R_e$ from the vector PDF
\EndFor
\end{algorithmic}
\end{minipage}
\end{algorithm}

\paragraph{Consequences.}
Because walls come from text roles, text that belongs elsewhere can enter a region only if the LLM fails to label it as a wall, whereas subfigure captions, axis labels, and legends are retained without being detected individually, and panels that share a caption grow into a single region. Because figures are enumerated from their recognized captions rather than detected, a figure can be missed only if its caption is.

\paragraph{Growth, resolution, and rendering.}
A region grows from the edge of its caption toward the nearest walls, alternating vertical and horizontal extension for three rounds. It grows upward first and downward only when no ink lies above the caption. If a region collapses, the ink is clustered into connected components instead. Figures placed side by side are separated symmetrically about their captions or at white gaps, and stacked figures at walls or horizontal gaps. Page-level arbitration, cross-page adoption, and column clipping resolve the remaining conflicts. Whitespace is trimmed on a $4\times$ raster (288 dpi), and the final crop is rendered again from the PDF at $6\times$ (432 dpi) without padding. No vision model is used at any point of the extraction.

\paragraph{Asset package.}
The package contains the figures (with their pixel sizes, pages, PDF boxes, sections, and priorities), their captions, the metadata, the numbered source lines, and the structured records. Body figures form the primary pool and appendix figures a reserve. Tables are transcribed but not included as images. A paper without figures still yields a complete package.

\subsection{Vision-Centric Layout Synthesis (VisComposer)}
\label{app:layout-synthesis}

\begin{algorithm}[t]
\caption{Closed-loop layout synthesis in VisComposer.}
\label{alg:method}
\begin{minipage}{0.62\textwidth}
\small
\begin{algorithmic}[1]
\Require assets $\mathcal{A}$, canvas $\Omega$, bound $K$
\State $(C_0,S_0)\gets\pi(\mathcal{A})$ \Comment{complete draft}
\For{$t=0,1,\dots,K$}
  \State $\{B_v\}_t\gets$ measure $\Gamma(C_t)$ by identifier
  \State compute $F_t$ and $g_t$ \Comment{Eq.~\ref{eq:feedback}}
  \State store checkpoint $(C_t,S_t,F_t)$
  \If{$g_t=1$ \textbf{or} $t=K$} \State \textbf{break} \EndIf
  \State $(\Delta_t,S_{t+1})\sim\pi(\cdot\mid\mathcal{A},\mathcal{H}_t,I_t,F_t)$
  \State $C_{t+1}\gets\operatorname{Apply}(C_t,\Delta_t)$ \Comment{all or none}
\EndFor
\State $t^{\star}\gets$ best stored checkpoint \Comment{App.~\ref{app:implementation}}
\If{$C_{t^{\star}}$ only overflows slightly inside containers}
  \State store a copy with reduced padding, re-check it, and update $t^{\star}$
\EndIf
\State \Return $C_{t^{\star}}$
\end{algorithmic}
\end{minipage}
\end{algorithm}

\paragraph{Inputs and instructions.}
The agent first receives contact sheets of the body figures (labeled thumbnails in two columns), of the appendix figures, and of the brand assets, followed by a structured package: the metadata, a catalog of assets (kind, scope, priority, label, page, pixel size, aspect ratio, caption, and palette candidates), the appendix policy, the brand requirements, the structured records, and all numbered source lines. Captions and sizes appear only in the package, and the design agent never receives figures at full resolution. The instructions ask the agent to design claim--evidence units rather than separate text and image columns, placing every figure, table, equation, or algorithm next to the claim it explains, supports, or derives. They also ask it to make one visual group the unmistakable primary anchor, to decide the claim--evidence graph and the stages first and then choose text amount, evidence size, territories, typography, and geometry jointly, to give every admitted figure enough area for its specific point to be read and never keep a figure as a decorative thumbnail, to treat appendix figures as a reserve, and to avoid a rigid reusable template. The only layout priors are a fixed landscape canvas ($2544\times1696$\,px in our runs), a narrative that advances from left to right through framing, overview, mechanism, and results, four to six stages, and two to five territories. Tables, equations, and algorithms are typeset in HTML from their records. Scripts are disabled at rendering, so no math-rendering library is available.

\paragraph{Design state.}
The declared state $S=(G,\mathcal{Q},\mathcal{T},v^{\star},o)$ contains four to six stages, each with a role (framing, overview, mechanism, result, or takeaway), a title, a purpose, and its member nodes. It also contains two to five territories, each with a relative width in $[0.5,8]$ and its member nodes, 6 to 36 nodes of kind claim, figure, table, equation, algorithm, transition, context, conclusion, or branding, each with a semantic role, a stage, a summary, and a priority from 1 to 3, and 3 to 80 typed edges (explains, supports, derives, qualifies, contrasts, or continues). Finally, it contains a reading order $o$, a visual hierarchy with the primary anchor $v^{\star}$ and at most three secondary anchors, a type scale, and a decision log. Elements of the HTML carry the node, stage, and territory identifiers as data attributes.

\paragraph{Rendering and measurement.}
Every source is rendered in headless Chromium with the viewport equal to the canvas, a device pixel ratio of 1, scripts disabled, and external requests blocked. The displayed size $(\tilde{w}_v,\tilde{h}_v)$ of a figure is the size of its image after aspect-preserving fitting into its box.

\paragraph{Predicates.}
A checkpoint is accepted if and only if no predicate below is violated. All of them are hard constraints.
\begin{itemize}
  \item \emph{Source contract}, checked before rendering: only allowed resources, unique identifiers that are all present in the HTML, existing referenced nodes, and no isolated evidence node. A checkpoint that fails it is neither rendered nor ranked.
  \item \emph{Design-state contract}: required fields, counts, and enumerations.
  \item \emph{Narrative structure}: stage roles are complete and in order, and each node's stage is consistent with its semantic role. Every non-identity node lies in exactly one stage and every node in exactly one territory. The reading order covers all nodes and never returns to an earlier stage, and every node and stage is rendered exactly once. An evidence node is the primary anchor, and title and metadata are reproduced verbatim.
  \item \emph{Geometry}: no overflow beyond 4\,px (with a tolerance for headings, paragraphs, and list items), no content more than 1\,px outside the canvas, no missing image, no empty container (a height of at least 180\,px with a blank ratio above 0.28), no flex gap above 4.5\% of the territory height, no side container holding only a title, and no rotated or vertical text.
  \item \emph{Typography}: a title of at least 64\,px, a median text size of at least 19\,px, a 10th percentile of at least 16.5\,px, at most four text leaves below 17\,px, chapter headings of at least 30\,px, and consistent chapter styles.
  \item \emph{Whitespace}: within each territory, the largest vertical void, the painted bottom void, and the spread of bottom baselines do not exceed 0.075, 0.035, and 0.025, respectively.
  \item \emph{Figure floor}: every figure node satisfies $\tilde{w}_v\ge\alpha W$ and $\tilde{h}_v\ge\beta H$ with $\alpha=0.12$ and $\beta=0.065$ (305.28\,px and 110.24\,px on our canvas). The violation message asks the agent to enlarge or omit the figure.
  \item \emph{Primary anchor}: rendered, with an area of at least $\gamma WH$, $\gamma=0.035$.
  \item \emph{Evidence proximity}: for every edge with at least one evidence endpoint, the distance between the centers of the two boxes is at most $\delta\sqrt{W^2+H^2}$, $\delta=0.42$.
  \item \emph{Left-to-right narrative}: ordered by role, the horizontal centers of the stage bounding boxes never decrease by more than $\tau=80$\,px.
  \item \emph{Brand and QR contract}.
\end{itemize}
Warnings that do not block acceptance cover fewer than three chapters, large dark blocks, a territory coverage below 0.47, and a horizontal stage progression below 40\%. The order of elements in the HTML is not checked against the declared reading order.

\paragraph{Revision.}
With every revision request, the agent receives the full rendering of the previous checkpoint, an evidence sheet that crops each figure node from its rendered rectangle and labels it with its displayed size, and a controller state with the acceptance flag, the violated predicates with their measurements, semantic warnings, font statistics, distant claim--evidence pairs, stage centers, and geometry. Intermediate rounds are asked to resolve evidence admission, figure scale, text quantity, claim--evidence proximity, and territory allocation together, and the last round is asked to converge without adding scientific figures or starting a redesign. This distinction is only instructed. The session is resumed across rounds through the session identifier of the backend. The first round returns the complete HTML, CSS, design state, and a critique. Each revision returns at most 16 replacements of exact substrings, each with a reason, together with the complete design state. The harness applies the replacements atomically and rejects the whole round if any replaced text does not occur exactly once. The design agent has no write access to files.

\paragraph{Stopping, selection, and spacing adjustment.}
The loop allows at most $K$ revisions ($K=2$ in our main runs, i.e., at most three LLM calls) and stops as soon as a checkpoint is accepted. The returned checkpoint maximizes a lexicographic order of 16 keys that starts with acceptance, followed by fewer semantic issues, fewer out-of-canvas and missing-image issues, fewer overflow pixels, fewer hard overflows, fewer empty containers, and fewer undersized figures, and ends with fewer issues in total. Ties favor later checkpoints. If the best checkpoint has no semantic, out-of-canvas, missing-image, or empty-container issue and its overflow is confined to containers (at most 4\,px horizontally, 4 to 48\,px per vertical overflow, and at most 64\,px in total), the harness appends CSS that only reduces padding and margins, renders and checks the result again, and ranks it with the other checkpoints. This adjustment never changes font sizes, line heights, content, evidence dimensions, or global tracks.

\section{Complete Quantitative Results}
\label{app:results}

\subsection{Key-Figure Presentation Scale}
\label{app:pilot-protocols}

\begin{figure}[h]
\centering
\includegraphics[width=0.7\linewidth]{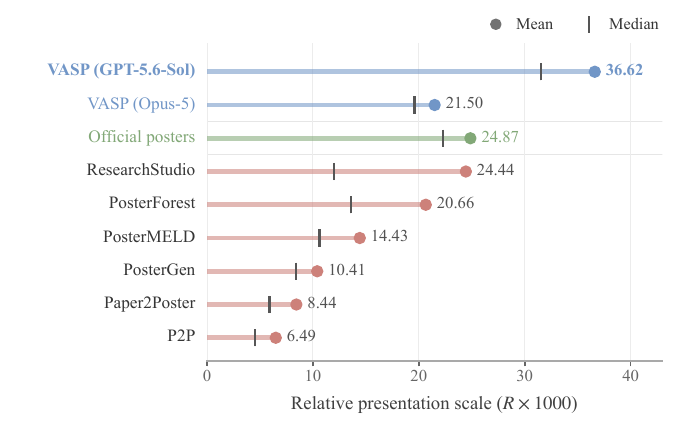}
\caption{Relative presentation scale ($R\times1000$) of each paper's most important figure on the 74 papers whose official posters display it. Dots and vertical ticks indicate means and medians. Full numerical results appear in Table~\ref{tab:pilot-scale}.}
\label{fig:pilot-study}
\end{figure}

To measure the figure-presentation gain in Section~\ref{sec:poster-level-quality} directly, we compare how large each paper's most important figure is displayed relative to the body text. On the 74 papers whose official posters show that figure, the median relative scale ($R\times10^{3}$) is 31.5 for VASP with GPT-5.6-Sol and 19.6 with Opus-5, compared with 22.3 for the official posters and at most 13.6 for existing approaches (Table~\ref{tab:pilot-scale} and Figure~\ref{fig:pilot-study}).

\begin{table}[h]
\centering
\caption{Relative presentation scale of each paper's most important figure on the 74 papers whose official posters display it. Mean and median report $R\times1000$. A missing figure receives zero. Displayed counts the posters that contain the figure. Higher is better, and among the generated posters the best result is in bold and the second best is underlined.}
\label{tab:pilot-scale}
\small
\setlength{\tabcolsep}{10pt}
\begin{tabular}{lccc}
    \hline
    Method & Mean$\uparrow$ & Median$\uparrow$ & Displayed$\uparrow$ \\
    \hline
    \multicolumn{4}{l}{\textit{Ours}} \\
    \textbf{VASP (Opus-5)}      & 21.5 & \underline{19.6} & \underline{73/74} \\
    \textbf{VASP (GPT-5.6-Sol)} & \textbf{36.6} & \textbf{31.5} & \textbf{74/74} \\
    \hline
    \multicolumn{4}{l}{\textit{Baseline methods}} \\
    Paper2Poster       & 8.4 & 5.9 & 68/74 \\
    PosterMELD         & 14.4 & 10.6 & 54/74 \\
    PosterForest       & 20.7 & 13.6 & 63/74 \\
    PosterGen          & 10.4 & 8.4 & 67/74 \\
    P2P                & 6.5 & 4.6 & \underline{73/74} \\
    ResearchStudio     & \underline{24.4} & 12.0 & 62/74 \\
    \hline
    \multicolumn{4}{l}{\textit{Human reference}} \\
    Official posters   & 24.9 & 22.3 & 74/74 \\
    \hline
\end{tabular}
\end{table}

\paragraph{Measurement protocol.}

GPT-5.6-Sol identifies the most important figure among the extracted visual assets, with human verification resolving ambiguous selections. Of the 100 papers, 74 have official posters that display this figure. We restrict the comparison to these 74 papers so that every method is compared with the authors' own presentation of the same figure. Each figure instance is aligned to the corresponding source PDF region. Its area scaling factor is measured from PDF square points to rendered-poster square pixels. For multiple corresponding subfigures or repeated instances, we use the smallest area scaling factor. If the reference figure is absent, its presentation scale is zero. Character area is the tight bounding box containing all visible strokes of the annotated uppercase body-text character, rather than the number of foreground pixels. For generated posters, we use the same uppercase character where available and another uppercase body-text character otherwise. The ratio of figure area scaling to character area cancels output-resolution changes, so no additional poster-area normalization is applied.

\subsection{Ablation Study}
\label{sec:ablation}

\begin{figure}[h]
\centering
\begin{minipage}[c]{0.49\textwidth}\centering
\includegraphics[width=\linewidth]{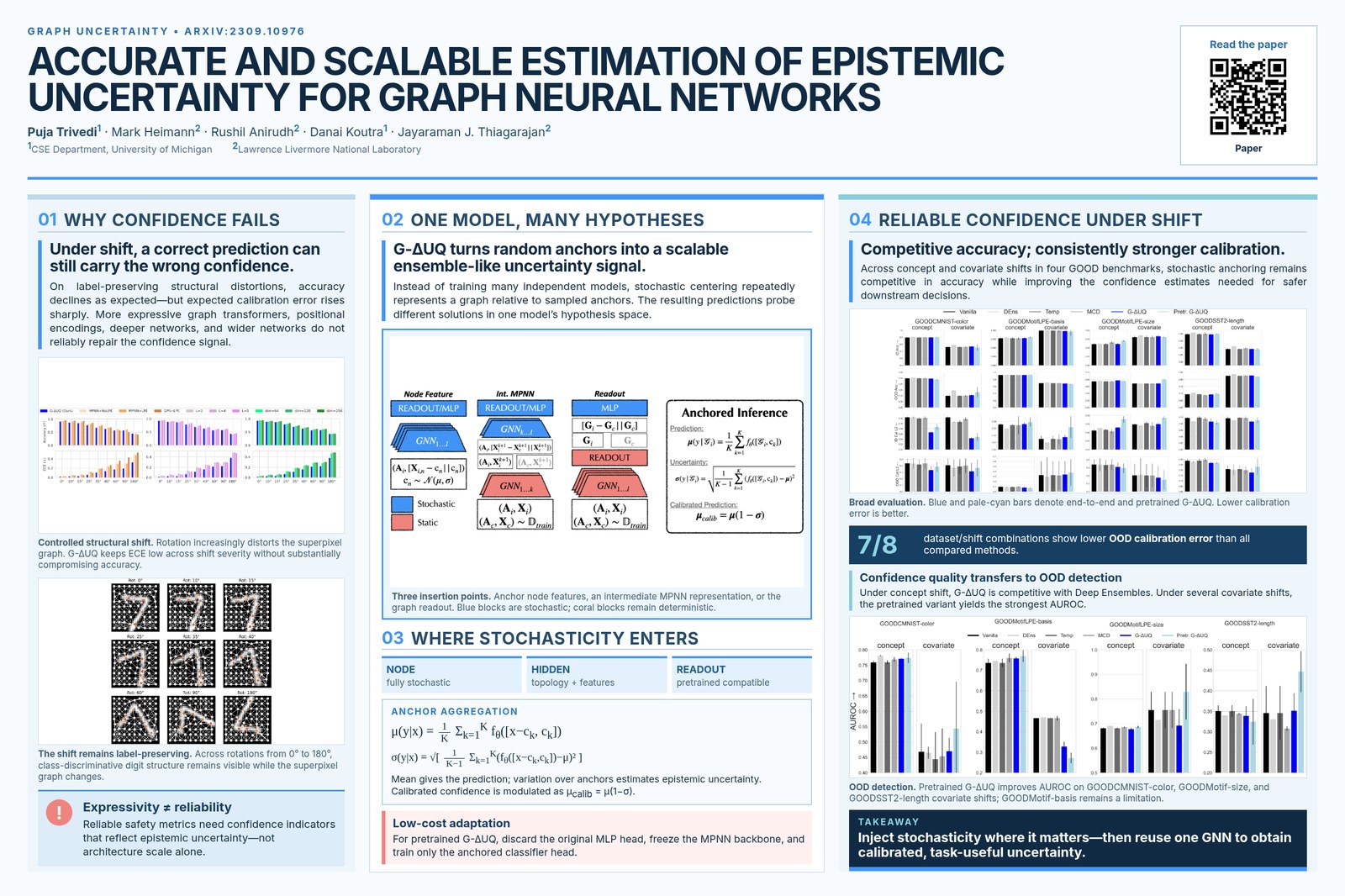}\\[2pt]
{\small (a) VASP (full)}
\end{minipage}\hfill
\begin{minipage}[c]{0.49\textwidth}\centering
\begin{minipage}[t]{0.49\linewidth}\centering
\includegraphics[width=\linewidth]{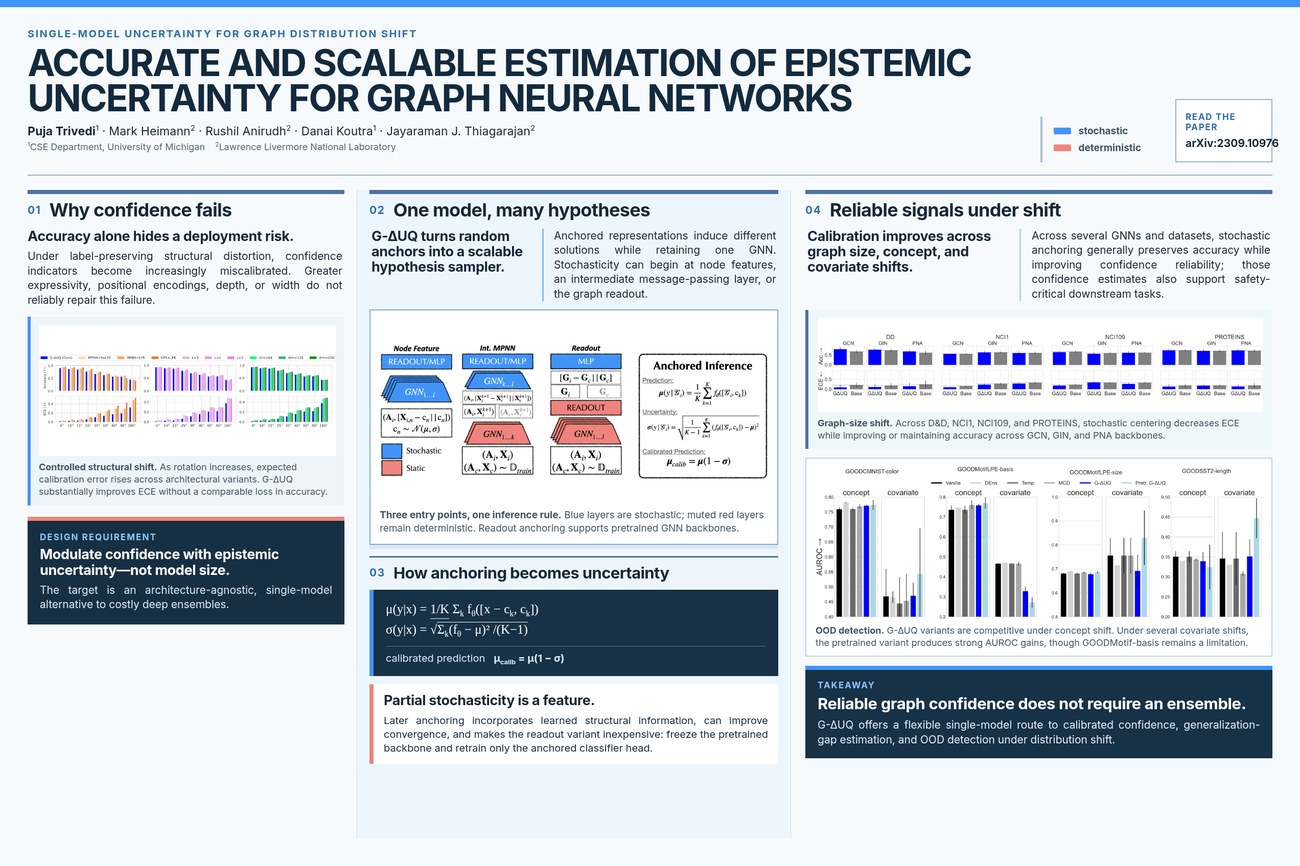}\\[1pt]
{\small (b) w/o SemParser}
\end{minipage}\hfill
\begin{minipage}[t]{0.49\linewidth}\centering
\includegraphics[width=\linewidth]{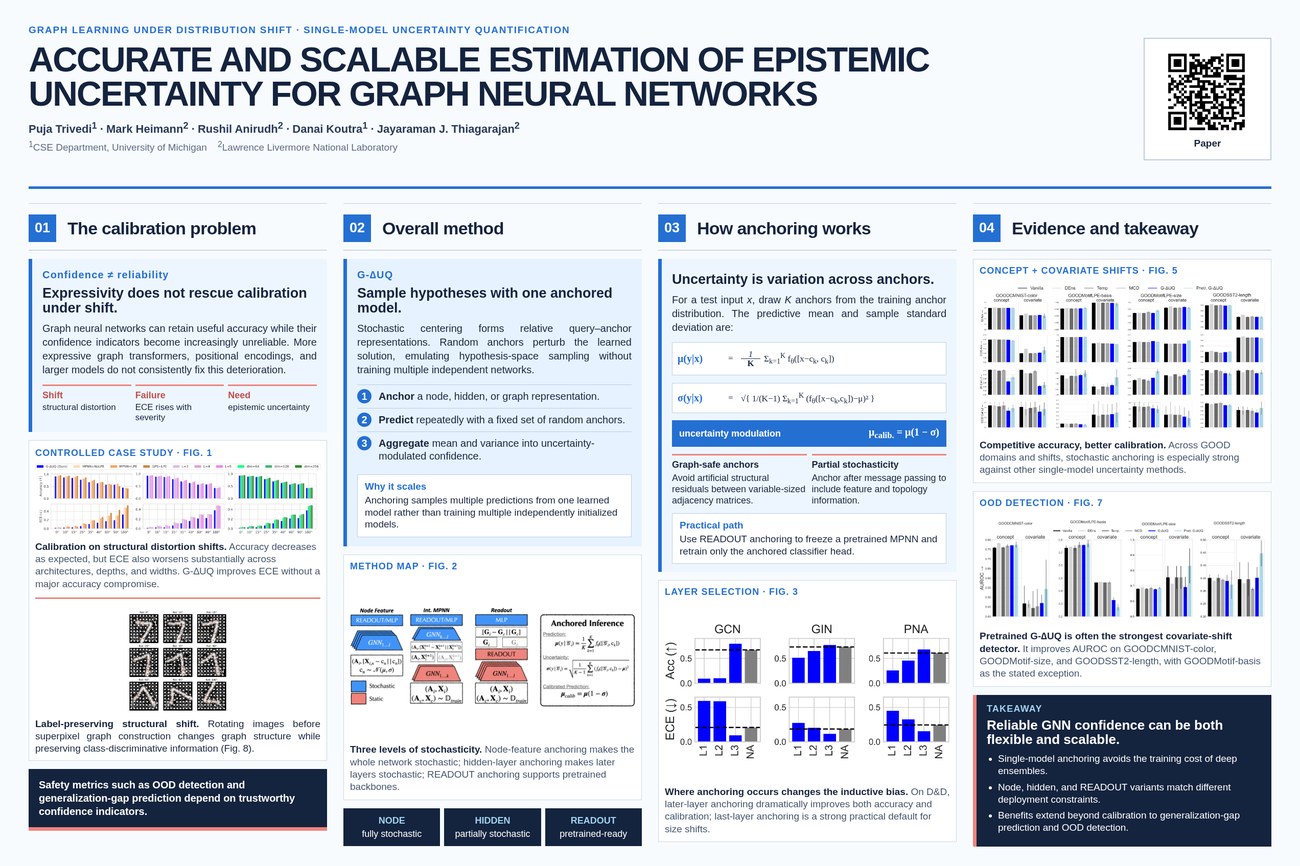}\\[1pt]
{\small (c) w/o anchored composition}
\end{minipage}\\[6pt]
\begin{minipage}[t]{0.49\linewidth}\centering
\includegraphics[width=\linewidth]{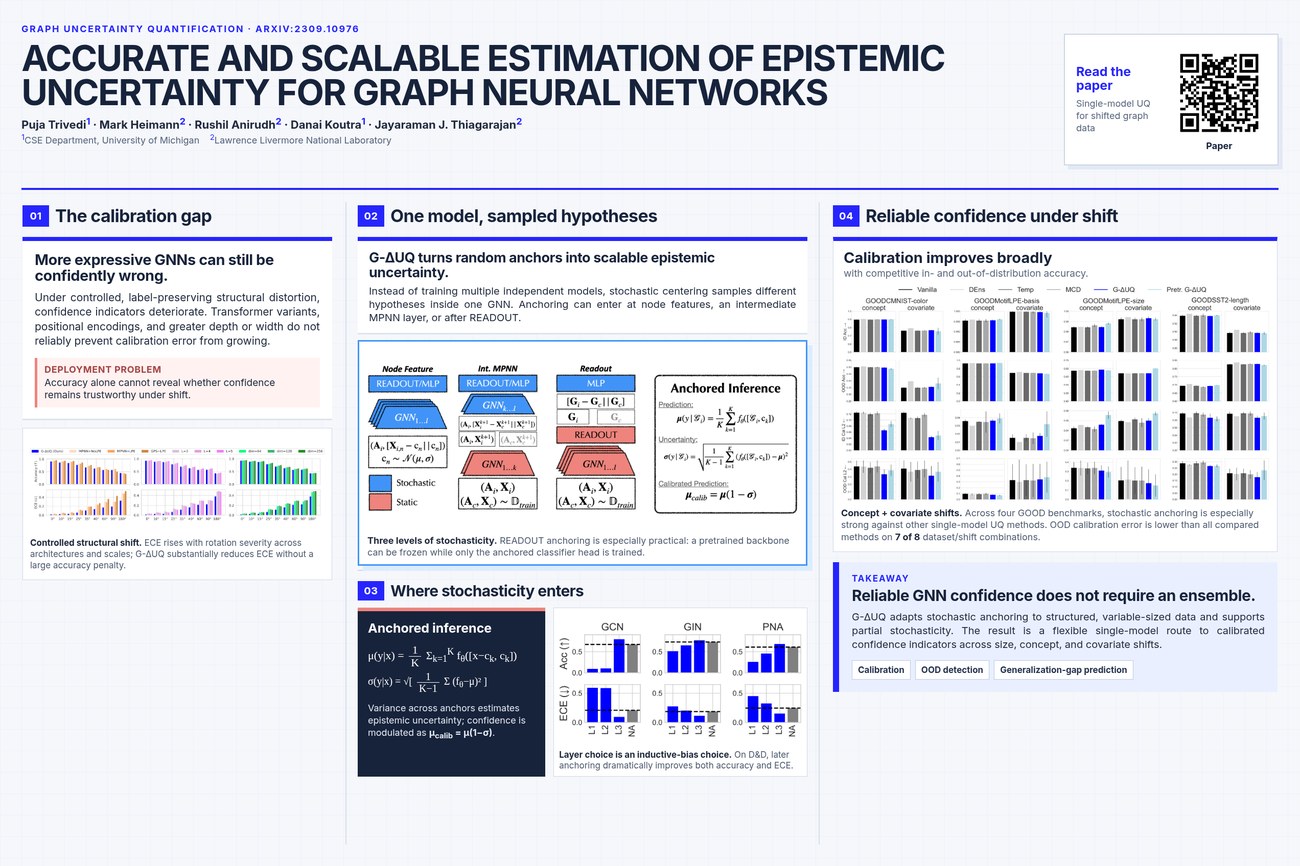}\\[1pt]
{\small (d) w/o semantic verification}
\end{minipage}\hfill
\begin{minipage}[t]{0.49\linewidth}\centering
\includegraphics[width=\linewidth]{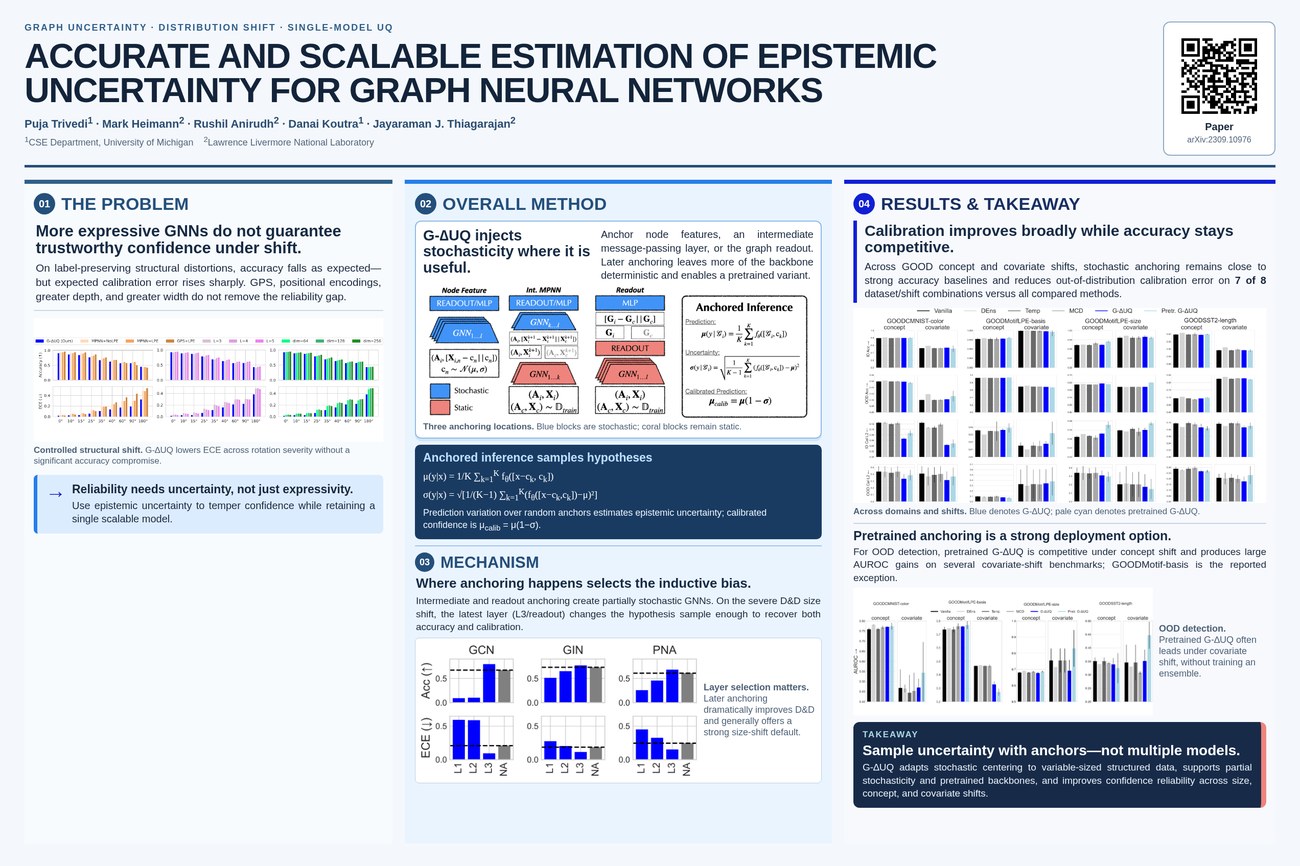}\\[1pt]
{\small (e) w/o refinement loop}
\end{minipage}
\end{minipage}
\caption{Posters generated for the same paper (\emph{Accurate and Scalable Estimation of Epistemic Uncertainty for Graph Neural Networks}) by VASP (GPT-5.6-Sol) and the four ablation variants of Table~\ref{tab:ablation}. The full framework displays the key figures at a larger scale than any variant.}
\label{fig:ablation-cases}
\end{figure}

Table~\ref{tab:ablation} removes one component of VASP at a time on a 20-paper subset, and Figure~\ref{fig:ablation-cases} shows the posters of all variants for one paper. Replacing SemParser with the assets that ResearchStudio extracts from the same papers lowers the visual-presentation score from 91.4 to 83.1, so intact evidence matters for composition. Within VisComposer, replacing anchored composition with a fixed layout template lowers the score to 80.8 and shrinks the largest figure from 7.3\% to 4.5\% of the canvas, which reproduces the layout mismatch of Section~\ref{sec:introduction}. Removing semantic verification lowers the score to 81.1, mostly through whitespace and density, because the agent must then judge the layout from rendered images, the execution mismatch of Section~\ref{sec:introduction}. Removing the refinement loop leaves only the first draft and lowers the score to 73.8. The anchored-composition variant tests Vision-Centric Anchoring, and the other three variants test Semantic-Geometric Decoupling.

\begin{table*}[h]
    \centering
    \caption{Ablations of VASP (GPT-5.6-Sol) on a 20-paper subset, removing one component at a time. Overall is the visual-presentation score. Fit, Text, Assoc., Space, and Density are the pass rates (\%) of the figure-size, in-figure-text, spatial-association, whitespace, and density criteria, and Largest is the area of the largest figure as a percentage of the canvas. The best result in each column is in bold and the second best is underlined.}
    \label{tab:ablation}
    \small
    \setlength{\tabcolsep}{4pt}
    \begin{tabular}{lccccccc}
        \hline
        & & \multicolumn{2}{c}{Figure (\%)} & \multicolumn{3}{c}{Layout (\%)} & Area (\%) \\
        \cmidrule(lr){3-4} \cmidrule(lr){5-7} \cmidrule(lr){8-8}
        Variant & Overall$\uparrow$ & Fit$\uparrow$ & Text$\uparrow$ & Assoc.$\uparrow$ & Space$\uparrow$ & Density$\uparrow$ & Largest$\uparrow$ \\
        \hline
        VASP (full)                             & \textbf{91.4} & \textbf{80} & \textbf{65} & \textbf{100} & \underline{80} & \textbf{90} & \textbf{7.3} \\
        \hline
        \multicolumn{8}{l}{\textit{SemParser}} \\
        \quad w/o SemParser                     & \underline{83.1} & \underline{70} & 55 & \underline{95} & 45 & \underline{70} & 6.7 \\
        \multicolumn{8}{l}{\textit{VisComposer}} \\
        \quad w/o anchored composition         & 80.8 & 30 & 30 & \textbf{100} & \textbf{95} & 65 & 4.5 \\
        \quad w/o semantic verification         & 81.1 & \underline{70} & \underline{60} & \textbf{100} & 10 & 65 & \underline{7.0} \\
        \quad w/o refinement loop               & 73.8 & 65 & 45 & 75 & 10 & 50 & 6.9 \\
        \hline
    \end{tabular}
\end{table*}

\paragraph{Decoupled ablations of composition priority and verification.}
To separate planning priority from layout flexibility and geometric verification from semantic verification, Table~\ref{tab:ablation-decoupled} evaluates decoupled controls on the same 20-paper subset. First, we include a \emph{text-first} variant that retains the same extracted assets, template-free CSS grid, and two-round refinement budget as VASP (full), changing only the planning order so that section prose is budgeted before figures are placed into the remaining space. Second, because \texttt{Assoc.} (\texttt{AL05}) measures local label proximity rather than whether body prose explains a figure---and the \emph{w/o semantic verification} row in Table~\ref{tab:ablation} disabled the entire revision controller (both iterative semantic checks and geometric/whitespace checks while retaining round-0 semantic planning)---we independently toggle Initial SV, Iterative SV, and geometric/whitespace verification (GV), and additionally evaluate \textbf{Claim--Evidence Grounding (CEG)}, a blind VLM check testing whether each displayed figure or table is substantively explained by its section's body text.

\begin{table*}[ht]
    \centering
    \caption{Decoupled ablations of VASP (GPT-5.6-Sol) on the 20-paper subset, isolating composition priority (visual-first vs.\ template vs.\ text-first on an identical flexible grid) and verification components (Initial SV, Iterative SV, and geometric/whitespace verification GV). CEG measures blind Claim--Evidence Grounding (\%) and Coverage reports author-informed key-information coverage (\%). $^\dagger$Corresponds to the \emph{w/o semantic verification} row of Table~\ref{tab:ablation}, which retained initial SV planning while disabling iterative SV and GV during revisions.}
    \label{tab:ablation-decoupled}
    \vspace{4pt}
    \small
    \renewcommand{\arraystretch}{1.08}
    \setlength{\tabcolsep}{3pt}
    \resizebox{\textwidth}{!}{%
    \begin{tabular}{lccccccccc}
        \hline
        & & \multicolumn{2}{c}{Figure (\%)} & \multicolumn{3}{c}{Layout (\%)} & Area (\%) & \multicolumn{2}{c}{Semantics (\%)} \\
        \cmidrule(lr){3-4} \cmidrule(lr){5-7} \cmidrule(lr){8-8} \cmidrule(lr){9-10}
        Variant & Overall$\uparrow$ & Fit$\uparrow$ & Text$\uparrow$ & Assoc.$\uparrow$ & Space$\uparrow$ & Density$\uparrow$ & Largest$\uparrow$ & CEG$\uparrow$ & Coverage$\uparrow$ \\
        \hline
        \multicolumn{10}{l}{\textit{Composition priority (with full SV and GV enabled)}} \\
        \textbf{VASP (full: visual-first)}      & \textbf{91.4}    & \textbf{80}    & \textbf{65}    & \textbf{100} & \underline{80} & \textbf{90}    & \underline{7.3} & 82.3             & 75.7             \\
        w/o anchored comp.\ (template)          & 80.8             & 30             & 30             & \textbf{100} & \textbf{95}    & 65             & 4.5             & 75.8             & \underline{81.4} \\
        w/o anchored comp.\ (text-first)        & 72.8             & 25             & 15             & \textbf{100} & 75             & 30             & 2.7             & \underline{87.0} & \textbf{89.3}    \\
        \hline
        \multicolumn{10}{l}{\textit{Verification decoupling (with visual-first composition fixed)}} \\
        w/o iterative SV (init.\ SV + GV)       & \underline{90.6} & \textbf{80}    & \textbf{65}    & \textbf{100} & 75             & \underline{85} & 6.9             & 78.2             & 77.1             \\
        w/o all SV (GV only)                    & 89.0             & 60             & \underline{60} & \textbf{100} & \textbf{95}    & \textbf{90}    & \textbf{8.9}    & 25.0             & 67.9             \\
        w/o GV (init.\ + iter.\ SV)             & 80.1             & 65             & 50             & 95           & 15             & 65             & 7.0             & \textbf{88.1}    & 80.0             \\
        w/o iter.\ SV \& GV (Table~\ref{tab:ablation})$^\dagger$ & 81.1 & \underline{70} & \underline{60} & \textbf{100} & 10 & 65 & 7.0 & 72.1 & 77.1 \\
        \hline
    \end{tabular}}
\end{table*}

Each factor separates cleanly in Table~\ref{tab:ablation-decoupled}. \textbf{Visual-first priority governs figure scale and layout balance}: with assets, flexible grid, and refinement budget held fixed, switching from visual-first to text-first drops Overall from 91.4 to 72.8, shrinking \texttt{Largest} figure area from 7.3\% to 2.7\%, \texttt{Fit} from 80\% to 25\%, in-figure \texttt{Text} legibility from 65\% to 15\%, and \texttt{Density} from 90\% to 30\%. \textbf{GV governs whitespace}: disabling GV drops \texttt{Space} to 15\% and 10\%, whereas keeping GV holds \texttt{Space} at 75\%--95\%, explaining the whitespace drop in Table~\ref{tab:ablation}. \textbf{Initial SV establishes claim--evidence grounding}: with GV fixed, removing initial SV collapses CEG from 78.2\% to 25.0\% and Coverage from 77.1\% to 67.9\%, as columns fill with generic bullets unrelated to the displayed figures. \textbf{Iterative SV preserves grounding across revisions}: with GV enabled, removing iterative SV allows geometry-only edits to trim figure-explaining prose (CEG drops from 82.3\% to 78.2\%), whereas without GV, adding iterative SV lifts CEG from 72.1\% to 88.1\%.

\subsection{Reasoning Effort and Revision Rounds}
\label{app:effort-rounds}

\begin{figure}[p]
\centering
\small
\setlength{\tabcolsep}{4pt}
\begin{tabular}{cc}
Reasoning effort ($K=2$) & Revision bound $K$ (medium effort) \\[2pt]
\includegraphics[width=0.38\linewidth]{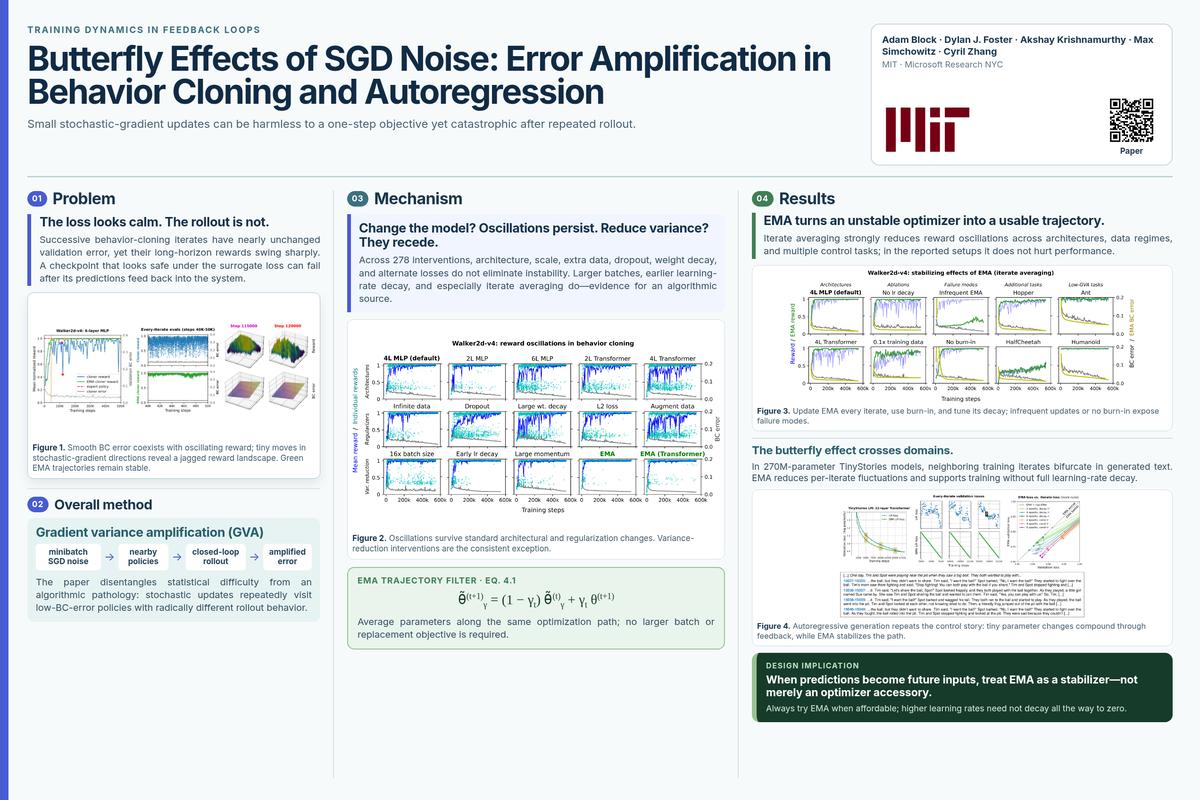} &
\includegraphics[width=0.38\linewidth]{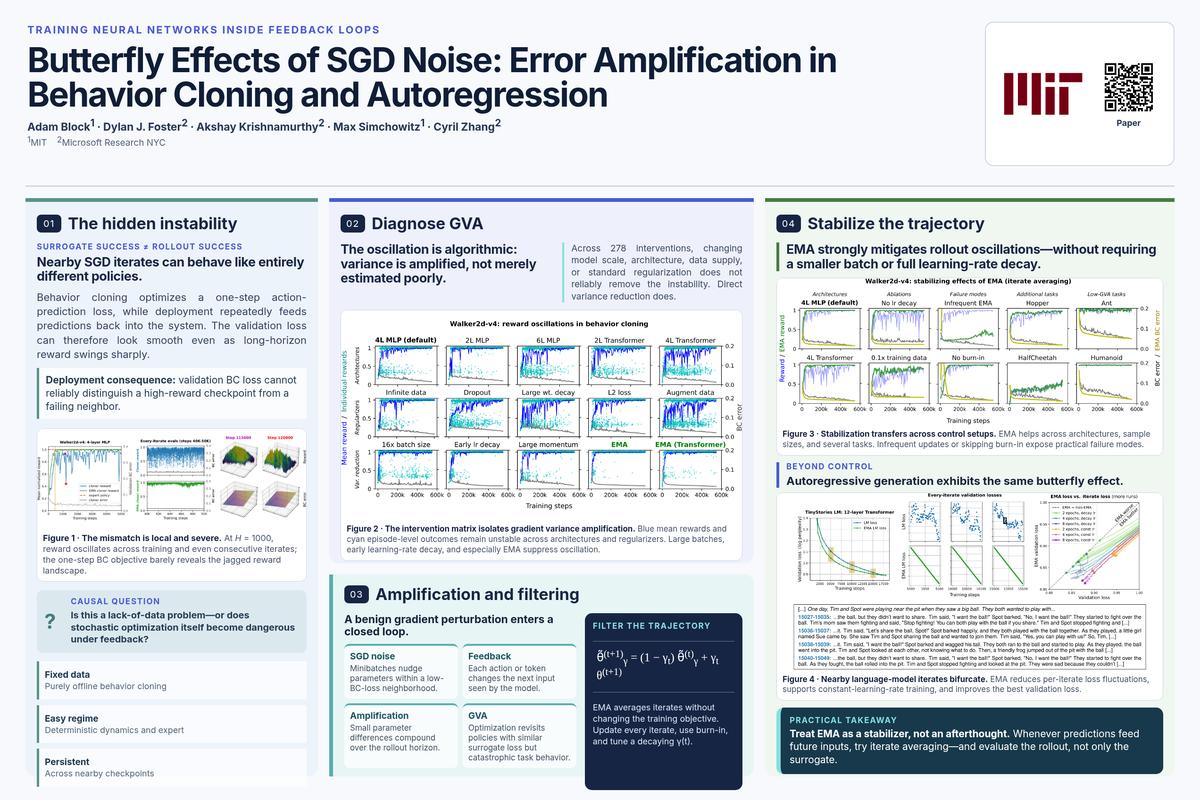} \\
low & $K=1$ \\[3pt]
\includegraphics[width=0.38\linewidth]{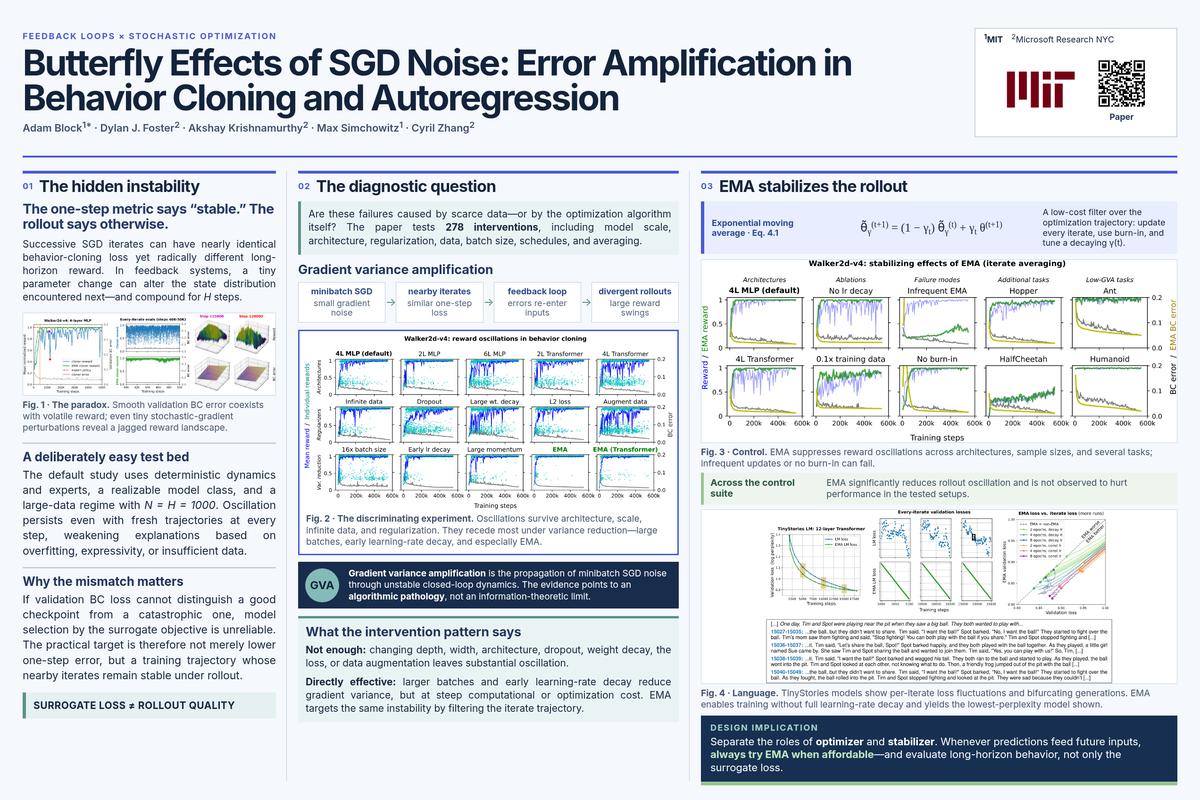} &
\includegraphics[width=0.38\linewidth]{figures/effort_rounds_medium_k2.jpg} \\
medium (default) & $K=2$ (default) \\[3pt]
\includegraphics[width=0.38\linewidth]{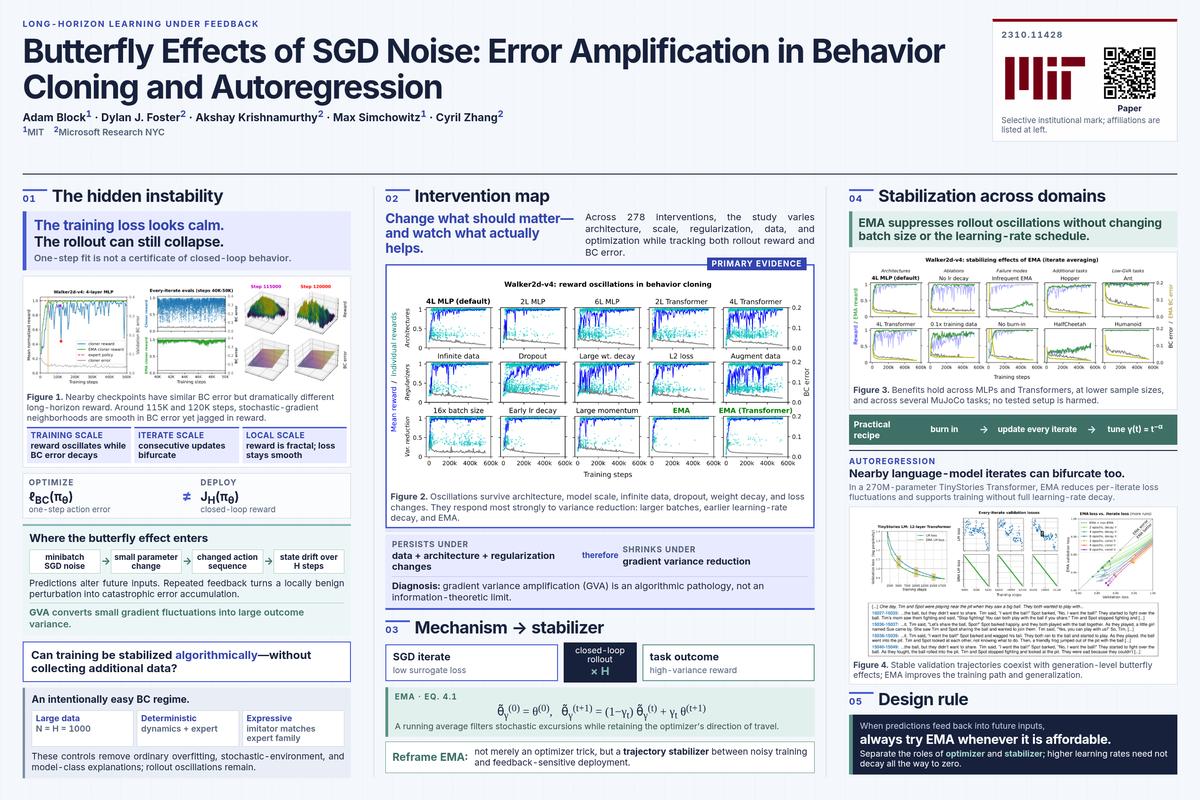} &
\includegraphics[width=0.38\linewidth]{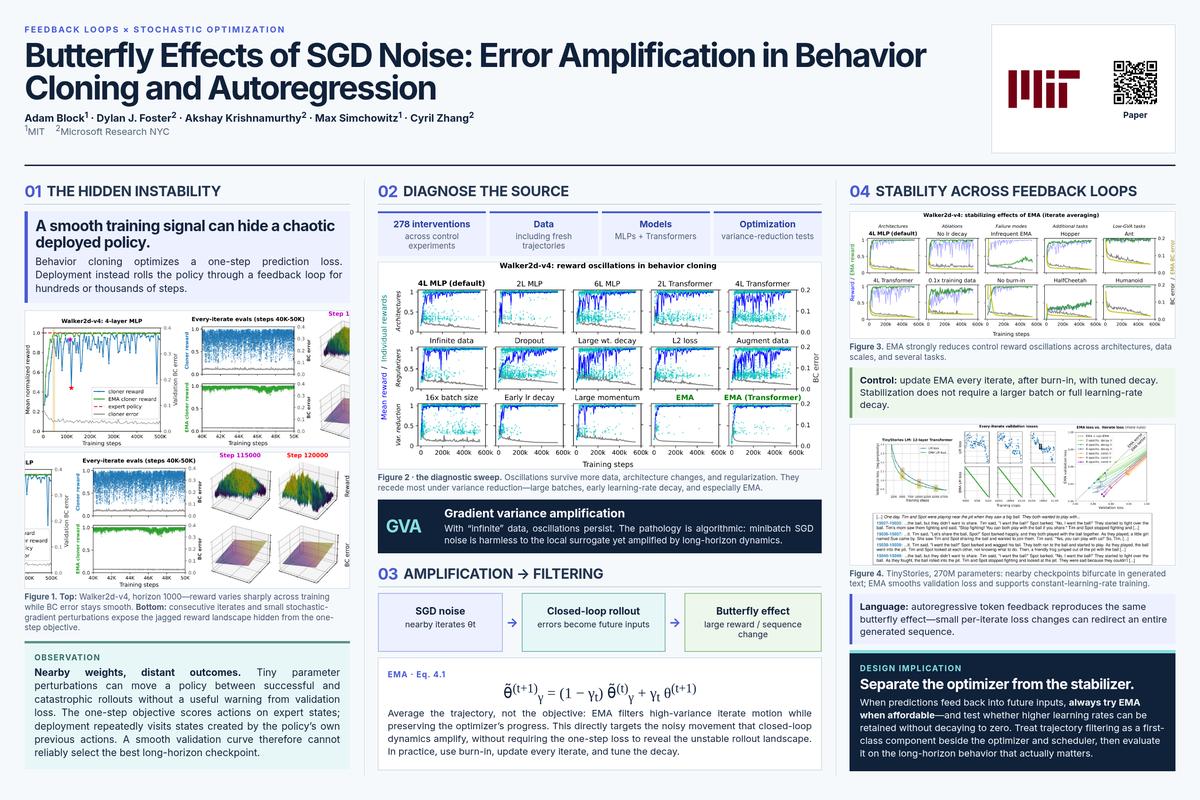} \\
high & $K=4$ \\[3pt]
\includegraphics[width=0.38\linewidth]{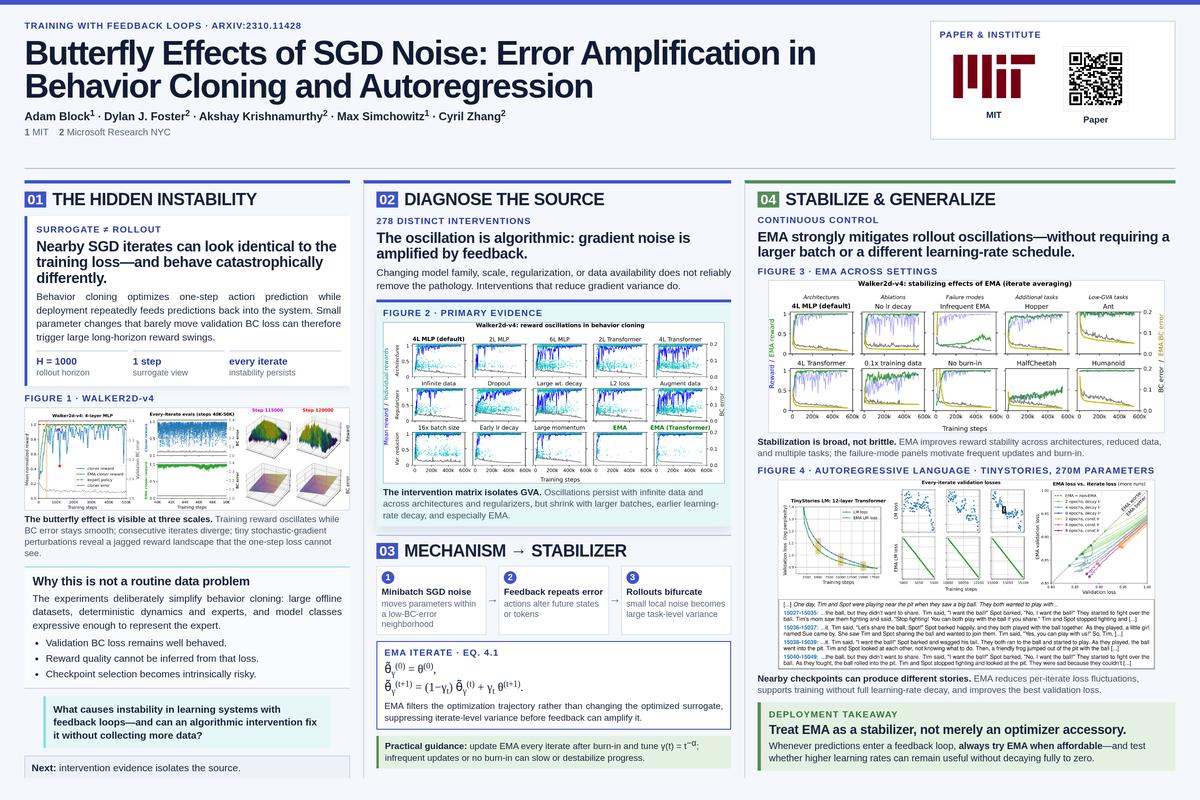} &
\includegraphics[width=0.38\linewidth]{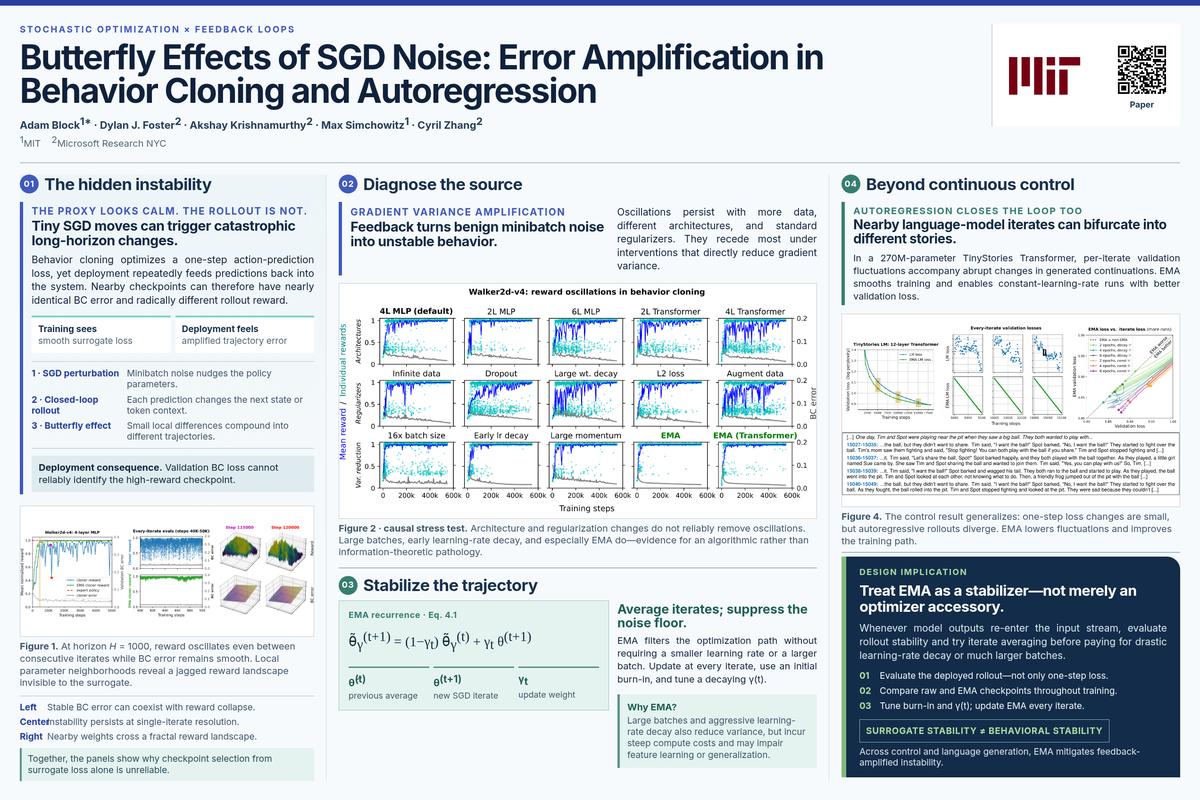} \\
xhigh & $K=8$ \\[3pt]
\includegraphics[width=0.38\linewidth]{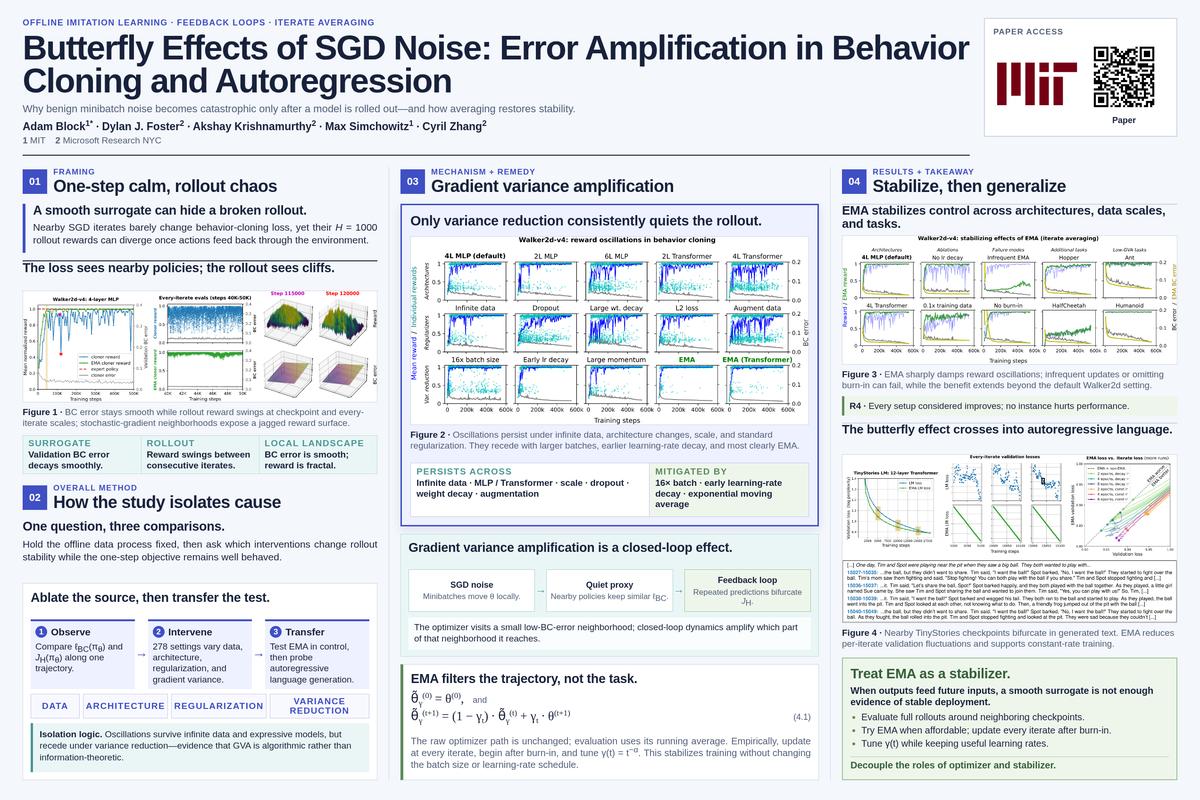} &
\includegraphics[width=0.38\linewidth]{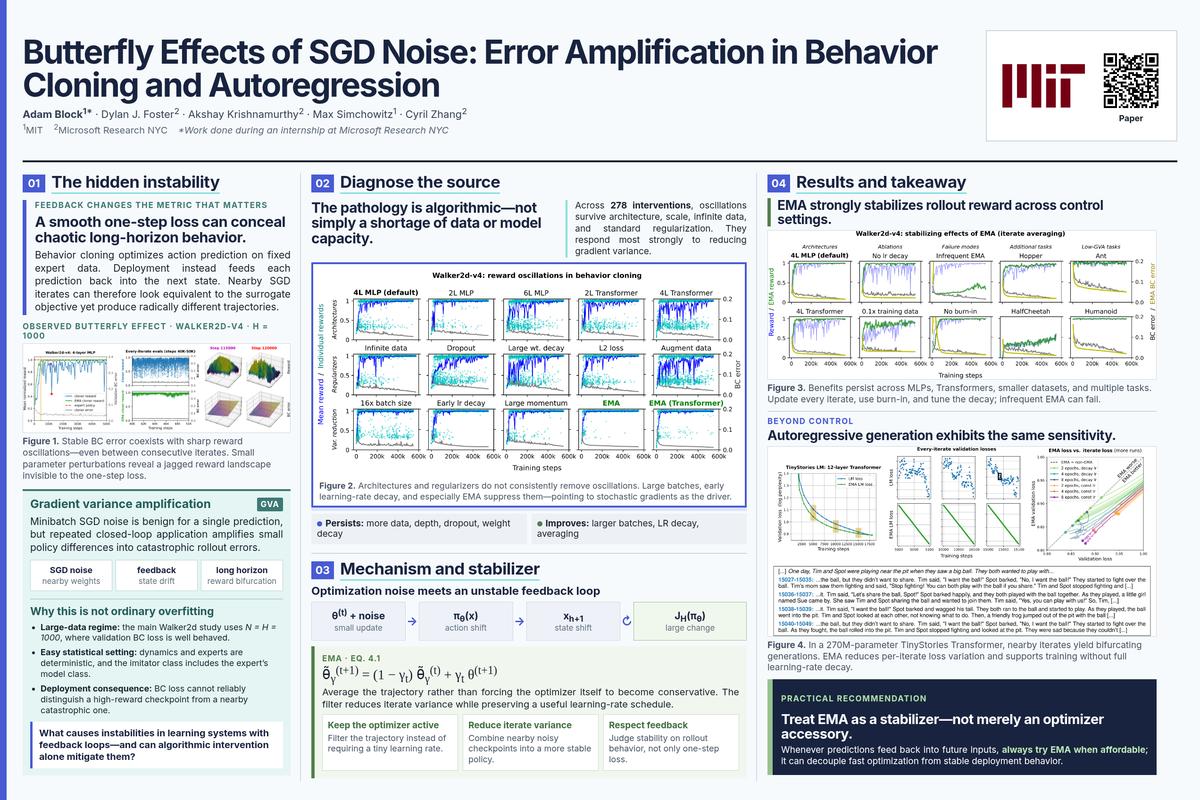} \\
max & $K=16$ \\
\end{tabular}
\caption{Posters generated by VASP (GPT-5.6-Sol) for \emph{Butterfly Effects of SGD Noise: Error Amplification in Behavior Cloning and Autoregression} when one setting changes at a time. Left: reasoning effort of the design agent with $K=2$. Right: bound $K$ on revision rounds with medium effort. The second row shows the default configuration in both columns.}
\label{fig:effort-rounds}
\end{figure}

Figure~\ref{fig:effort-rounds} shows the posters that VASP (GPT-5.6-Sol) generates for one paper, \emph{Butterfly Effects of SGD Noise: Error Amplification in Behavior Cloning and Autoregression}, when one setting changes at a time from the default of medium reasoning effort and at most $K=2$ revisions. The left column varies the reasoning effort of the design agent with $K=2$, and the right column varies the revision bound $K$ with medium effort. A checkpoint was accepted with high and max effort after two revisions, and with $K=4$, $8$, and $16$ after four, six, and five revisions, so the larger bounds were not exhausted. With low, medium, and xhigh effort at $K=2$ and with $K=1$, no checkpoint was accepted within the bound, and the best-ranked checkpoint was returned, which for low effort is the first draft. The comparison is qualitative and covers a single paper.

\clearpage

\subsection{Backbone Comparison}
\label{app:backbone-comparison}

Table~\ref{tab:backbone-comparison} compares VASP driven by three backbone LLMs, GPT-5.6-Sol, Opus-5, and GPT-5.6-Luna, with ResearchStudio, the strongest existing system, on the VLM-as-Judge component of the Paper2Poster benchmark, together with the API cost and time per poster. VASP scores above ResearchStudio overall with all three backbones (3.71, 3.70, and 3.62 versus 3.52) while costing less and finishing faster. GPT-5.6-Sol gives the highest aesthetic and overall scores in the shortest time. Opus-5 gives the highest information score (3.83) at 4.2 times the cost of GPT-5.6-Sol. GPT-5.6-Luna is the cheapest backbone at 0.55 USD per poster, about 21 times cheaper than ResearchStudio. Its information score equals that of GPT-5.6-Sol (3.69), and its lower aesthetic score (3.55 versus 3.72) comes mainly from engagement, one of the three aesthetic sub-scores (3.31 versus 3.63). Compared paper by paper, VASP (GPT-5.6-Luna) scores above ResearchStudio overall on 55 papers and below it on 17 (Wilcoxon $p<10^{-4}$). For the asset stage alone, SemParser takes 3.6 min and 0.41 USD per paper, whereas ResearchStudio takes 11.5--12.6 min and 3.93--4.33 USD, with the range reflecting two ways of separating its asset and poster stages in its logs; SemParser is thus 3.2--3.5$\times$ faster and 9.6--10.6$\times$ cheaper.

\begin{figure}[h]
\centering
\begin{minipage}[t]{0.49\textwidth}\centering
\includegraphics[width=\linewidth]{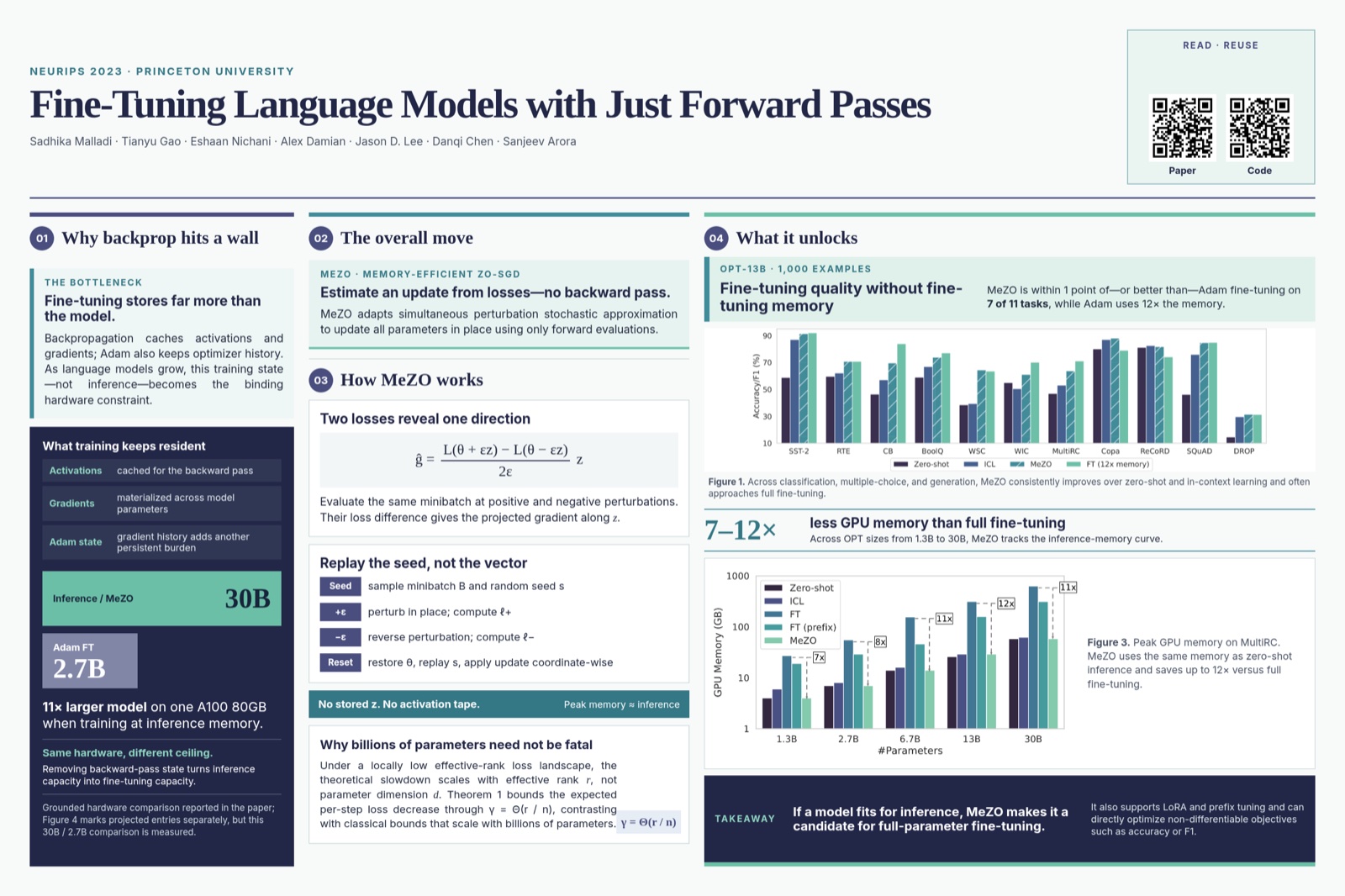}\\[2pt]
{\small (a) VASP (GPT-5.6-Sol)}
\end{minipage}\hfill
\begin{minipage}[t]{0.49\textwidth}\centering
\includegraphics[width=\linewidth]{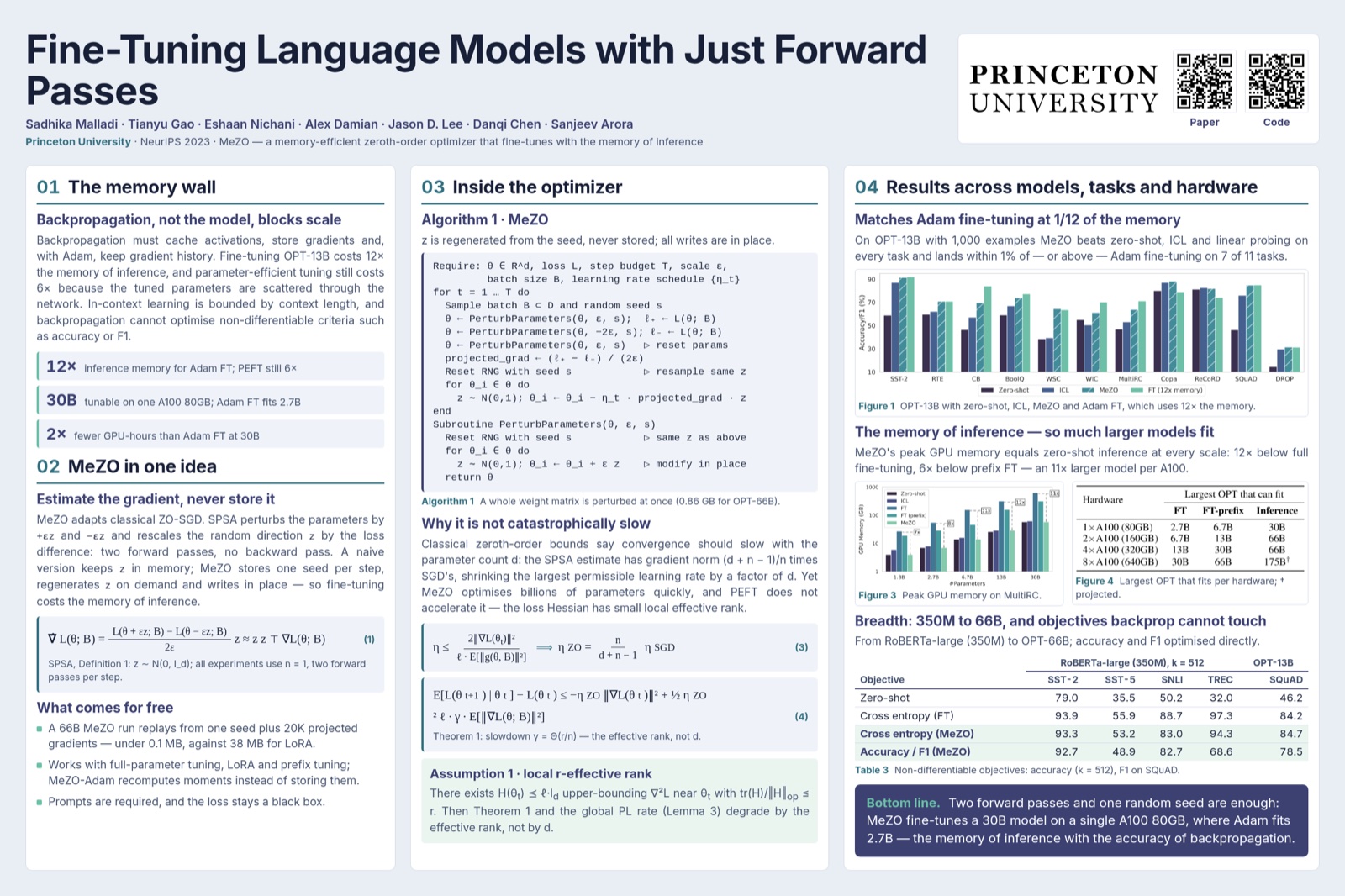}\\[2pt]
{\small (b) VASP (Opus-5)}
\end{minipage}\\[6pt]
\begin{minipage}[t]{0.49\textwidth}\centering
\includegraphics[width=\linewidth]{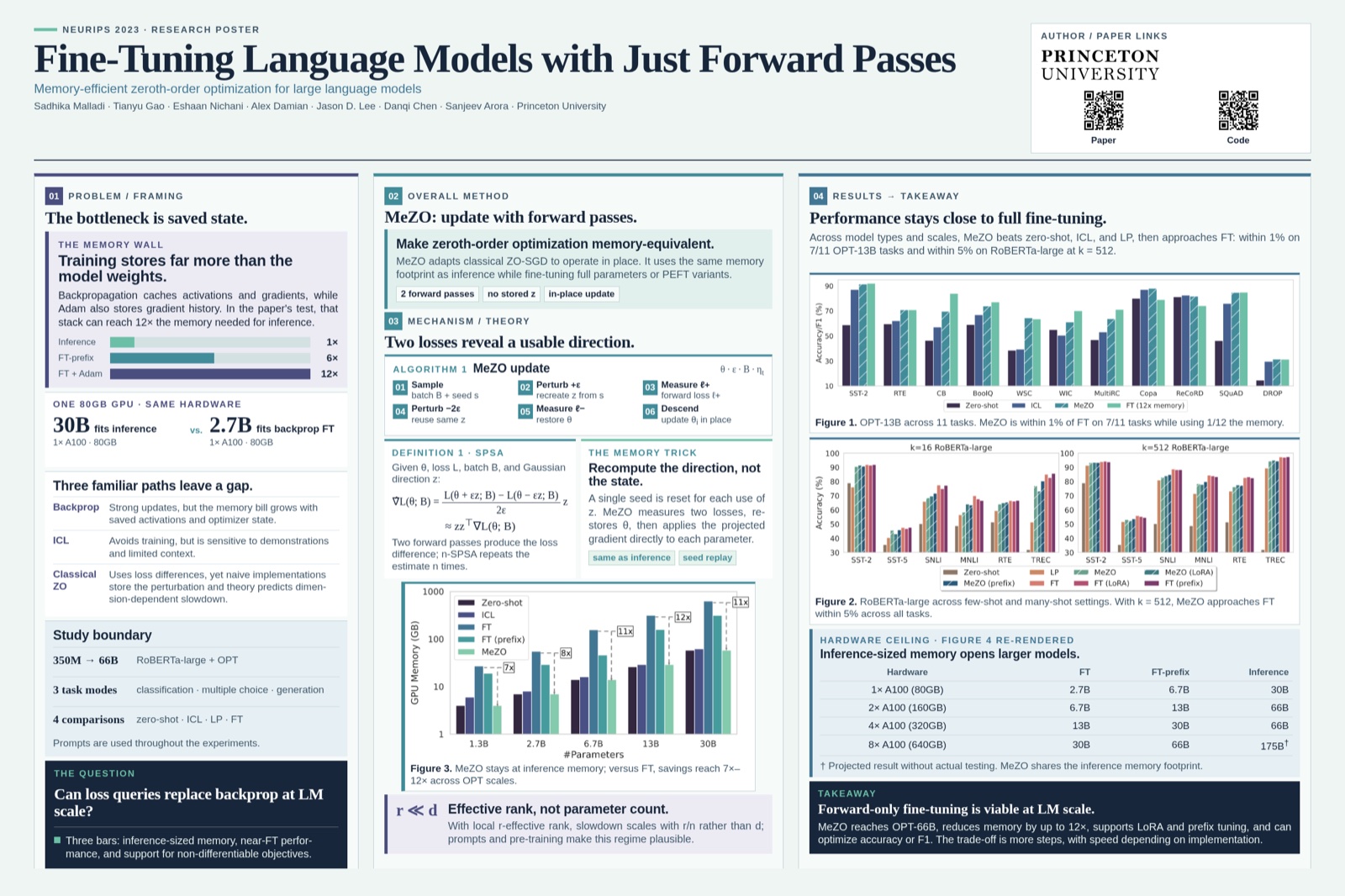}\\[2pt]
{\small (c) VASP (GPT-5.6-Luna)}
\end{minipage}\hfill
\begin{minipage}[t]{0.49\textwidth}\centering
\includegraphics[width=\linewidth]{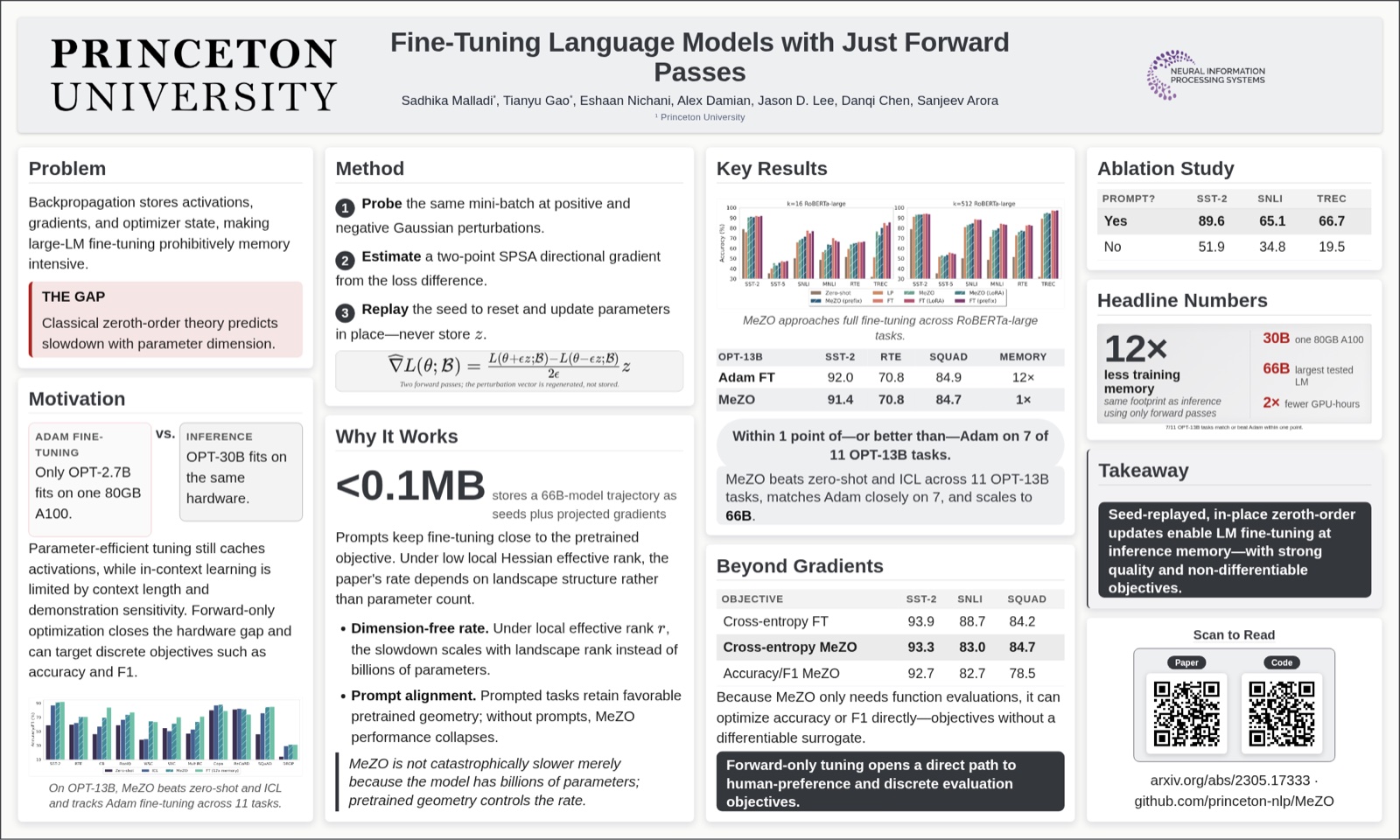}\\[2pt]
{\small (d) ResearchStudio (GPT-5.6-Sol)}
\end{minipage}
\caption{Posters generated for \emph{Fine-Tuning Language Models with Just Forward Passes} by VASP with three backbone LLMs and by ResearchStudio. The VLM-as-Judge gives each VASP poster an overall score of 3.83 and the ResearchStudio poster 3.33.}
\label{fig:backbone-cases}
\end{figure}

Figure~\ref{fig:backbone-cases} shows the four posters for one paper. All three VASP posters display the paper's main accuracy chart over eleven OPT-13B tasks at a scale at which its axes and legend remain legible, whereas ResearchStudio shrinks this chart into a corner of its motivation panel and fills most of the canvas with text and tables.

\begin{table}[h]
    \centering
    \caption{VLM-as-Judge scores of VASP with three backbone LLMs and of ResearchStudio, together with the API cost (USD) and time (min) per poster. Aesthetic and Information are the means of their three sub-scores, and Overall is the mean of all six. Higher is better for the scores and lower is better for cost and time. $^\dagger$VASP (GPT-5.6-Luna) uses the assets parsed by GPT-5.6-Sol, so its cost and time include the GPT-5.6-Sol asset-parsing stage.}
    \label{tab:backbone-comparison}
    \small
    \setlength{\tabcolsep}{5pt}
    \begin{tabular}{lccccc}
        \hline
        & \multicolumn{3}{c}{VLM-as-Judge} & \multicolumn{2}{c}{Efficiency} \\
        \cmidrule(lr){2-4} \cmidrule(lr){5-6}
        Method & Aesthetic$\uparrow$ & Information$\uparrow$ & Overall$\uparrow$ & Cost$\downarrow$ & Time$\downarrow$ \\
        \hline
        \multicolumn{6}{l}{\textit{Ours}} \\
        \textbf{VASP (GPT-5.6-Sol)}    & \textbf{3.72} & \underline{3.69} & \textbf{3.71} & \underline{1.64} & \textbf{11.4} \\
        \textbf{VASP (Opus-5)}         & \underline{3.58} & \textbf{3.83} & \underline{3.70} & 6.90 & \underline{27.9} \\
        \textbf{VASP (GPT-5.6-Luna)}   & 3.55 & \underline{3.69} & 3.62 & \textbf{0.55}$^\dagger$ & 31.1$^\dagger$ \\
        \hline
        \multicolumn{6}{l}{\textit{Baseline method}} \\
        ResearchStudio (GPT-5.6-Sol)   & 3.44 & 3.59 & 3.52 & 11.70 & 32.7 \\
        \hline
    \end{tabular}
\end{table}

\paragraph{Same-backbone comparisons under Claude Opus-5 and Qwen3.8-Flash-Next.}
We also compare VASP with three representative baselines---ResearchStudio (strongest visual baseline), PosterGen (strongest coverage baseline), and Paper2Poster---when all methods share the same alternative backbones (Claude Opus-5 and Qwen3.8-Flash-Next) across all 100 papers under the same evaluation protocol (Table~\ref{tab:cross-backbone-100}).

\begin{table*}[ht]
    \centering
    \caption{Same-backbone comparison across all 100 papers under Claude Opus-5 and Qwen3.8-Flash-Next. Higher is better. Best results within each backbone group are in bold and second best are underlined.}
    \label{tab:cross-backbone-100}
    \vspace{4pt}
    \small
    \setlength{\tabcolsep}{7pt}
    \begin{tabular}{lcccc}
        \hline
        Method & \makecell{VLM-as-Judge\\(Table~\ref{tab:paper2poster-vlm-efficiency})$\uparrow$} & \makecell{PaperQuiz\\(\%)$\uparrow$} & \makecell{Visual Presentation\\(\%)$\uparrow$} & \makecell{Key-Information\\Coverage (\%)$\uparrow$} \\
        \hline
        \multicolumn{5}{l}{\textit{Claude Opus-5 backbone}} \\
        \textbf{VASP}                    & \textbf{3.70}    & \textbf{78.08}    & \textbf{92.5}    & \textbf{88.7}    \\
        ResearchStudio                   & \underline{3.58} & 71.77             & \underline{88.7} & 77.9             \\
        PosterGen                        & 3.00             & \underline{76.05} & 65.9             & \underline{86.6} \\
        Paper2Poster                     & 2.76             & 74.55             & 44.0             & 78.0             \\
        \hline
        \multicolumn{5}{l}{\textit{Qwen3.8-Flash-Next backbone}} \\
        \textbf{VASP}                    & \textbf{3.58}    & \textbf{81.35}    & \textbf{89.2}    & \textbf{89.3}    \\
        ResearchStudio                   & \underline{3.54} & \underline{73.68} & \textbf{89.2}    & 80.4             \\
        PosterGen                        & 3.04             & 73.28             & 65.7             & \underline{82.6} \\
        Paper2Poster                     & 2.42             & 73.16             & 46.7             & 74.7             \\
        \hline
        \multicolumn{5}{l}{\textit{Human reference}} \\
        Official posters                 & 3.12             & 67.55             & 83.0             & 96.4             \\
        \hline
    \end{tabular}
\end{table*}

Under both backbones, VASP ranks first (or tied first) across all four benchmarks. ResearchStudio approaches or matches VASP on visual presentation (88.7\% and 89.2\%) with lower content retention, trailing VASP by 6.31 to 7.67 points on PaperQuiz and by 8.9 to 10.8 points on key-information coverage. Stronger backbones raise text recall and coverage in PosterGen and Paper2Poster, while their visual presentation remains at 65.7\%--65.9\% and 44.0\%--46.7\% due to canvas crowding.

\clearpage
\subsection{Comparison with General-Purpose Multi-Format Frameworks}
\label{app:paperx-comparison}

Both ResearchStudio \citep{xiao2026researchstudio} and PaperX \citep{yu2026paperx} are general-purpose presentation frameworks: ResearchStudio produces a poster, video deck, and blog from one asset bundle, and PaperX produces slides, posters, and promotional posts. Table~\ref{tab:paperx-comparison} compares VASP with both frameworks on the posters generated for all 100 papers under GPT-5.6-Sol, scored by GPT-5.6-Sol under the same evaluation protocols.

\begin{table*}[!ht]
    \centering
    \caption{Comparison of VASP with the general-purpose multi-format frameworks ResearchStudio and PaperX across all 100 papers (all generated and scored with GPT-5.6-Sol). Higher is better.}
    \label{tab:paperx-comparison}
    \vspace{4pt}
    \small
    \setlength{\tabcolsep}{5pt}
    \begin{tabular}{lcccccc}
        \hline
        & \multicolumn{3}{c}{VLM-as-Judge} & \multicolumn{3}{c}{PaperQuiz (\%)} \\
        \cmidrule(lr){2-4}\cmidrule(lr){5-7}
        Method & Aesthetic$\uparrow$ & Information$\uparrow$ & Overall$\uparrow$ & Verbatim$\uparrow$ & Interpretive$\uparrow$ & Overall$\uparrow$ \\
        \hline
        \textbf{VASP}  & \textbf{3.72} & \textbf{3.69} & \textbf{3.71} & \textbf{52.62} & \textbf{87.30} & \textbf{69.96} \\
        ResearchStudio & 3.44          & 3.59          & 3.52          & 49.68          & 85.28          & 67.48 \\
        PaperX         & 2.35          & 3.14          & 2.74          & 31.94          & 77.56          & 54.75 \\
        \hline
    \end{tabular}

    \vspace{6pt}

    \begin{tabular}{lcccccc}
        \hline
        & \multicolumn{4}{c}{Visual Presentation (\%)} & \multicolumn{2}{c}{Key-Info.\ Coverage (\%)} \\
        \cmidrule(lr){2-5}\cmidrule(lr){6-7}
        Method & Figure Pres.$\uparrow$ & Visual Guid.$\uparrow$ & Layout$\uparrow$ & Overall$\uparrow$ & Overall$\uparrow$ & Full$\uparrow$ \\
        \hline
        \textbf{VASP}  & \textbf{88.0} & \textbf{100.0} & \textbf{95.0} & \textbf{94.3} & \textbf{82.0} & \textbf{20} \\
        ResearchStudio & 73.7          & 97.0           & 93.5          & 88.1          & 73.0          & 18 \\
        PaperX         & 18.3          & 60.2           & 51.0          & 43.2          & 60.9          & 7 \\
        \hline
    \end{tabular}
\end{table*}

Across all 100 papers, VASP outperforms both ResearchStudio and PaperX on every reported metric: VLM-as-Judge overall (3.71 vs.\ 3.52 and 2.74), PaperQuiz overall (69.96\% vs.\ 67.48\% and 54.75\%), visual presentation (94.3\% vs.\ 88.1\% and 43.2\%), and key-information coverage (82.0\% vs.\ 73.0\% and 60.9\%, with 20 vs.\ 18 and 7 full-coverage posters). Because SemParser constructs a layout-agnostic asset package once per paper (Section~\ref{sec:asset-extraction}), the same assets can be reused across output formats such as slides, short videos, and promotional summaries.

\subsection{Cost and Time per Poster}
\label{app:cost-time}

Table~\ref{tab:cost-time} lists the API cost and time per poster behind Figure~\ref{fig:teaser}. All systems were run one job at a time on the same machine, and API cost is priced per request with the same token prices for every system. No request exceeded the long-context threshold, so no long-context surcharge applies. For VASP, cost and time are the sums of the asset and poster stages. For Paper2Poster, P2P, and PosterMELD, whose document parsing runs as a separate local step without API cost, the time of that step is added: the pre-parsing of Paper2Poster, the DocLayout-YOLO detection of P2P, and the MinerU parsing of PosterMELD.

\begin{table}[h]
\centering
\caption{API cost and time per poster for the systems in Figure~\ref{fig:teaser}. Lower is better, and the best result is in bold and the second best is underlined.}
\label{tab:cost-time}
\small
\setlength{\tabcolsep}{10pt}
\begin{tabular}{lcc}
    \hline
    Method & Cost (USD)$\downarrow$ & Time (min)$\downarrow$ \\
    \hline
    \multicolumn{3}{l}{\textit{Ours}} \\
    \textbf{VASP (GPT-5.6-Sol)} & 1.64 & \underline{11.4} \\
    \hline
    \multicolumn{3}{l}{\textit{Baseline methods}} \\
    Paper2Poster       & 2.72 & 14.5 \\
    PosterMELD         & 2.35 & 16.3 \\
    PosterForest       & 1.75 & 23.9 \\
    PosterGen          & \textbf{1.03} & \textbf{9.3} \\
    P2P                & \underline{1.57} & 13.6 \\
    ResearchStudio     & 11.70 & 32.7 \\
    \hline
\end{tabular}
\end{table}

\clearpage

\section{Additional Evaluations and Analyses}
\label{app:validation-robustness}

This appendix reports cross-family VLM evaluation, blind human evaluation, visual-theme sensitivity and criterion-subset analyses, \texttt{N/A}-handling sensitivity, and paper-paired bootstrap confidence intervals across the 100-paper evaluation set.

\subsection{Cross-Family VLM Judge Evaluation (Gemini-4-Argon)}
\label{app:cross-family-judge}
\label{app:judge-human}

We also evaluate all 900 posters across all 100 papers on all three poster-level benchmarks---Paper2Poster VLM-as-Judge and PaperQuiz (Table~\ref{tab:argon-table1}), Visual Presentation (Table~\ref{tab:argon-table3}), and Author-Informed Key-Information Coverage (Table~\ref{tab:argon-table4})---using Gemini-4-Argon, a VLM from an unrelated model family that is not used by VASP or any baseline. Prompts, schemas, and scoring rules are unchanged.

\begin{table*}[ht]
    \centering
    \caption{Paper2Poster VLM-as-Judge and PaperQuiz (\%) results across all 100 papers re-evaluated with the cross-family judge Gemini-4-Argon. Higher is better. $^\dagger$P2P renders a fixed-width page of unbounded height and is not ranked on PaperQuiz.}
    \label{tab:argon-table1}
    \vspace{4pt}
    \small
    \setlength{\tabcolsep}{5pt}
    \begin{tabular}{lcccccc}
        \hline
        & \multicolumn{3}{c}{VLM-as-Judge} & \multicolumn{3}{c}{PaperQuiz (\%)} \\
        \cmidrule(lr){2-4} \cmidrule(lr){5-7}
        Method & Aesthetic$\uparrow$ & Information$\uparrow$ & Overall$\uparrow$ & Verbatim$\uparrow$ & Interpretive$\uparrow$ & Overall$\uparrow$ \\
        \hline
        \multicolumn{7}{l}{\textit{Ours}} \\
        \textbf{VASP (Opus-5)}      & \underline{3.62} & \textbf{3.83}    & \textbf{3.72}    & \textbf{60.46}    & \textbf{86.70}    & \textbf{73.58}    \\
        \textbf{VASP (GPT-5.6-Sol)} & \textbf{3.72}    & 3.62             & \underline{3.67} & \underline{45.68} & \underline{84.68} & \underline{65.18} \\
        \hline
        \multicolumn{7}{l}{\textit{Baseline methods}} \\
        Paper2Poster                & 1.72             & 2.81             & 2.27             & 43.96             & 82.28             & 63.12             \\
        PosterMELD                  & 1.82             & 2.69             & 2.26             & 34.72             & 77.74             & 56.23             \\
        PosterForest                & 2.06             & 2.83             & 2.44             & 41.46             & 81.84             & 61.65             \\
        PosterGen                   & 2.34             & 3.30             & 2.82             & 43.76             & 80.56             & 62.16             \\
        P2P$^\dagger$               & 2.21             & \underline{3.65} & 2.93             & 66.46             & 91.36             & 78.91             \\
        ResearchStudio              & 3.30             & 3.26             & 3.28             & 42.94             & 81.38             & 62.16             \\
        \hline
        \multicolumn{7}{l}{\textit{Human reference}} \\
        Official posters            & 3.07             & 3.29             & 3.18             & 39.70             & 78.60             & 59.15             \\
        \hline
    \end{tabular}
\end{table*}

\begin{table*}[ht]
    \centering
    \caption{Visual-presentation (\%) results across all 100 papers re-evaluated with Gemini-4-Argon. Higher is better.}
    \label{tab:argon-table3}
    \vspace{4pt}
    \small
    \setlength{\tabcolsep}{7pt}
    \begin{tabular}{lcccc}
        \hline
        Method & Figure Presentation$\uparrow$ & Visual Guidance$\uparrow$ & Layout$\uparrow$ & Overall$\uparrow$ \\
        \hline
        \multicolumn{5}{l}{\textit{Ours}} \\
        \textbf{VASP (Opus-5)}      & \underline{75.3} & \underline{90.0} & \textbf{94.0}    & \textbf{86.4}    \\
        \textbf{VASP (GPT-5.6-Sol)} & \textbf{83.7}    & 82.8             & \underline{90.3} & \underline{85.6} \\
        \hline
        \multicolumn{5}{l}{\textit{Baseline methods}} \\
        Paper2Poster                & 12.0             & 35.6             & 36.3             & 28.0             \\
        PosterMELD                  & 50.3             & 58.4             & 44.3             & 51.0             \\
        PosterForest                & 20.7             & 39.2             & 36.5             & 32.1             \\
        PosterGen                   & 21.0             & 48.2             & 63.5             & 44.2             \\
        P2P                         & 2.0              & 52.8             & 60.3             & 38.4             \\
        ResearchStudio              & 65.7             & \textbf{94.6}    & 88.3             & 82.8             \\
        \hline
        \multicolumn{5}{l}{\textit{Human reference}} \\
        Official posters            & 61.0             & 75.6             & 92.0             & 76.2             \\
        \hline
    \end{tabular}
\end{table*}

\begin{table*}[ht]
    \centering
    \caption{Author-informed key-information coverage (\%) across all 100 papers re-evaluated with Gemini-4-Argon. Higher is better. Full: share of posters covering all seven units. $^\dagger$P2P renders a page of unbounded height and is not ranked.}
    \label{tab:argon-table4}
    \vspace{4pt}
    \small
    \setlength{\tabcolsep}{7pt}
    \begin{tabular}{lccccc}
        \hline
        & \multicolumn{3}{c}{Coverage by information type} & & \\
        \cline{2-4}
        Method & Problem$\uparrow$ & Method$\uparrow$ & Results$\uparrow$ & Overall$\uparrow$ & Full$\uparrow$ \\
        \hline
        \multicolumn{6}{l}{\textit{Ours}} \\
        \textbf{VASP (Opus-5)}      & \textbf{94.0}    & \textbf{88.3}    & \textbf{85.4}    & \textbf{88.3}    & \textbf{41}    \\
        \textbf{VASP (GPT-5.6-Sol)} & \underline{87.2} & \underline{82.5} & 68.2             & 79.1             & \underline{21} \\
        \hline
        \multicolumn{6}{l}{\textit{Baseline methods}} \\
        Paper2Poster                & 72.5             & 72.9             & 73.2             & 73.1             & 12             \\
        PosterMELD                  & 81.2             & 72.5             & 46.0             & 65.0             & 5              \\
        PosterForest                & 66.4             & 75.9             & \underline{76.8} & 72.7             & 10             \\
        PosterGen                   & 85.2             & 81.8             & 74.2             & \underline{79.3} & \underline{21} \\
        P2P$^\dagger$               & 90.6             & 94.5             & 92.4             & 93.3             & 65             \\
        ResearchStudio              & 74.5             & 74.2             & 68.2             & 71.6             & 14             \\
        \hline
        \multicolumn{6}{l}{\textit{Human reference}} \\
        Official posters            & 99.3             & 96.9             & 94.9             & 96.4             & 80             \\
        \hline
    \end{tabular}
\end{table*}

Under Gemini-4-Argon, VASP continues to place first and second on VLM-as-Judge (3.72 and 3.67 vs.\ 3.28 for ResearchStudio), fixed-canvas PaperQuiz (73.58\% and 65.18\%), and Visual Presentation (86.4\% and 85.6\% vs.\ 82.8\% for ResearchStudio and 76.2\% for official posters). On Key-Information Coverage, VASP (Opus-5) ranks first among fixed-canvas methods (88.3\%, Full 41), while VASP (GPT-5.6-Sol) matches PosterGen (79.1\% vs.\ 79.3\%, Full 21) and exceeds ResearchStudio by 7.5 points. Across all 900 posters, Gemini-4-Argon achieves 82.7\% item-level agreement with GPT-5.6-Sol across all 10,800 Visual Presentation judgments and 93.0\% unit-level agreement across all 6,300 Key-Information Coverage judgments.

\subsection{Human Evaluation and Judge Agreement}
\label{app:human-eval}

We conduct two blind human studies with three evaluators not involved in developing VASP.

\paragraph{Full 100-paper blind ranking.}
On every paper in the 100-paper set ($N = 300$ rankings in total), method identities were anonymized and shuffled across five representative poster sets---VASP (GPT-5.6-Sol), ResearchStudio, Official posters, PosterGen, and Paper2Poster---and each evaluator ranked the five posters from 1 (best) to 5 on visual readability (Table~\ref{tab:human-ranking-100}).

\begin{table*}[ht]
    \centering
    \caption{Independent blind visual-readability rankings by three human evaluators across all 100 papers ($N = 300$ rankings) compared with Table~\ref{tab:visual-presentation-results} Visual Presentation scores. Lower mean rank and higher score are better.}
    \label{tab:human-ranking-100}
    \vspace{4pt}
    \small
    \setlength{\tabcolsep}{9pt}
    \begin{tabular}{lcccc}
        \hline
        & \multicolumn{2}{c}{Human Blind Ranking} & \multicolumn{2}{c}{Table~\ref{tab:visual-presentation-results} Visual Presentation} \\
        \cmidrule(lr){2-3} \cmidrule(lr){4-5}
        Method & Mean Rank$\downarrow$ & Rank$\downarrow$ & Score (\%)$\uparrow$ & Rank$\downarrow$ \\
        \hline
        \textbf{VASP (GPT-5.6-Sol)} & \textbf{1.47}    & \textbf{1}    & \textbf{94.3}    & \textbf{1}    \\
        ResearchStudio              & \underline{2.20} & \underline{2} & \underline{88.1} & \underline{2} \\
        Official posters            & 2.65             & 3             & 83.0             & 3             \\
        PosterGen                   & 4.06             & 4             & 65.6             & 4             \\
        Paper2Poster                & 4.62             & 5             & 46.5             & 5             \\
        \hline
    \end{tabular}
\end{table*}

Blind human rankings (1.47, 2.20, 2.65, 4.06, 4.62) reproduce the exact 1-to-5 ordering of Table~\ref{tab:visual-presentation-results} (94.3, 88.1, 83.0, 65.6, 46.5). On standalone readability, both human evaluators and Table~\ref{tab:visual-presentation-results} rank VASP and ResearchStudio ahead of official posters, whereas official posters lead on key-information coverage (96.4\%, Table~\ref{tab:key-information-coverage}). Author-designed posters are typically prepared to accompany live talks and vary widely in layout density, often packing denser text and smaller figures onto one board (72.7\% figure presentation and 86.4\% visual guidance in Table~\ref{tab:visual-presentation-results}).

\paragraph{Criterion-level human annotation on a 20-paper subset.}
On a randomly sampled 20-paper subset, the three independent evaluators annotated all 12 Visual Presentation criteria across the same five anonymized, per-paper shuffled poster sets (1,200 judgments per evaluator). Table~\ref{tab:human-12crit} compares the resulting scores with GPT-5.6-Sol on the same 20 papers, and Table~\ref{tab:human-agreement} reports pairwise criterion-level agreement.

\begin{table*}[ht]
    \centering
    \caption{Criterion-level human evaluation (mean across three independent evaluators, 1,200 judgments each) versus GPT-5.6-Sol on the same randomly sampled 20-paper subset.}
    \label{tab:human-12crit}
    \vspace{4pt}
    \small
    \renewcommand{\arraystretch}{1.08}
    \setlength{\tabcolsep}{4.5pt}
    \resizebox{\textwidth}{!}{%
    \begin{tabular}{lccccccc}
        \hline
        & \multicolumn{5}{c}{Human Evaluation (3-Evaluator Mean)} & \multicolumn{2}{c}{GPT-5.6-Sol (Same 20 Papers)} \\
        \cmidrule(lr){2-6} \cmidrule(lr){7-8}
        Method & \makecell{Figure\\Pres.$\uparrow$} & \makecell{Visual\\Guid.$\uparrow$} & Layout$\uparrow$ & Overall$\uparrow$ & Rank$\downarrow$ & Overall$\uparrow$ & Rank$\downarrow$ \\
        \hline
        \textbf{VASP (GPT-5.6-Sol)} & \textbf{95.6}    & \textbf{98.7}    & 93.3             & \textbf{95.9}    & \textbf{1}    & \textbf{94.0}    & \textbf{1}    \\
        ResearchStudio              & \underline{87.2} & \underline{97.3} & \textbf{94.2}    & \underline{92.9} & \underline{2} & \underline{89.2} & \underline{2} \\
        Official posters            & 83.3             & 83.6             & \underline{93.8} & 86.9             & 3             & 79.5             & 3             \\
        PosterGen                   & 72.5             & 79.8             & 73.3             & 75.2             & 4             & 71.3             & 4             \\
        Paper2Poster                & 59.7             & 66.5             & 55.4             & 60.6             & 5             & 52.4             & 5             \\
        \hline
    \end{tabular}}
\end{table*}

\begin{table*}[ht]
    \centering
    \caption{Criterion-level agreement across all 1,200 Visual Presentation judgments (20 papers $\times$ 5 methods $\times$ 12 criteria) between pairs of human evaluators and between each evaluator and GPT-5.6-Sol.}
    \label{tab:human-agreement}
    \vspace{4pt}
    \small
    \setlength{\tabcolsep}{4.5pt}
    \begin{tabular}{ccccc}
        \hline
        Category & Evaluator Pair & \makecell{Matched\\(/1,200)$\uparrow$} & \makecell{Agreement\\(\%)$\uparrow$} & \makecell{Mean\\Agreement (\%)$\uparrow$} \\
        \hline
        \multirow{3}{*}{Human--Human}       & H1 vs.\ H2          & 978   & 81.5 & \multirow{3}{*}{\textbf{83.3}} \\
                                            & H1 vs.\ H3          & 1,014 & 84.5 &                                \\
                                            & H2 vs.\ H3          & 1,007 & 83.9 &                                \\
        \hline
        \multirow{3}{*}{Human--GPT-5.6-Sol} & H1 vs.\ GPT-5.6-Sol & 981   & 81.8 & \multirow{3}{*}{\textbf{83.1}} \\
                                            & H2 vs.\ GPT-5.6-Sol & 1,023 & 85.3 &                                \\
                                            & H3 vs.\ GPT-5.6-Sol & 988   & 82.3 &                                \\
        \hline
    \end{tabular}
\end{table*}

Human evaluators and GPT-5.6-Sol again produce the identical 1-to-5 ordering; all three evaluators independently rank VASP (GPT-5.6-Sol) first (98.2, 92.9, and 96.5), with a human mean exceeding GPT-5.6-Sol's own score (95.9 vs.\ 94.0). Human--GPT-5.6-Sol criterion-level agreement (83.1\%) closely matches pairwise Human--Human agreement (83.3\%).

\subsection{Visual Theme Sensitivity and Criterion-Subset Analysis}
\label{app:robustness}

\paragraph{Visual theme sensitivity.}
To separate spatial composition from surface styling, we freeze VASP (GPT-5.6-Sol)'s HTML text, extracted figures, and grid layout on the 20-paper ablation subset and vary only the CSS stylesheet across six dimensions: canvas palette, typography, header banner, section title bars, container borders and card surfaces, and conclusion callouts. We evaluate nine themes: VASP's default (\texttt{T0}), six themes matching the styles of the six baselines (\texttt{T1}--\texttt{T6}), a monochrome wireframe with 1\,px square borders (\texttt{T7}), and a classic \LaTeX{} Beamer serif theme (\texttt{T8}) (Table~\ref{tab:theme-sensitivity}).

\begin{table*}[ht]
    \centering
    \caption{Visual-presentation scores (\%) of VASP (GPT-5.6-Sol) with frozen content and layout across nine CSS visual themes on the 20-paper subset, compared with ResearchStudio and official posters on the same 20 papers.}
    \label{tab:theme-sensitivity}
    \vspace{4pt}
    \small
    \setlength{\tabcolsep}{6.5pt}
    \begin{tabular}{lcccc}
        \hline
        Visual Theme / Method & \makecell{Figure\\Presentation$\uparrow$} & \makecell{Visual\\Guidance$\uparrow$} & Layout$\uparrow$ & Overall$\uparrow$ \\
        \hline
        \multicolumn{5}{l}{\textit{VASP (GPT-5.6-Sol) with fixed layout across 9 visual themes}} \\
        \texttt{T0}: VASP Default               & 80.0             & \textbf{100.0}   & \textbf{91.2}    & \textbf{90.4}    \\
        \texttt{T1}: Paper2Poster Style         & 73.3             & \textbf{100.0}   & \underline{90.0} & 87.8             \\
        \texttt{T2}: PosterForest Style         & \underline{83.3} & 99.0             & 87.5             & \underline{89.9} \\
        \texttt{T3}: PosterMELD Style           & 75.0             & \textbf{100.0}   & 87.5             & 87.5             \\
        \texttt{T4}: PosterGen Style            & 70.8             & \textbf{100.0}   & \textbf{91.2}    & 87.4             \\
        \texttt{T5}: ResearchStudio Style       & 73.3             & \textbf{100.0}   & 88.8             & 87.4             \\
        \texttt{T6}: P2P Style                  & 72.5             & \textbf{100.0}   & \underline{90.0} & 87.5             \\
        \texttt{T7}: Monochrome Wireframe       & 74.2             & \textbf{100.0}   & 88.8             & 87.6             \\
        \texttt{T8}: \LaTeX{} Beamer Classic    & \textbf{84.2}    & \textbf{100.0}   & 82.5             & 88.9             \\
        \hline
        Mean $\pm$ SD across 9 themes           & 76.3 $\pm$ 5.0   & 99.9 $\pm$ 0.3   & 88.6 $\pm$ 2.7   & 88.3 $\pm$ 1.2   \\
        \hline
        \multicolumn{5}{l}{\textit{Baseline and human references on the same 20 papers}} \\
        ResearchStudio                          & 69.2             & 95.0             & 87.5             & 83.9             \\
        Official posters                        & 67.5             & 89.0             & 86.2             & 80.9             \\
        \hline
    \end{tabular}
\end{table*}

Across all nine themes, overall scores stay between 87.4 and 90.4 (88.3 $\pm$ 1.2), including the monochrome wireframe (\texttt{T7}, 87.6) and the ResearchStudio-matched style (\texttt{T5}, 87.4), compared with 83.9 for ResearchStudio and 80.9 for official posters on the same 20 papers. Dropping the two style-related criteria in Table~\ref{tab:visual-presentation-results} (\texttt{COL01} color harmony and \texttt{EN05} subordinate branding/decoration) across all 100 papers yields 10-criterion scores of 94.3 (GPT-5.6-Sol) and 92.8 (Opus-5), compared with 88.1 for ResearchStudio.

\paragraph{Criteria without programmatic counterparts.}
Five of the twelve Visual Presentation criteria correspond to geometric properties checked by VisComposer's controller (Appendix~\ref{app:layout-synthesis}): figure size (\texttt{IS01}), overlap (\texttt{AL01}), spatial association (\texttt{AL05}), density (\texttt{DEN01}), and whitespace (\texttt{WS01}). The remaining seven criteria (\texttt{AE04}, \texttt{AE13}, \texttt{ENH01}, \texttt{DIST01}, \texttt{COL01}, \texttt{EN05}, \texttt{EN07}) assess rendered visual qualities without programmatic counterparts in the controller, and Key-Information Coverage (Table~\ref{tab:key-information-coverage}) is scored against frozen paper-audited units extracted from the official poster. Table~\ref{tab:heldout-criteria} reports scores on these seven criteria across all 100 papers alongside the full benchmark and coverage, with VASP ranking first and second (97.1\% and 94.4\%, vs.\ 91.6\% for ResearchStudio and 82.3\% for official posters).

\begin{table*}[ht]
    \centering
    \caption{Visual Presentation on the 7 criteria without programmatic counterparts in VisComposer (\texttt{AE04}, \texttt{AE13}, \texttt{ENH01}, \texttt{DIST01}, \texttt{COL01}, \texttt{EN05}, \texttt{EN07}) across all 100 papers, shown alongside full 12-criterion Visual Presentation and Key-Information Coverage. $^\dagger$P2P renders a page of unbounded height and is not ranked on coverage.}
    \label{tab:heldout-criteria}
    \vspace{4pt}
    \small
    \setlength{\tabcolsep}{7pt}
    \begin{tabular}{lccc}
        \hline
        Method & \makecell{Visual Presentation\\(All 12 Criteria)$\uparrow$} & \makecell{Visual Presentation\\(7 Held-Out Criteria)$\uparrow$} & \makecell{Key-Information\\Coverage (Table~\ref{tab:key-information-coverage})$\uparrow$} \\
        \hline
        \multicolumn{4}{l}{\textit{Ours}} \\
        \textbf{VASP (Opus-5)}      & \underline{92.5} & \underline{94.4} & \textbf{88.7}    \\
        \textbf{VASP (GPT-5.6-Sol)} & \textbf{94.3}    & \textbf{97.1}    & \underline{82.0} \\
        \hline
        \multicolumn{4}{l}{\textit{Baseline methods}} \\
        Paper2Poster                & 46.5             & 53.7             & 74.1             \\
        PosterMELD                  & 66.3             & 79.6             & 66.9             \\
        PosterForest                & 45.5             & 52.7             & 73.9             \\
        PosterGen                   & 65.6             & 70.3             & 79.7             \\
        P2P$^\dagger$               & 49.6             & 58.6             & 93.6             \\
        ResearchStudio              & 88.1             & 91.6             & 73.0             \\
        \hline
        \multicolumn{4}{l}{\textit{Human reference}} \\
        Official posters            & 83.0             & 82.3             & 96.4             \\
        \hline
    \end{tabular}
\end{table*}

\subsection{Scoring Sensitivity and Paired Confidence Intervals}
\label{app:statistical-robustness}

\paragraph{\texttt{N/A} frequency and \texttt{Excl.\ N/A} scoring sensitivity.}
Across all 900 posters (10,800 judgments), 7 of the 12 Visual Presentation criteria have zero \texttt{N/A} verdicts on every poster. Nearly all \texttt{N/A} verdicts occur on \texttt{EN05} (subordinate branding/decoration in Visual Guidance, triggered on minimalist posters without decorative elements), with the remaining 0\%--0.8\% occurring on the three Figure Presentation criteria when a poster displays no figures (plus 2 official posters without a conclusion block on \texttt{EN07}). Under \texttt{Excl.\ N/A}, for each poster $i$ and group $g$, we replace the fixed group size $|\mathcal{Q}_g| \in \{3, 5, 4\}$ in Equation~\ref{eq:vp-score} with the number of applicable (non-\texttt{N/A}) criteria $|\mathcal{Q}_{g,i}^{\mathrm{app}}|$ (setting a group score to zero if no criteria in that group are applicable, e.g., when a poster displays no figures), and average the three groups equally (Table~\ref{tab:na-sensitivity}).

\begin{table*}[ht]
    \centering
    \caption{\texttt{N/A} frequency (\%) by method across all 900 posters and comparison between the fixed-denominator score (Table~\ref{tab:visual-presentation-results}) and \texttt{Excl.\ N/A} per-group renormalization.}
    \label{tab:na-sensitivity}
    \vspace{4pt}
    \small
    \setlength{\tabcolsep}{4.2pt}
    \begin{tabular}{lcccccccc}
        \hline
        & \multicolumn{2}{c}{N/A Rate (\%)} & & \multicolumn{2}{c}{Visual Guidance} & & \multicolumn{2}{c}{Overall} \\
        \cmidrule(lr){2-3} \cmidrule(lr){5-6} \cmidrule(lr){8-9}
        Method & \texttt{EN05} & Total & \makecell{Figure\\Pres.$\uparrow$} & Paper$\uparrow$ & Excl.\ N/A$\uparrow$ & Layout$\uparrow$ & Paper$\uparrow$ & Excl.\ N/A$\uparrow$ \\
        \hline
        \multicolumn{9}{l}{\textit{Ours}} \\
        \textbf{VASP (Opus-5)}      & 5.0  & 0.7 & \underline{81.3} & \underline{98.4} & \underline{99.4} & \textbf{97.8}    & \underline{92.5} & \underline{92.8} \\
        \textbf{VASP (GPT-5.6-Sol)} & 0.0  & 0.0 & \textbf{88.0}    & \textbf{100.0}   & \textbf{100.0}   & \underline{95.0} & \textbf{94.3}    & \textbf{94.3}    \\
        \hline
        \multicolumn{9}{l}{\textit{Baseline methods}} \\
        Paper2Poster                & 84.0 & 7.8 & 35.0             & 61.6             & 73.9             & 43.0             & 46.5             & 50.6             \\
        PosterMELD                  & 40.0 & 3.8 & 71.7             & 80.6             & 87.6             & 46.5             & 66.3             & 68.6             \\
        PosterForest                & 91.0 & 8.3 & 38.0             & 56.4             & 68.8             & 42.0             & 45.5             & 49.6             \\
        PosterGen                   & 28.0 & 2.8 & 53.0             & 75.4             & 79.7             & 68.3             & 65.6             & 67.0             \\
        P2P                         & 2.0  & 0.9 & 11.0             & 76.4             & 76.7             & 61.3             & 49.6             & 49.6             \\
        ResearchStudio              & 0.0  & 0.0 & 73.7             & 97.0             & 97.0             & 93.5             & 88.1             & 88.1             \\
        \hline
        \multicolumn{9}{l}{\textit{Human reference}} \\
        Official posters            & 3.0  & 0.8 & 72.7             & 86.4             & 87.4             & 90.0             & 83.0             & 83.5             \\
        \hline
    \end{tabular}
\end{table*}

Excluding \texttt{N/A} raises minimalist baselines such as Paper2Poster ($+4.1$) and PosterForest ($+4.1$), while keeping the top-6 ranking unchanged and VASP ahead of ResearchStudio (94.3 and 92.8 vs.\ 88.1) and PosterMELD/PosterGen ($>24$ points). Dropping \texttt{EN05} altogether yields similar overall scores for the top six methods (94.3, 92.9, 88.3, 82.6, 68.3, and 65.9).

\paragraph{Paper-paired 95\% bootstrap confidence intervals.}
Table~\ref{tab:paired-ci} reports paper-paired 95\% bootstrap confidence intervals (10,000 resamples) for $\Delta = \text{VASP (GPT-5.6-Sol)} - \text{Baseline}$ on PaperQuiz and Key-Information Coverage across all 100 papers, alongside Visual Presentation.

\begin{table*}[ht]
    \centering
    \caption{Paper-paired 95\% bootstrap confidence intervals (10,000 resamples) for $\Delta = \text{VASP (GPT-5.6-Sol)} - \text{Method}$ on PaperQuiz and Key-Information Coverage across all 100 papers, alongside Table~\ref{tab:visual-presentation-results} Visual Presentation. Intervals excluding zero with positive $\Delta$ are shown in bold.}
    \label{tab:paired-ci}
    \vspace{4pt}
    \small
    \renewcommand{\arraystretch}{1.15}
    \setlength{\tabcolsep}{4pt}
    \resizebox{\textwidth}{!}{%
    \begin{tabular}{lccccc}
        \hline
        Method & \makecell{PaperQuiz\\(\%)$\uparrow$} & \makecell{VASP $\Delta$\\{[Paired 95\% CI]}} & \makecell{Coverage\\(\%)$\uparrow$} & \makecell{VASP $\Delta$\\{[Paired 95\% CI]}} & \makecell{Visual\\Pres.\ (\%)$\uparrow$} \\
        \hline
        \textbf{VASP (GPT-5.6-Sol)} & \textbf{69.96}    & ---                               & \textbf{82.0}    & ---                                  & \textbf{94.3}    \\
        \hline
        Paper2Poster                & \underline{69.25} & $+0.71\ [-1.20,\ +2.68]$          & 74.1             & $\mathbf{+7.86\ [+4.00,\ +11.71]}$   & 46.5             \\
        PosterForest                & 68.86             & $+1.10\ [-0.66,\ +2.92]$          & 73.9             & $\mathbf{+8.14\ [+4.29,\ +12.00]}$   & 45.5             \\
        PosterGen                   & 67.72             & $\mathbf{+2.24\ [+0.62,\ +3.84]}$ & \underline{79.7} & $+2.29\ [-0.57,\ +5.14]$             & 65.6             \\
        ResearchStudio              & 67.48             & $\mathbf{+2.48\ [+1.14,\ +3.82]}$ & 73.0             & $\mathbf{+9.00\ [+5.29,\ +12.86]}$   & \underline{88.1} \\
        PosterMELD                  & 62.36             & $\mathbf{+7.60\ [+5.93,\ +9.24]}$ & 66.9             & $\mathbf{+15.14\ [+11.86,\ +18.57]}$ & 66.3             \\
        \hline
        Official posters            & 67.55             & $\mathbf{+2.41\ [+0.32,\ +4.68]}$ & 96.4             & $-14.43\ [-17.14,\ -11.86]$          & 83.0             \\
        \hline
    \end{tabular}}
\end{table*}

VASP significantly outperforms ResearchStudio on both PaperQuiz ($+2.48$, $[+1.14, +3.82]$) and Coverage ($+9.00$, $[+5.29, +12.86]$), significantly outperforms PosterGen on PaperQuiz ($+2.24$, $[+0.62, +3.84]$) while leading Visual Presentation by 28.7 points, and significantly outperforms Paper2Poster and PosterForest on Coverage ($+7.86$ and $+8.14$) while leading Visual Presentation by over 47 points.

\clearpage

\section{Additional Examples and Failure Cases}
\label{app:examples}

\begin{figure}[p]
\centering
\includegraphics[width=\linewidth]{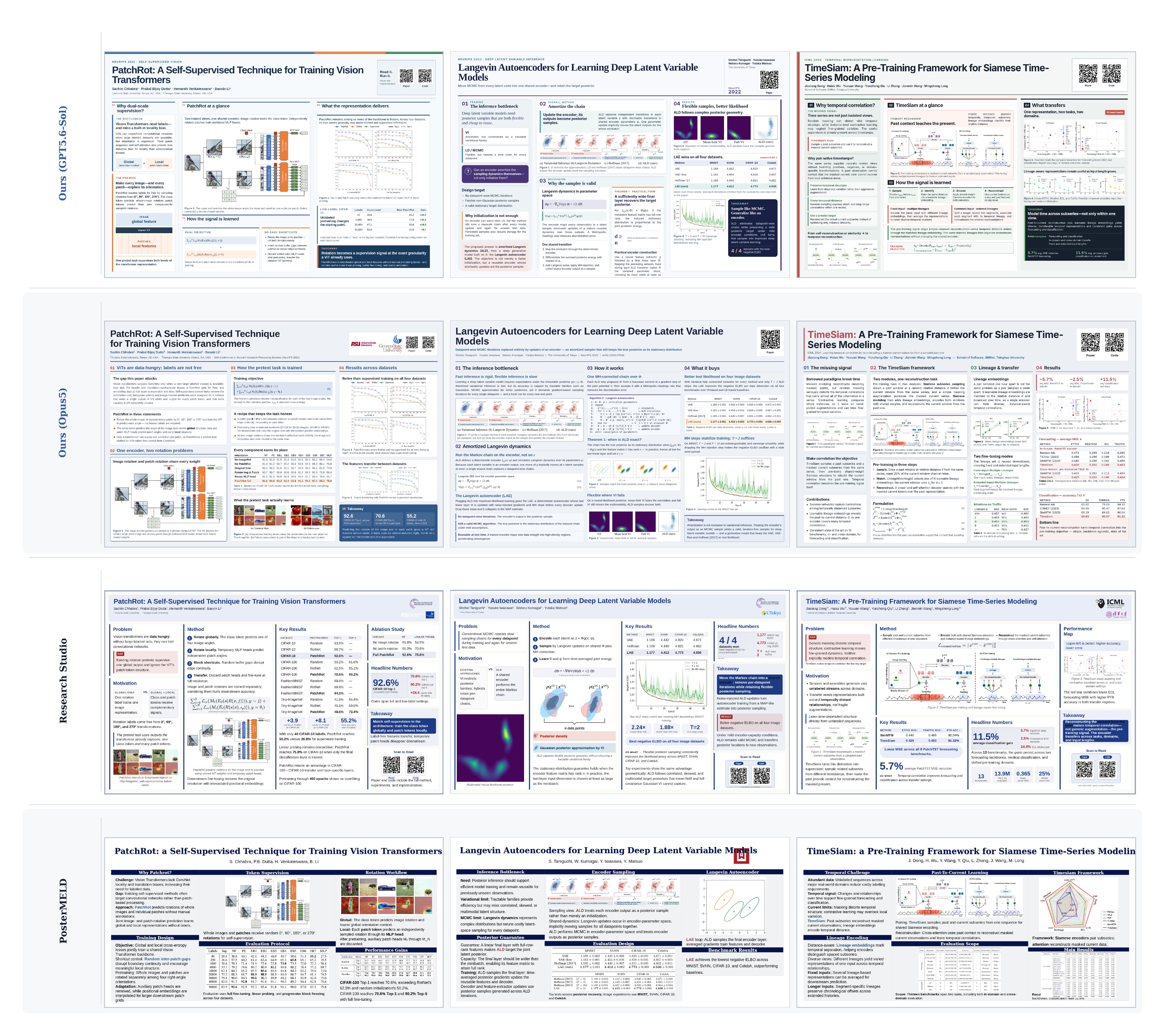}
\caption{Posters generated for three papers of the evaluation set (columns) by VASP (GPT-5.6-Sol), VASP (Opus-5), ResearchStudio, and PosterMELD (rows, from top to bottom). Rows labeled Ours show VASP.}
\label{fig:examples-a}
\end{figure}

\begin{figure}[p]
\centering
\includegraphics[width=\linewidth]{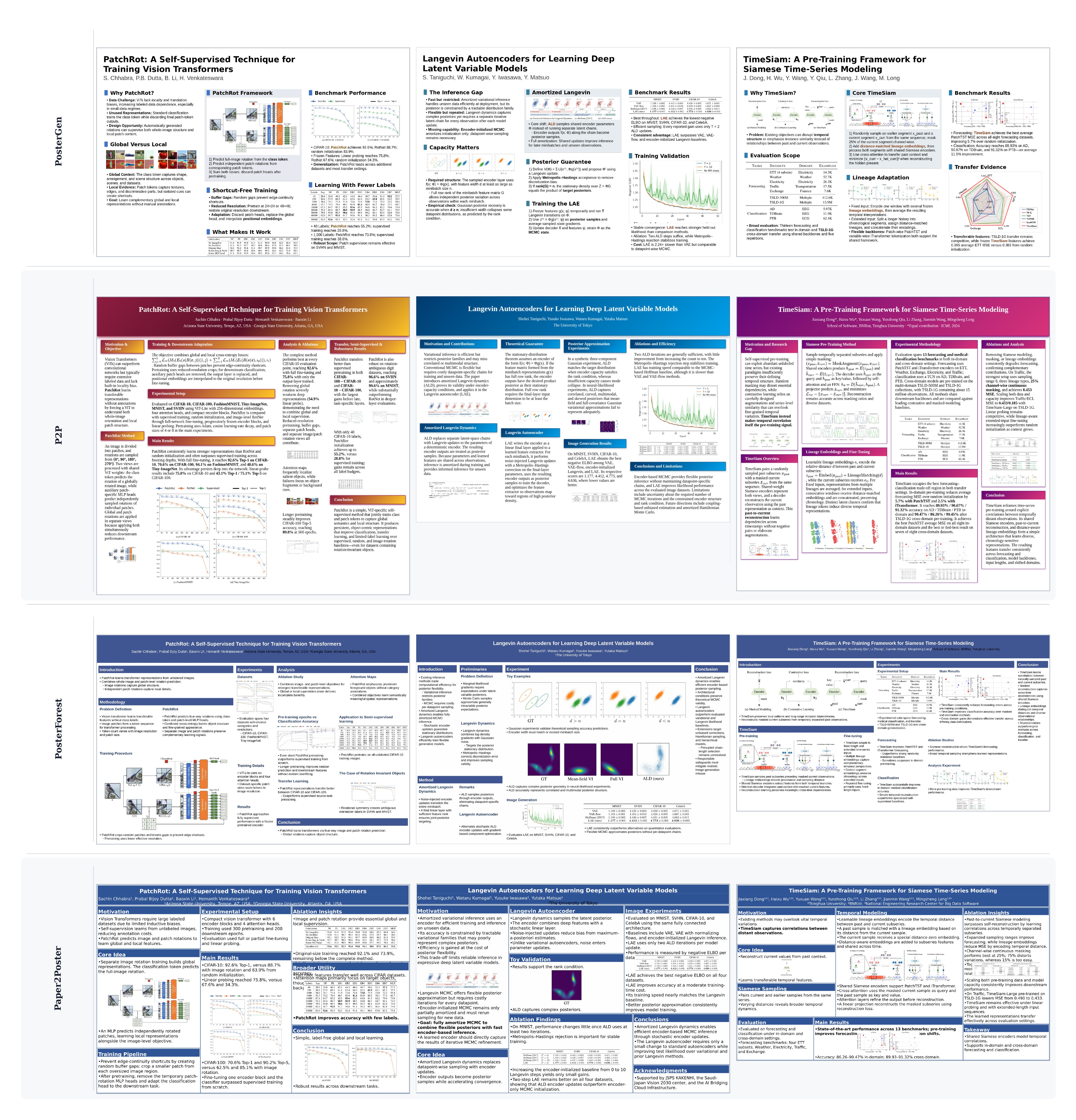}
\caption{Posters generated for the three papers of Figure~\ref{fig:examples-a} (columns) by PosterGen, P2P, PosterForest, and Paper2Poster (rows, from top to bottom).}
\label{fig:examples-b}
\end{figure}

\paragraph{Additional examples.}
Figures~\ref{fig:examples-a} and~\ref{fig:examples-b} show the posters of all eight methods for three papers of the evaluation set, \emph{PatchRot: A Self-Supervised Technique for Training Vision Transformers}, \emph{Langevin Autoencoders for Learning Deep Latent Variable Models}, and \emph{TimeSiam: A Pre-Training Framework for Siamese Time-Series Modeling}. Each row shows one method and each column one paper.

\begin{figure}[h]
\centering
\begin{minipage}[t]{0.325\textwidth}\centering
\includegraphics[width=\linewidth]{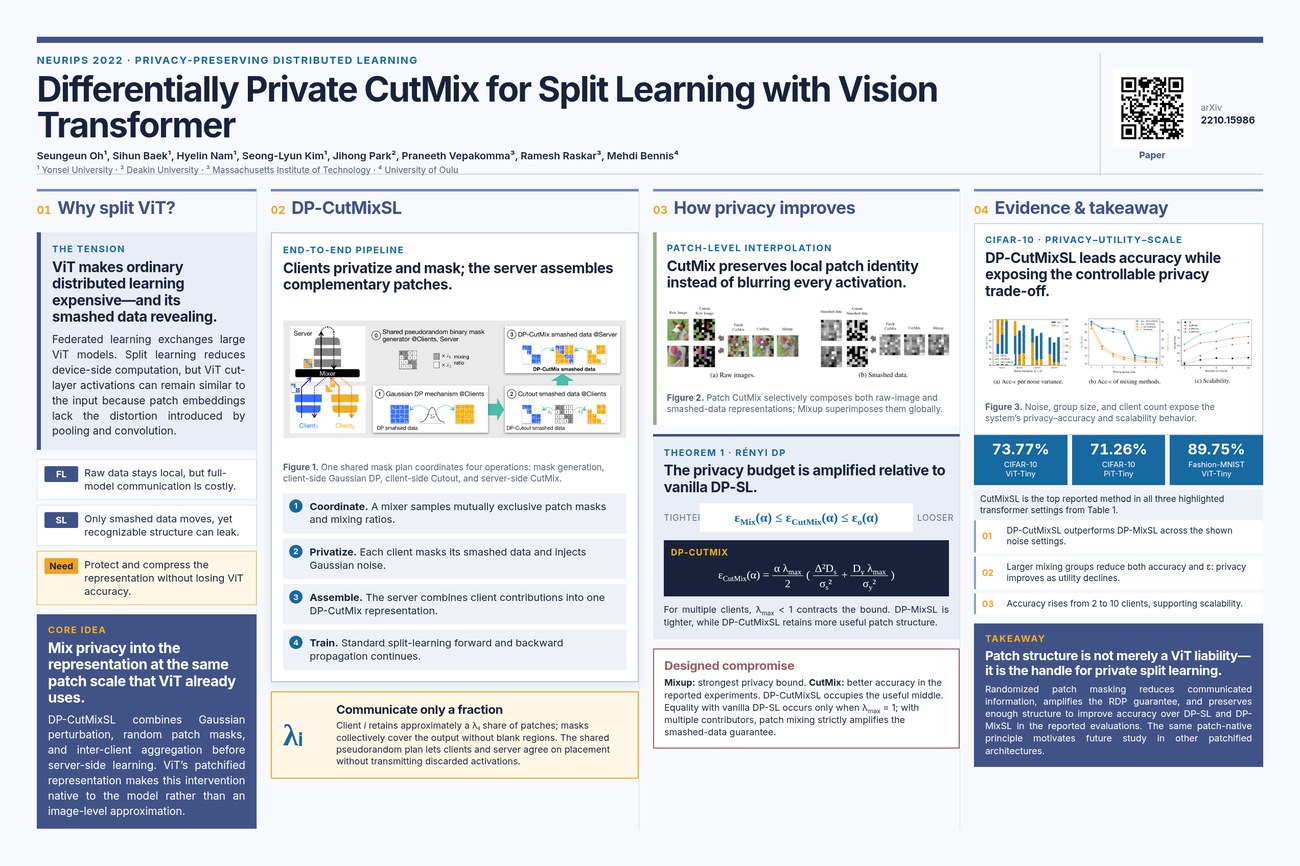}\\[1pt]
{\small (a) \texttt{AE13}, \texttt{IS01}, \texttt{DEN01}}
\end{minipage}\hfill
\begin{minipage}[t]{0.325\textwidth}\centering
\includegraphics[width=\linewidth]{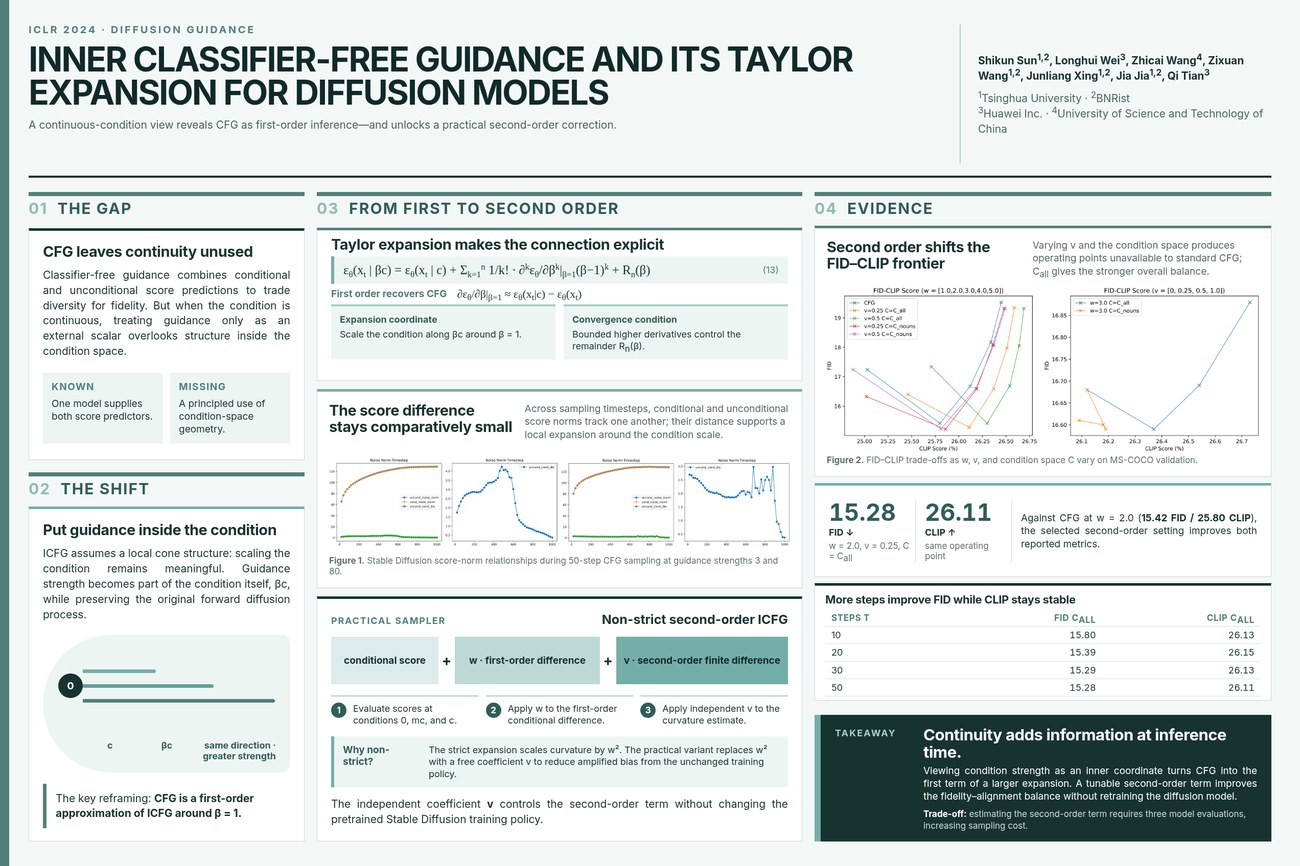}\\[1pt]
{\small (b) \texttt{AE13}, \texttt{IS01}, \texttt{DEN01}}
\end{minipage}\hfill
\begin{minipage}[t]{0.325\textwidth}\centering
\includegraphics[width=\linewidth]{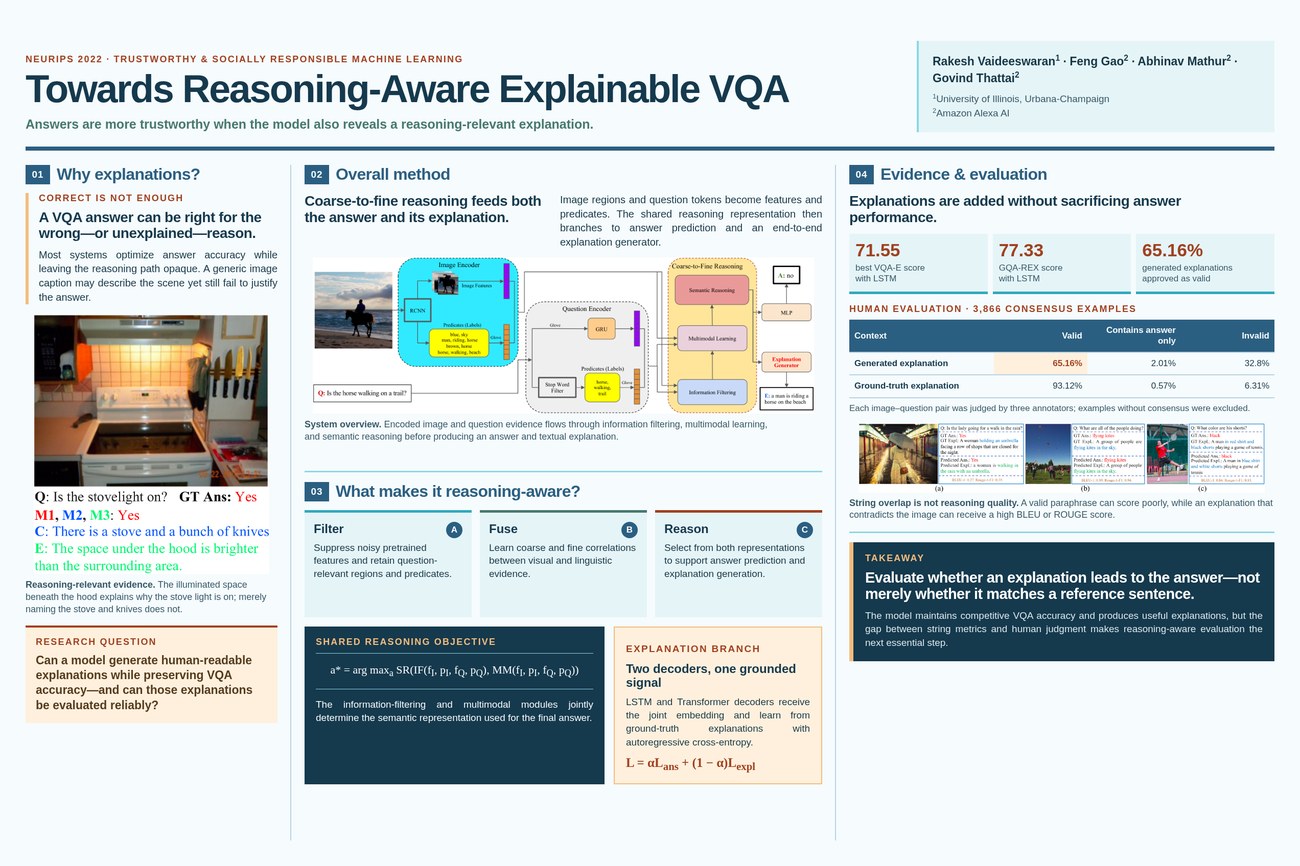}\\[1pt]
{\small (c) \texttt{AE13}, \texttt{IS01}, \texttt{WS01}}
\end{minipage}
\caption{The three lowest-scoring posters of VASP (GPT-5.6-Sol) on the visual-presentation benchmark: (a)~\emph{Differentially Private CutMix for Split Learning with Vision Transformer}, (b)~\emph{Inner Classifier-Free Guidance and Its Taylor Expansion for Diffusion Models}, and (c)~\emph{Towards Reasoning-Aware Explainable VQA}. Labels list the criteria judged \textsc{Fail} (Appendix~\ref{app:visual-definitions}).}
\label{fig:failure-cases}
\end{figure}

\paragraph{Failure cases.}
Figure~\ref{fig:failure-cases} shows the three lowest-scoring posters of VASP (GPT-5.6-Sol) on the visual-presentation benchmark, each with an overall score of 69.4. All three share one failure mode: a multi-panel result figure is compressed into a shallow strip, so its legends, axis labels, and tick values become too small to read. The judge therefore fails the legibility of necessary in-figure text (\texttt{AE13}) and the size relative to information complexity (\texttt{IS01}), and for the first two posters also the visual information density (\texttt{DEN01}). The third poster additionally leaves a large empty region below its takeaway panel (\texttt{WS01}). These cases fall under the figure floor of Appendix~\ref{app:layout-synthesis}, which bounds the displayed size of a whole figure but not the size of the text inside each of its panels.

\clearpage

\section{Visual Presentation Benchmark}
\label{app:visual-benchmark}

\subsection{Overview and Scope}
\label{app:visual-overview}

The visual-presentation benchmark evaluates properties that can be judged directly from a rendered academic poster. It contains 12 atomic criteria organized into three groups: figure presentation, visual guidance and emphasis, and layout and spatial organization. It does not assess scientific correctness, faithfulness to the source paper, or completeness of paper content. Inserted paper figures are treated as fixed assets: the judge evaluates their caption association, displayed scale, and poster-level placement, without reassessing their internal scientific content, connector logic, source resolution, aspect ratio, or color semantics.

Each poster--criterion pair is evaluated independently. The judge receives one poster image and the complete rule for one criterion, and returns a structured verdict with visible evidence. Except for \texttt{DIST01}, all criteria use the original poster rendering. \texttt{DIST01} uses the deterministic reduced-scale stimulus described in Appendix~\ref{app:visual-definitions}.

\subsection{Detailed Criterion Definitions}
\label{app:visual-definitions}

This section gives the operational definition of each criterion, grouped into figure presentation (Appendix~\ref{app:crit-figure}), visual guidance and emphasis (Appendix~\ref{app:crit-guidance}), and layout and spatial organization (Appendix~\ref{app:crit-layout}). Appendix~\ref{app:benchmark-prompts} reproduces the complete prompts, including detailed edge cases, evidence requirements, and output schemas.

\subsubsection{Figure Presentation}
\label{app:crit-figure}

\paragraph{\texttt{AE04}: nearby figure description.}
Every displayed figure must have a nearby caption or short identifying label with unambiguous ownership. One unlabeled or ambiguously labeled figure causes failure. A poster without figures is \textsc{Not-Applicable}.

\paragraph{\texttt{AE13}: legibility of necessary in-figure text.}
For each primary method or result figure, functionally necessary in-figure text must be displayed at least approximately 30\% as tall as typical body text. Text is necessary when hiding it would prevent interpretation of the figure's main objects, conditions, visual encoding, process, axes, or emphasized result. Borderline size comparisons are \textsc{Uncertain}. The criterion is \textsc{Not-Applicable} when the required comparison cannot be formed.

\paragraph{\texttt{IS01}: size relative to information complexity.}
Information-bearing visuals must receive enough area for their main structure and evidence to remain distinguishable. Logos and decorative images are excluded, and an isolated minor label does not by itself cause failure.

\subsubsection{Visual Guidance and Emphasis}
\label{app:crit-guidance}

\paragraph{\texttt{ENH01}: visual hierarchy.}
A successful poster provides a clear entry point and distinguishable primary, secondary, and supporting levels. Competing focal regions, inverted salience, or persistent style conflicts cause failure.

\begin{figure}[t]
    \centering
    \includegraphics[width=0.98\linewidth]{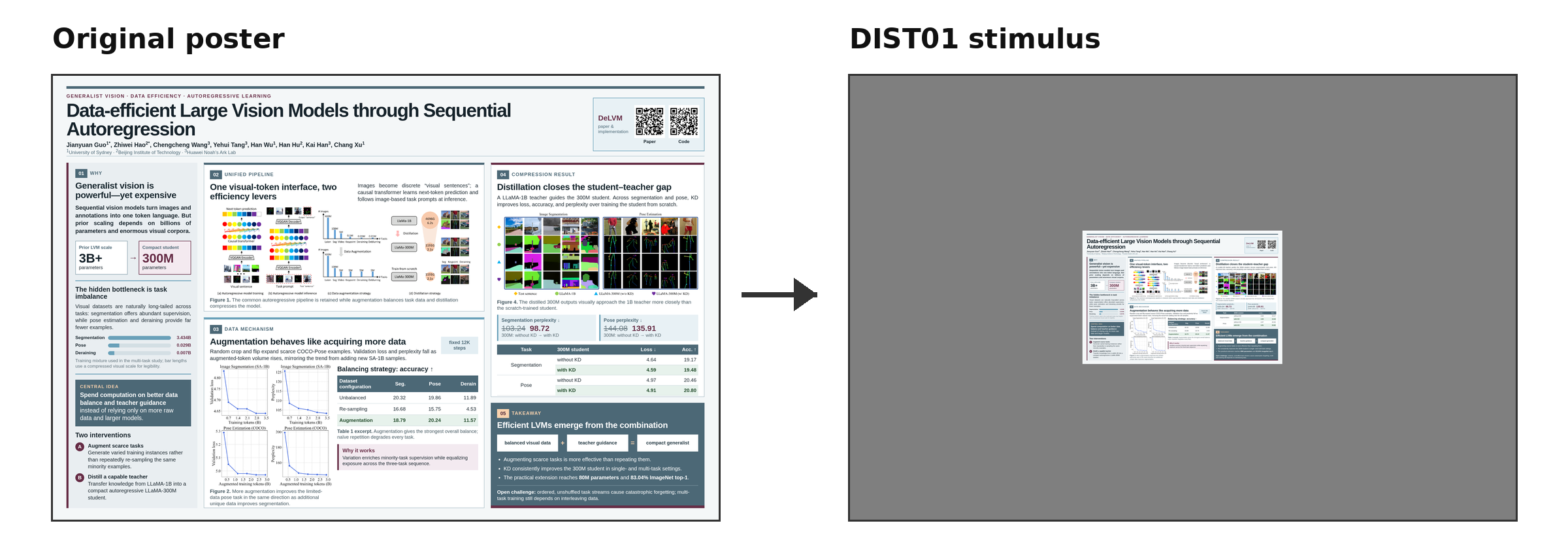}
    \caption{Construction of the \texttt{DIST01} stimulus. The original poster (left) is reduced to 30\% of its width and height and centered, without subsequent enlargement, on a neutral-gray canvas with the original dimensions (right).}
    \label{fig:dist01-stimulus}
\end{figure}

\paragraph{\texttt{DIST01}: reduced-scale recognizability.}
For a poster of width $W$ and height $H$, we first compute
\begin{equation}
    W'=\max(1,\operatorname{round}(0.3W)),\qquad
    H'=\max(1,\operatorname{round}(0.3H)).
\label{eq:dist01-size}
\end{equation}
The poster is resized once to $W'\times H'$ from Equation~\ref{eq:dist01-size} with LANCZOS resampling and centered on a $W\times H$ neutral-gray canvas (\texttt{\#7F7F7F}), without subsequent enlargement (Figure~\ref{fig:dist01-stimulus}). A pass requires the title or topical entry, main focal region, core sections, broad reading progression, and a key result or conclusion region to remain recognizable. Small body text need not remain readable. This operation is a controlled image-scale proxy rather than a calibrated physical viewing distance.

\paragraph{\texttt{COL01}: color harmony.}
Poster-level colors should form a coherent palette with visible priority and adequate foreground--background distinction. Figure-internal color encodings are treated as fixed.

\paragraph{\texttt{EN05}: subordinate branding and decoration.}
Branding and decoration must not dominate the title or scientific content or consume disproportionate space. The criterion is \textsc{Not-Applicable} when no substantive branding or decorative treatment is present.

\paragraph{\texttt{EN07}: visually enhanced conclusion information.}
At least one readable, substantively complete conclusion unit must be visually distinguished from nearby ordinary content by one strong emphasis cue or two independent weaker cues. A section heading, isolated number, or promotional fragment is insufficient. When no readable conclusion unit exists, the criterion is \textsc{Not-Applicable}.

\subsubsection{Layout and Spatial Organization}
\label{app:crit-layout}

\paragraph{\texttt{AL01}: harmful overlap.}
Separately placed poster elements must not intersect in ways that obscure content or create ownership ambiguity. Proximity without intersection and overlap internal to an inserted paper figure do not cause failure.

\paragraph{\texttt{AL05}: spatial association.}
Headings, captions, legends, bullet continuations, and explanatory labels must be spatially grouped with the content they describe. The criterion is \textsc{Not-Applicable} when none of these auxiliary elements exists.

\paragraph{\texttt{DEN01}: visual information density.}
The amount and complexity of visible content should match the area it occupies. Systematic crowding, severely imbalanced density, or an overall overfilled or underfilled composition causes failure.

\paragraph{\texttt{WS01}: ineffective whitespace.}
A large empty region is acceptable when it supports grouping, emphasis, reading flow, de-crowding, or rhythm. Failure requires a clearly large region without an observable layout function. No fixed area threshold is used.

\subsection{Judgment Output and Scoring}
\label{app:visual-scoring}
For each poster $i$ and criterion $q$, the judge returns a verdict $r_{i,q}$ from \textsc{Pass}, \textsc{Fail}, \textsc{Uncertain}, and \textsc{Not-Applicable}, together with visible evidence, an approximate location, and confidence. Only \textsc{Pass} receives credit. For criterion set $\mathcal{Q}_g$ in group $g$, the group score is the proportion of criteria receiving \textsc{Pass},
\begin{equation}
\begin{aligned}
v_{i,g} &=
\frac{\big|\{q \in \mathcal{Q}_g :
r_{i,q}=\text{\textsc{Pass}}\}\big|}
{|\mathcal{Q}_g|},\\
V &= \frac{1}{3N}\sum_{i=1}^{N}\sum_{g=1}^{3}v_{i,g}.
\end{aligned}
\label{eq:vp-score}
\end{equation}
In Equation~\ref{eq:vp-score}, criteria are equally weighted within each group, the three groups are equally weighted, and posters are equally weighted. Appendix~\ref{app:statistical-robustness} reports \texttt{N/A} frequencies by method and compares fixed-denominator scores with \texttt{N/A}-excluded renormalization.

\subsection{Evaluation Execution}
\label{app:visual-execution}

Visual judgments use GPT-5.6-Sol with medium reasoning effort, called through an OpenAI-compatible chat-completions API. Every poster--criterion pair is judged in its own stateless request, so a decision for one criterion cannot enter another criterion's context. The system message contains the shared verifier instructions and the group-specific scope, and the user message contains the rule for a single criterion followed by one image, the rendered poster or, for \texttt{DIST01}, the distance-view stimulus. The JSON Schema of Appendix~\ref{app:benchmark-prompts} constrains the response in strict mode, transport failures are retried, and a response that fails parsing or validation is re-asked once in a fresh request. The evaluation program then verifies the requested dimension and criterion ID, the allowed verdict set, non-empty evidence and location fields, and a confidence value in $[0,1]$.

\section{Author-Informed Key-Information Coverage Benchmark}
\label{app:coverage-benchmark}

\subsection{Overview and Scope}
\label{app:coverage-overview}

This benchmark measures whether a generated poster preserves the information that authors chose to foreground in their official poster. For each of the 100 papers, exactly seven reference units are extracted from the author-designed poster, yielding 700 units. The fixed budget limits the extraction of secondary details and does not claim to exhaust the paper or the official poster. The source paper is used only to audit and revise the reference units before evaluation. During candidate-poster judgment, the evaluator receives the candidate image and its seven frozen units, but neither the paper nor the official-poster image.

\subsection{Construction of Reference Information Units}
\label{app:coverage-construction}

Initial extraction uses only the visible official poster and asks for seven semantically distinct, independently judgeable units that follow its own emphasis. The extractor may select motivation, methods, theory, equations, datasets, findings, applications, or limitations, depending on the poster. It excludes acknowledgements, citations, QR codes, decorative text, and minor implementation details.

Each record contains: (1) a stable \texttt{unit\_id}, (2) an information type, (3) a short topic, (4) two or three sentences of local context, (5) a self-contained statement, (6) the minimum semantic content required for coverage, and (7) a description of visible evidence in the official poster. Coverage requirements accept paraphrase and compression and include exact values only when they materially define the claim. For the analysis by information type in Table~\ref{tab:key-information-coverage}, the 700 units are grouped into problem and background (149 units), method and theory (291 units), and experimental and theoretical results (198 units). The remaining 62 units count toward the overall coverage only.

\subsection{Paper-Grounding Audit and Reference Revision}
\label{app:coverage-audit}

All 700 initial units are audited against page-delimited text from the corresponding paper version (Appendix~\ref{app:prompt-audit}). The official poster determines the information theme, while the paper constrains factual content, scope, assumptions, comparison sets, and numerical detail. The first pass flagged 89 units as not fully supported. An independent second pass (Appendix~\ref{app:prompt-adjudication}) confirmed 57 flags, revised the interpretation of 13, and dismissed 19. Final adjudication produced 70 revisions across 48 papers: 57 units were narrowed, eight contradicted units were corrected, and five unsupported units were replaced by related paper-grounded content. The revised records were frozen before any candidate method was scored.

\subsection{Candidate-Poster Coverage Judgment}
\label{app:coverage-judgment}

For each candidate poster, the judge assigns \textsc{Covered} or \textsc{Not-Covered} to all seven frozen units. \textsc{Covered} requires visible content that satisfies the unit's minimum semantic requirement. Faithful paraphrases, formulas, diagrams, plot labels, captions, and legible tables are accepted even when they appear in a different section. A unit is \textsc{Not-Covered} when its core meaning is absent, too vague, unreadable, represented only by an unrelated keyword, or contradicted. Extra information is not penalized and partial credit is not used. The evidence field quotes a short visible phrase or identifies a specific visual, table, or section.

For an evaluated method $m$, let $c_{m,i,j}\in\{0,1\}$ indicate whether its poster for paper $i$ covers reference unit $j$. The method-level coverage rate is
\begin{equation}
    C_m=\frac{1}{7N}\sum_{i=1}^{N}\sum_{j=1}^{7}c_{m,i,j}.
\label{eq:coverage}
\end{equation}
Coverage in Equation~\ref{eq:coverage} measures preservation of the frozen author-informed reference and is not a comprehensive factuality score. Coverage judgments use GPT-5.6-Sol through the same API as the visual-presentation benchmark (Appendix~\ref{app:visual-execution}). The system message is the coverage prompt of Appendix~\ref{app:prompt-coverage}, the user message contains the poster identifier, the paper title, and the seven frozen units followed by the poster image, and the response is constrained by a strict JSON Schema.

\subsection{Scope and Limitations}
\label{app:coverage-limitations}

The benchmark inherits the editorial choices of the official poster and the fixed seven-unit budget. It therefore measures retention of selected author-emphasized information rather than exhaustive paper coverage. The paper-grounding audit reduces unsupported or overbroad references but remains LLM-assisted. Coverage should therefore be interpreted as preservation of the frozen author-informed reference rather than as a comprehensive measure of paper understanding or factual correctness.

\section{Limitations}
\label{app:limitations}

In the main tables, GPT-5.6-Sol serves as the evaluator for Paper2Poster and both of our benchmarks while also driving our GPT-5.6-Sol instantiation and the primary baseline runs. Although we supplement these results with an unrelated VLM judge across all 100 papers (Gemini-4-Argon, Appendix~\ref{app:cross-family-judge}) and blind human evaluations (Appendix~\ref{app:human-eval}), full-scale evaluation remains automated and criterion-level human annotation currently covers a 20-paper subset. In addition, our benchmarks assess standalone visual legibility, figure preservation, claim--evidence organization, and retention of author-emphasized takeaways, rather than directly measuring audience comprehension in live poster sessions. Finally, figure extraction in SemParser assumes born-digital PDFs with an embedded text layer, so scanned papers require OCR prior to extraction.

% Required in the main document preamble:
% \usepackage{listings}
% \usepackage{tcolorbox}
% \tcbuselibrary{skins,breakable,listings}

\definecolor{promptTitle}{HTML}{92945F}
\definecolor{promptBody}{HTML}{F6F6DE}

\newtcblisting{promptblock}[1][Prompt]{
    enhanced jigsaw,
    breakable,
    listing only,
    listing engine=listings,
    title={#1},
    colframe=promptTitle,
    colback=promptBody,
    colbacktitle=promptTitle,
    coltitle=white,
    fonttitle=\bfseries\small,
    boxrule=0.6pt,
    arc=2mm,
    outer arc=2mm,
    left=1.5mm,
    right=1.5mm,
    top=1.2mm,
    bottom=1.2mm,
    before skip=8pt,
    after skip=8pt,
    listing options={
        basicstyle=\rmfamily\small,
        columns=fullflexible,
        keepspaces=true,
        showstringspaces=false,
        breaklines=true,
        breakatwhitespace=false,
        tabsize=2,
        literate={“}{{\textquotedblleft}}1 {”}{{\textquotedblright}}1 {`}{{\`{}}}1 {-}{{-}}1
    }
}

\section{Complete Prompt Templates}
\label{app:benchmark-prompts}

This appendix reproduces the prompt components used by the two benchmarks. For visual presentation, each call concatenates the shared verifier instructions, the corresponding group-specific scope, one criterion-specific instruction, and the output contract below. Reference extraction, paper-grounding audit, independent adjudication, and candidate-poster coverage use the four complete prompts reproduced in Sections~\ref{app:prompt-extraction}--\ref{app:prompt-coverage}. Text inside the shaded boxes is supplied to the model, whereas the short descriptions outside the boxes are explanatory and are not part of the prompts.

\subsection{Visual-Presentation Prompts}
\label{app:prompt-visual}

The visual evaluator uses a modular prompt rather than one monolithic template. The following blocks first specify the shared evidence policy (Appendix~\ref{app:prompt-shared}), then restrict the scope to one criterion group (Appendix~\ref{app:prompt-scope}), provide the operational rule for one atomic criterion (Appendix~\ref{app:prompt-criteria}), and finally impose a common structured-output contract (Appendix~\ref{app:prompt-output}).

\subsubsection{Shared verifier instructions}
\label{app:prompt-shared}

The shared instructions establish the image-only evidence boundary and define the four verdict labels used across all 12 visual criteria.

\begin{promptblock}[Prompt: shared verifier instructions]
EVALUATION-ONLY TARGETED RERUN

The attached poster image is the only evidence. Do not inspect files, run shell
commands, browse the web, call tools, or use outside knowledge. Judge only the
single requested atomic question.

You are a visual verification agent for academic posters.
You are a verifier, not a holistic scorer. Evaluate every supplied atomic
question independently and only from directly observable evidence in the
attached poster image(s).

Verdict rules:
- PASS: the positive requirement is clearly satisfied by every relevant,
  visible instance. For a universal requirement, one clear counterexample is
  enough to prevent PASS.
- FAIL: there is clear visual or textual evidence that the requirement is
  violated.
- UNCERTAIN: the relevant content exists, but resolution, ambiguity, occlusion,
  or missing context prevents a reliable judgement.
- NOT_APPLICABLE: the question's explicit applicability condition is not met.
  Do not use this merely because the answer is difficult.

Evidence rules:
- Every result must cite a concrete observation.
- For FAIL, identify the exact defect and its approximate poster location.
- For PASS, cite representative positive evidence and its location.
- For UNCERTAIN or NOT_APPLICABLE, explain why that verdict is necessary.
- Do not use outside knowledge, infer invisible content, compare against an
  imagined ideal poster, or let one question's verdict determine another.
- Do not return a 1--5 rating or an overall aesthetic opinion.

Only one atomic question is supplied in each call. Return exactly one result
with the supplied question ID and output only one valid JSON object satisfying
the supplied JSON Schema.
\end{promptblock}

\subsubsection{Group-specific scope instructions}
\label{app:prompt-scope}

Exactly one of the following scope modules is inserted into each evaluation call. These modules prevent the judge from importing concerns assigned to another criterion group.

\begin{promptblock}[Prompt: figure-presentation scope]
Treat every inserted figure, chart, table, diagram, and image crop as a trusted, immutable asset taken from the source paper. Do not judge source-image resolution, sharpness, compression, pixelation, aspect ratio, internal color encoding, internal contrast, scientific correctness, label-to-mark relationships, connector logic, or any other relationship inside a figure. Judge only the figure asset's poster-level caption association, displayed text size, and displayed area relative to visible information complexity. Do not judge overlaps or general spacing here because the layout dimension handles those properties consistently. Evaluate the single supplied question according to its complete question-specific instructions.
\end{promptblock}

\begin{promptblock}[Prompt: visual-engagement scope]
Judge observable whole-poster attention and engagement properties only, never whether the research topic itself is interesting. Infer information importance only from the poster's visible wording, headings, layout, and emphasis; do not use the source paper or outside knowledge. For color harmony, treat the internal palette and color semantics of each inserted paper-derived figure as trusted and judge only how that figure coexists with the poster-level background, borders, panels, and surrounding colors. Unless the supplied question explicitly provides a derived distance-view stimulus, use direct evidence from the normal poster view and do not simulate a thumbnail view. For DIST01, follow its question-specific stimulus protocol and judge only the attached derived image. Evaluate the single supplied question according to its complete question-specific instructions.
\end{promptblock}

\begin{promptblock}[Prompt: layout scope]
Judge whole-poster spatial organization only. Check for poster-level overlap, ambiguous element ownership, the match between visible content complexity and occupied space, and whether large empty regions serve an observable layout function. Do not estimate grid-alignment percentages, gap ratios, header-height percentages, section-area percentages, column-bottom differences, caption-distance ratios, or whitespace-area percentages. Evaluate the single supplied question according to its complete question-specific instructions.
\end{promptblock}

\subsubsection{Criterion-specific instructions}
\label{app:prompt-criteria}

Each block below defines one independently evaluated atomic criterion. Only the block matching the requested criterion is combined with the shared instructions, its group-specific scope, and the output contract.

\paragraph{Figure presentation.}
The first three criteria evaluate caption association, the displayed size of necessary in-figure text, and whether each information-bearing visual receives sufficient area for its visible complexity.

\begin{promptblock}[Prompt: AE04]
Question: Does every displayed figure have a corresponding caption or short identifying label placed nearby?

PASS condition:
Each displayed figure has a nearby caption or short label that can be associated with it unambiguously.

FAIL condition:
At least one displayed figure has no nearby caption or identifying label, or the nearest candidate is too far away or ambiguously associated.

Applicability:
Applicable when at least one figure is displayed; otherwise NOT_APPLICABLE. A nearby caption is sufficient; it does not need to state a takeaway.
\end{promptblock}

\begin{promptblock}[Prompt: AE13]
Question: For each primary method or result figure, is every item of functionally necessary in-figure text displayed at least approximately 30% as tall as the poster's typical body text?

Evaluation instructions:
{
  "primary_figure_definition": "A primary figure is a figure that visibly carries the poster's core method explanation, key result, or principal evidence. Identify primary figures only from their displayed size, placement, section heading, nearby caption, and visual emphasis. Do not automatically treat every inserted image, logo, decorative visual, or supporting example as primary.",
  "necessary_text_definition": "Functionally necessary in-figure text is text whose loss of recognizability would prevent a viewer from completing the figure's main communication function on this poster. Infer that function only from the poster's visible heading, nearby caption, placement, size, and emphasis; do not use the source paper or outside knowledge.",
  "functional_deletion_test": "Mentally hide the candidate text without hiding the surrounding visual. Treat it as necessary when the viewer would then be unable to identify a principal object, distinguish a key condition, decode an essential visual encoding, follow the main process, interpret a key coordinate relationship, or locate the figure's visibly emphasized result.",
  "likely_inclusions": [
    "Legend entries, color-category labels, line-style labels, and data-series names when they are needed to decode the figure's principal marks.",
    "Axis names and units when the variables or quantities are part of the figure's main communication function.",
    "Representative tick values when the figure's main visible function requires quantitative comparison, scale interpretation, or reading a value range; not every redundant tick must qualify.",
    "Main node, stage, module, or path labels in a flow diagram, framework diagram, or architecture schematic when they are needed to follow the principal process or structure.",
    "Condition names, method names, sample categories, or subpanel headings when they are needed to distinguish the figure's principal comparisons.",
    "Category, condition, or stage labels in an image group when they are needed to understand the central comparison among images.",
    "Visibly emphasized key numbers, result annotations, or short result statements when they perform the figure's central result-communication function.",
    "Subpanel identifiers only when a nearby caption or poster-level callout visibly refers to those identifiers and the identifiers are needed to establish the correspondence."
  ],
  "likely_exclusions": [
    "Citations, copyright notices, data-source notes, and watermarks unless the poster visibly makes them part of the figure's central message.",
    "Logos, branding text, and decorative lettering.",
    "Repeated labels whose loss would not impair the main interpretation.",
    "Secondary parameter values, sample IDs, file names, or auxiliary technical details that are not needed for the principal visible message.",
    "Dense collections of minor values that do not participate in the figure's main comparison.",
    "Local annotations whose removal would not alter the figure's principal visible function.",
    "Ordinary superscripts and subscripts as separately undersized text; compare the base characters of the containing text instead.",
    "Poster body text and captions outside the figure. They may establish the figure's main function but are not themselves measured as in-figure text."
  ],
  "typical_body_text_reference": "Use the recurring regular paragraph text in the poster's main content as the typical body-text reference. Exclude the poster title, section headings, figure captions, references, affiliations, footer or contact text, and specially enlarged conclusion or callout text. If ordinary body-text sizes vary slightly, use the most common representative paragraph size.",
  "comparison_procedure": [
    "Identify each primary method or result figure.",
    "State that figure's main visible communication function and apply the functional deletion test to identify its necessary in-figure text.",
    "For each such figure, locate the smallest displayed item among the necessary text, ignoring normal superscript or subscript reduction.",
    "Compare the visible height of its base characters with the visible character height of the typical body-text reference."
  ],
  "threshold_guidance": "Treat 30% as an approximate visual threshold. A required text item clearly at or above the threshold satisfies the size requirement; one clearly below it violates the requirement. If the smallest required text is too close to the threshold for a reliable visual comparison, or the evidence cannot support the comparison, return UNCERTAIN rather than forcing PASS or FAIL.",
  "source_asset_scope": "Compare only the displayed character sizes in the attached poster. Do not judge source-image resolution, sharpness, compression, pixelation, scientific correctness, or relationships inside the figure. Do not assume that the figure can be cropped or enlarged separately.",
  "evidence_requirements": "For PASS, name a representative primary figure, its main function, the smallest necessary text class inspected, and the body-text reference. For FAIL, identify the primary figure, its location, the exact undersized text class, and why that text is functionally necessary. For NOT_APPLICABLE, state which required comparison object is absent."
}

PASS condition:
PASS only when every functionally necessary in-figure text item in every applicable primary method or result figure is displayed at least approximately 30% as tall as the poster's typical body text. Do not fail for excluded minor text, ordinary superscript or subscript reduction, or non-primary visuals.

FAIL condition:
FAIL when at least one functionally necessary in-figure text item in a primary method or result figure is clearly displayed below approximately 30% of the typical body-text character height. Identify the figure, location, undersized text class, and the communication function that makes it necessary.

Applicability:
Applicable when at least one primary method or result figure contains functionally necessary in-figure text and representative regular body text is visible. Return NOT_APPLICABLE when no primary figure exists, no primary figure contains functionally necessary in-figure text, or no representative body-text reference is visible. Do not use NOT_APPLICABLE merely because comparison is difficult.
\end{promptblock}

\begin{promptblock}[Prompt: IS01]
Question: Do the displayed sizes of the poster's information-bearing visuals match their visible information complexity?

Evaluation instructions:
{
  "definition": "A visual's displayed size matches its information complexity when the poster allocates enough area for the visual's main structure and key details to remain distinguishable. A visual containing more subpanels, text labels, nodes, arrows, data series, axes, legends, or annotations generally requires more display area than a visually simple image.",
  "relevant_visuals": "Evaluate information-bearing figures, charts, plots, flow diagrams, schematics, and image groups. Ignore logos, branding marks, decorative icons, background ornaments, and purely decorative images.",
  "scope": "Judge only poster-level displayed size relative to visible complexity. Treat the source visual as a trusted immutable asset: do not judge source resolution, sharpness, compression, pixelation, aspect ratio, scientific correctness, or correctness of relationships inside the visual. Do not infer complexity or importance from the source paper or outside knowledge. When importance matters, infer it only from the poster's visible heading, caption, placement, size, and emphasis.",
  "relationship_to_AE13": "AE13 separately applies the 30-percent body-text threshold to functionally necessary in-figure text in primary figures, using its functional deletion test. Do not fail IS01 solely because one isolated secondary label appears small or falls below that threshold. Fail IS01 for text-size evidence only when the visual's overall underscaling prevents necessary labels, legends, axes, or annotations from supporting the visual's main information function.",
  "pass_criteria": [
    "Visuals containing many subpanels, steps, nodes, arrows, curves, legends, or text labels receive enough display area for their visible complexity.",
    "The main components within each relevant visual remain visually separable rather than merging into a crowded or indistinguishable detail mass because of underscaling.",
    "Titles, labels, axes, legends, numbers, and key annotations necessary for the visual's main function are sufficiently discernible at the poster's presented scale; do not assume the reader can crop and enlarge the visual separately.",
    "Each panel in a multi-panel visual retains enough displayed area for its own main content to be distinguished, rather than only the multi-panel visual's overall silhouette being recognizable.",
    "A visually simple image may occupy a large area when it clearly serves as a focal element. Its size is a mismatch only when the allocation is visibly purposeless and clearly forces a more complex information-bearing visual into an inadequate area.",
    "Secondary visuals may be smaller than primary visuals, provided each still has enough displayed area to perform its own visible information function."
  ],
  "minor_variation_rule": "Minor details or nonessential labels may be difficult to read without causing FAIL when the visual's main structure, key evidence, and intended visible function remain discernible.",
  "fail_triggers": [
    "At least one important or functionally necessary visual contains substantial internal detail but is displayed so small that its main components merge or cannot be distinguished.",
    "A multi-panel visual is undersized to the point that one or more necessary subpanels communicate only an outline or color block rather than their main content.",
    "The overall scale of a visual prevents a necessary legend, axis, annotation, structural path, or principal data trend from supporting the visual's main information function.",
    "A clearly purposeless oversized simple visual consumes space in a way that visibly compresses a more complex, information-bearing visual below a functional size."
  ]
}

PASS condition:
PASS when every important information-bearing visual, and every smaller visual needed for the poster's visible message, has enough displayed area for its main structure and key details to remain distinguishable. Apply the minor-variation rule and do not require all fine print to be readable.

FAIL condition:
FAIL when at least one important or functionally necessary visual is clearly too small for its visible complexity and therefore cannot perform its main information function. Identify the visual, its location, and the specific structure, subpanel, legend, axis, annotation, or data trend made indistinguishable by underscaling.

Applicability:
Applicable when at least one information-bearing visual is displayed; otherwise NOT_APPLICABLE. Do not count logos or decorative imagery as information-bearing visuals.
\end{promptblock}

\paragraph{Visual guidance and emphasis.}
These five criteria examine hierarchy, reduced-scale recognizability, poster-level color harmony, the visual priority of branding and decoration, and emphasis on substantive conclusions.

\begin{promptblock}[Prompt: ENH01]
Question: Does the poster have a clear and reasonable visual hierarchy?

Evaluation instructions:
{
  "definition": "A reasonable visual hierarchy means that information of different apparent importance has a clear and stable priority relationship. Importance must be inferred only from the poster's visible headings, wording, layout, and declared emphasis. A reader should first recognize one main visual focus, then the major sections, key figures, and important findings, and finally the body text, captions, references, and supporting notes.",
  "identify_three_levels": [
    "Level 1: the main visual focus that attracts attention first, such as the main title, a concluding message, or a core figure.",
    "Level 2: major section headings, key figures other than any figure already serving as Level 1, and important results.",
    "Level 3: ordinary body text, captions, references, and supporting explanations."
  ],
  "pass_criteria": [
    "There is one clear and relatively unique Level-1 focus; several regions do not compete with approximately equal strength for first attention.",
    "Level-2 elements are visibly subordinate to Level 1 but visibly stronger than ordinary Level-3 text and supporting information.",
    "Level-3 text and captions remain readable but do not overpower the more important content.",
    "Visual salience broadly matches the information importance declared by the poster; secondary decoration, logos, references, or minor notes do not appear more prominent than core content.",
    "Elements serving the same hierarchy level are broadly consistent in font size, weight, color, contrast, and spacing.",
    "The viewer's attention can move naturally from the main focus to major content and then to supporting information. Judge this attention progression without requiring one fixed column-by-column reading path."
  ],
  "minor_variation_rule": "Minor local inconsistencies in size, color, or spacing are acceptable when the main focus, the separation among the three levels, and the overall attention progression remain clear.",
  "fail_triggers": [
    "No main visual focus can be identified.",
    "Multiple regions compete for first attention with approximately equal visual strength.",
    "Major headings, key figures, or important results are not visibly separated from ordinary body text.",
    "A secondary element such as decoration, branding, references, or a minor note visibly overpowers core content.",
    "Elements intended to occupy the same level repeatedly use conflicting visual treatments, making the hierarchy unstable.",
    "The viewer cannot form a natural attention progression from core information to major content and then supporting information."
  ]
}

PASS condition:
PASS only when the poster has an identifiable and relatively unique first focus, clear separation among the three hierarchy levels, no obvious salience inversion, broadly consistent same-level styling, and a natural attention progression. Minor local inconsistencies are allowed under the stated minor-variation rule.

FAIL condition:
FAIL when any listed fail trigger is clearly visible or when the three hierarchy levels cannot be distinguished reliably. Name the competing, inverted, or inconsistent elements in the evidence.

Applicability:
Always applicable. Base the judgement only on the poster's own visible content and normal-view appearance.
\end{promptblock}

\begin{promptblock}[Prompt: DIST01]
Question: Does the poster retain good long-distance recognizability when shown at 30% of its original width and height?

Evaluation instructions:
{
  "stimulus_protocol": "The attached image is a distance-view stimulus. The actual poster is the smaller rectangle centered on a neutral gray canvas. Its width and height are each 30% of the source poster dimensions, so it occupies 9% of the source poster area. The outer gray canvas is not part of the poster. Ignore that canvas completely: do not judge it as poster whitespace, background color, imbalance, or unused area.",
  "definition": "Good long-distance recognizability means that, even when body text, captions, and details inside charts are difficult or impossible to read, a viewer can still quickly identify the poster's title or topic-entry region, its main visual focus, its core content groups, and the location or visual form of its most important information.",
  "observation_restrictions": "Use only what is directly visible in the attached distance-view stimulus. Do not zoom, crop, enlarge, enhance, sharpen, use OCR, call tools, inspect files, or reconstruct text that is illegible at this scale. Do not use the source poster or outside knowledge. Recognizing a region as the title entry or a conclusion-like emphasis area does not require reading its exact wording.",
  "pass_criteria": [
    "The main title region remains visually prominent and functions as a clear entry point. It is visibly separated from author information, logos, and ordinary body text, even if the exact title wording cannot be read.",
    "One relatively clear main visual focus remains discoverable, such as a core method figure, key result region, or prominently framed concluding message. The viewer can quickly identify where to look first after the title.",
    "Major content regions remain distinguishable through columns, panel boundaries, section-heading treatments, color blocks, alignment, or intentional whitespace rather than merging into one continuous mass.",
    "A visible hierarchy remains among the title, major figures or result regions, section-level content, and supporting body text. Unreadable body text does not visually overpower the more important regions.",
    "The approximate reading entry, broad progression across the poster, and a conclusion-like or key-result region can be inferred from spatial placement and visual treatment without decoding illegible text.",
    "The reduced poster does not collapse into a broad continuous gray text wall, nor into many equally weighted tiny boxes, images, and color fragments with no dominant organization.",
    "At least one key-result or conclusion-like region remains discoverable through its position, scale, contrast, enclosure, or emphasis. The viewer need not read its exact values or sentences."
  ],
  "explicit_non_failures": [
    "Ordinary body text and captions are unreadable at this scale.",
    "Small labels, legends, axes, and annotations inside figures cannot be distinguished.",
    "Supporting explanations, references, affiliations, and other auxiliary information visually recede or disappear."
  ],
  "verdict_guidance": "Unreadable fine detail is expected and is not a reason to return UNCERTAIN. Return PASS or FAIL from the directly visible structural cues. Use UNCERTAIN only if the centered poster stimulus itself is missing, corrupted, substantially occluded, or otherwise unavailable for structural inspection.",
  "minor_variation_rule": "Minor ambiguity in one secondary region is acceptable when the title entry, main focus, core groups, and broad reading progression remain clearly recognizable.",
  "fail_triggers": [
    "The main title region is not visually identifiable or is mixed with author information, logos, or body text at approximately the same visual strength.",
    "No principal visual focus survives at the reduced scale, or several regions compete with approximately equal strength for first attention.",
    "Major content regions merge into an undifferentiated mass and their boundaries cannot be inferred.",
    "The poster becomes a broad, continuous gray text wall with no strong structural landmarks.",
    "The poster becomes a collection of many equally weighted tiny panels, images, or color fragments without a clear organizing hierarchy.",
    "No broad reading entry or progression can be inferred, and no key-result or conclusion-like region is visually discoverable."
  ]
}

PASS condition:
PASS only when the reduced poster still exposes all four essential cues: a clear title or topic-entry region, a main visual focus, distinguishable core content groups, and a broadly inferable reading progression with a discoverable key-result or conclusion-like region. Apply the explicit non-failures and minor-variation rule.

FAIL condition:
FAIL when one or more essential cues are clearly lost and the poster consequently becomes visually flat, merged, fragmented, or directionless at the supplied distance-view scale. Identify the missing cue and its approximate poster region without inferring illegible text.

Applicability:
Always applicable. Judge only the centered 30-percent poster rectangle and ignore the surrounding neutral gray simulation canvas.

Input transform:
{
  "type": "centered_distance_view",
  "linear_scale": 0.3,
  "canvas_size": "source_image_size",
  "canvas_background": "#7F7F7F",
  "placement": "center"
}
\end{promptblock}

\begin{promptblock}[Prompt: COL01]
Question: Does the academic poster have good overall color harmony?

Evaluation instructions:
{
  "definition": "Good color harmony means that the poster-level background, primary colors, secondary colors, accent colors, and the colors surrounding inserted visuals form a unified and stable whole through their hue, lightness, saturation, and relative area of use. The poster should not contain clear color conflicts, locally glaring treatments, or several unrelated color systems competing without hierarchy.",
  "scope": "Judge the poster-level color system: backgrounds, title and heading treatments, panel fills, borders, callouts, decorative elements, and the visible coexistence between inserted visuals and their surroundings. Treat each paper-derived figure, chart, heatmap, or photograph as a trusted immutable asset. Do not judge its internal palette, color encoding, internal color semantics, or scientific correctness. Judge only whether the poster-level colors placed around it create an obvious clash or prevent it from coexisting with the overall poster palette. Infer information importance only from visible headings, wording, placement, size, and emphasis.",
  "pass_criteria": [
    "The poster has an identifiable overall color direction: most poster-level headings, borders, backgrounds, and module treatments recur within a recognizable palette rather than each region using an unrelated color system.",
    "The number and visual strength of the main poster-level colors are controlled. A rich palette is acceptable when colors have a clear hierarchy; many highly saturated colors should not compete across large regions without a dominant system.",
    "Primary, secondary, and accent colors have observable roles. Primary colors establish the overall style, secondary colors organize or distinguish modules, and the strongest accents are used selectively for emphasis rather than repeatedly dominating large areas.",
    "Highly saturated or strongly contrasting colors are used with restraint. Vivid color may highlight important content, but several equally intense poster-level regions should not compete for attention.",
    "Adjacent poster-level color combinations appear visually compatible rather than clearly glaring, muddy, oppressive, or discordant.",
    "Backgrounds, poster text, icons, rules, and panel fills coexist with sufficient lightness and color separation for the foreground elements to remain visually distinguishable. For inserted figures, judge only their boundary and coexistence with surrounding poster colors, not their internal readability or palette.",
    "Color-rich figures, heatmaps, charts, or photographs can differ from the poster palette without causing failure when their surrounding panel, border, or background allows them to coexist without an obvious clash.",
    "Poster-level elements serving the same role, such as peer section headings, repeated module types, or the same type of callout, use broadly consistent color treatments. Do not inspect color consistency or data meaning inside trusted figures.",
    "The visual intensity of poster-level colors broadly matches the information importance declared by the poster. Secondary decoration, logos, borders, and backgrounds should not appear more chromatically dominant than visibly central methods, findings, or concluding information.",
    "Monochrome, analogous, complementary, and multicolor schemes are all acceptable. Do not fail a poster merely because it uses few colors, many colors, or strong contrast; judge whether the choices are organized, hierarchical, and unified."
  ],
  "minor_variation_rule": "A single figure or chart may use colors that differ from the surrounding palette, and a small accent may be unusually vivid, without causing FAIL when the overall color direction, hierarchy, and visual stability remain intact.",
  "fail_triggers": [
    "Prominent poster-level colors form a clearly jarring or discordant combination that disrupts the overall composition.",
    "Many highly saturated colors occupy substantial regions and compete without a recognizable primary, secondary, and accent hierarchy.",
    "Major regions appear to use unrelated color systems, making the poster look fragmented or assembled from incompatible templates.",
    "A local poster-level color block is abnormally intense or visually isolated enough to dominate unrelated core content.",
    "Background and foreground poster-level colors are too similar for important text, icons, rules, or shapes to remain distinguishable.",
    "Same-role poster-level elements repeatedly change color without an observable organizational purpose, weakening consistency."
  ]
}

PASS condition:
PASS when the poster-level colors form a recognizable, organized, and stable system with clear hierarchy, compatible adjacent combinations, restrained high-intensity accents, consistent same-role treatments, and no substantial clash with the surroundings of inserted visuals. Apply the minor-variation rule.

FAIL condition:
FAIL when at least one listed fail trigger is clearly visible and materially disrupts overall color unity, hierarchy, stability, or foreground distinction. Identify the specific colors, elements, and poster regions involved rather than giving a general taste-based opinion.

Applicability:
Always applicable, including monochrome posters. A monochrome or strongly contrasting palette is not automatically a failure.
\end{promptblock}

\begin{promptblock}[Prompt: EN05]
Question: Do logos, branding, decorative illustrations, and background treatments remain subordinate to the scientific content?

PASS condition:
No decorative or branding element attracts more attention than the title and primary scientific content or consumes disproportionate space.

FAIL condition:
A logo, banner, illustration, pattern, or background treatment visibly dominates primary scientific content.

Applicability:
Applicable when branding or decorative treatment beyond plain fills and rules is present; otherwise NOT_APPLICABLE.
\end{promptblock}

\begin{promptblock}[Prompt: EN07]
Question: Does the poster contain at least one visible, substantively complete conclusion-information unit that is clearly visually enhanced relative to nearby ordinary body text and peer non-concluding content?

Evaluation instructions:
{
  "conclusion_unit_definition": "A conclusion-information unit is a visible textual unit that can communicate a concluding point with enough context to be understood as such from the poster alone. It need not be a grammatically complete sentence, but it must identify enough of the finding, comparison, implication, or takeaway to avoid functioning as an isolated heading, number, or promotional fragment.",
  "qualifying_functions": [
    "State a principal finding: what the study visibly reports as having been found.",
    "State a key comparative result: what method, condition, or object performs differently and in what respect.",
    "State the implication, consequence, or meaning of a key result.",
    "State the final takeaway, practical value, recommendation, or message the viewer is meant to retain.",
    "Present a key quantitative result when its visible label or nearby wording identifies the object, metric, condition, or comparison well enough for the unit to carry a concluding meaning."
  ],
  "permitted_forms": "The unit may be a sentence, bullet, short paragraph, callout, result card, labeled metric card, or a caption that explicitly states a takeaway rather than merely describing a visual. It may appear under any heading or without a heading. Do not require the literal words Results, Conclusion, Takeaway, or equivalent fixed section names.",
  "non_qualifying_content": [
    "A Results, Conclusion, Takeaway, or similar heading without substantive conclusion content.",
    "Background, motivation, objective, problem-definition, or method-procedure text that does not state a finding or implication.",
    "A descriptive caption that only says what a figure contains or shows, without expressing a finding, comparison, implication, or takeaway.",
    "An isolated number, percentage, arrow, or metric name without enough visible context to identify what it concludes.",
    "A conclusion that could be obtained only by interpreting a figure, consulting the source paper, or using outside knowledge rather than reading a visible textual unit.",
    "An isolated promotional fragment such as better, significant improvement, or our method when the object, comparison, or conclusion remains unspecified."
  ],
  "candidate_selection": "Scan the whole poster for readable units that satisfy at least one qualifying function. Use only visible wording and local poster context. If several candidates exist, judge whether at least one substantively complete candidate receives sufficient visual enhancement; do not choose a fragment merely because it is already colorful or large.",
  "visual_enhancement_definition": "Visual enhancement must act on the conclusion-information unit itself or on its core concluding clause, and must make that content clearly more discoverable than nearby ordinary body text and peer non-concluding content. Emphasis applied only to a section heading does not transfer automatically to the conclusion text below it.",
  "strong_visual_cues": [
    "The complete conclusion unit is placed in a dedicated callout, result card, or clearly differentiated enclosing panel.",
    "A distinct background or enclosure clearly separates the complete conclusion unit from ordinary content.",
    "The complete conclusion unit, not merely its heading or one token, uses type that is clearly and substantially larger than nearby ordinary body text."
  ],
  "weak_visual_cues": [
    "The conclusion unit or its core clause is bold while nearby ordinary body text is not.",
    "The conclusion unit or its core clause uses a distinct accent color with sufficient contrast.",
    "The conclusion unit uses a modest but visible size increase over nearby body text.",
    "Purposeful whitespace or spatial isolation separates the conclusion unit from surrounding prose.",
    "An arrow, icon, divider, label treatment, or other attention marker points to or frames the conclusion unit.",
    "Prominent placement supports discovery when combined with another independent cue; placement alone is not sufficient."
  ],
  "sufficiency_rule": "The visual enhancement is sufficient only when the conclusion unit has at least one clear strong visual cue or at least two independent weak visual cues. Two instances of the same treatment do not count as two independent cues. A cue is not distinctive when the same treatment is applied broadly to ordinary content across the poster.",
  "whole_unit_rule": "The enhancement must make the complete conclusion unit or its core concluding clause discoverable. Emphasizing only the section heading, an unexplained number, or one promotional keyword is insufficient when the substantive conclusion remains styled like ordinary body text or buried in a long paragraph.",
  "scope_boundaries": "Judge only whether visible conclusion information is sufficiently and distinctly enhanced. Do not judge its scientific correctness, completeness with respect to the source paper, strength of evidence, overall poster hierarchy, or overall color harmony. ENH01 handles whole-poster hierarchy and COL01 handles color harmony. A poster may pass EN07 even if those other properties fail, and vice versa.",
  "pass_criteria": [
    "At least one readable candidate performs a qualifying conclusion function and contains enough visible context to communicate a substantive concluding point.",
    "The enhancement acts on the candidate itself or its core concluding clause, not only on a heading or isolated token.",
    "The candidate has at least one strong visual cue or at least two independent weak visual cues.",
    "The cues remain visibly distinctive relative to nearby body text and peer non-concluding content rather than being treatments used indiscriminately throughout the poster."
  ],
  "fail_triggers": [
    "Readable conclusion information exists but uses essentially the same visual treatment as nearby ordinary body text.",
    "Only a Results, Conclusion, Takeaway, or similar heading is emphasized while the substantive conclusion text is not.",
    "Only an isolated number or keyword is emphasized and the surrounding text needed to understand the conclusion is not enhanced.",
    "The candidate has only one weak cue and no strong cue, or its apparent cues are too slight to create reliable distinction.",
    "The same apparent emphasis treatment is used so widely for ordinary content that the conclusion unit is not distinctly discoverable.",
    "The conclusion remains buried within a long ordinary paragraph and does not form a discoverable unit despite minor local emphasis."
  ],
  "uncertainty_guidance": "Return UNCERTAIN when a likely conclusion unit is present but its wording is not readable enough to determine whether it performs a qualifying function, or when image evidence is insufficient to decide whether an apparent cue is visibly distinct. Do not use UNCERTAIN merely because the conclusion is not under a conventional heading.",
  "evidence_requirements": "For PASS, identify the unit's location, briefly paraphrase its concluding function, and name the strong cue or the two independent weak cues. For FAIL, identify the readable conclusion unit and the exact reason its enhancement is insufficient. For NOT_APPLICABLE, state that no directly visible and readable conclusion-information unit satisfying the definition is present."
}

PASS condition:
PASS only when at least one visible, readable, and substantively complete conclusion-information unit is enhanced by at least one clear strong visual cue or at least two independent weak visual cues, and the enhancement makes the unit itself or its core concluding clause distinctly discoverable relative to nearby body text and peer non-concluding content.

FAIL condition:
FAIL when readable conclusion information is present but no qualifying unit receives sufficient distinctive enhancement, including when only a heading, unexplained number, or isolated keyword is emphasized; when only one weak cue is present; when the treatment is used too broadly to distinguish the unit; or when the substantive conclusion remains buried in ordinary prose.

Applicability:
Applicable when at least one directly visible and readable textual unit satisfies the conclusion-unit definition. Return NOT_APPLICABLE when no such unit is present, regardless of section names. Return UNCERTAIN instead when a likely unit exists but its wording or visual treatment cannot be judged reliably from the supplied image. Do not use the source paper or infer a conclusion from a figure alone.
\end{promptblock}

\paragraph{Layout and spatial organization.}
The final four criteria address harmful overlap, spatial association, the match between visible content and occupied space, and whether large whitespace regions serve a clear layout function.

\begin{promptblock}[Prompt: AL01]
Question: Is the poster free from unintended poster-level overlaps that obscure or materially interfere with meaningful content?

Evaluation instructions:
{
  "definition": "An unintended poster-level overlap is a directly visible spatial intersection between two or more separately placed poster elements that obscures meaningful content, reduces its recognizability, or creates clear ambiguity about element ownership. Operationalize unintended through the harmful visible effect; do not speculate about the designer's intent.",
  "scope": "Inspect poster-level titles, author information, body-text blocks, figures, tables, captions, legends, formulas, callouts, panel borders, logos, badges, and decorative elements. Treat every inserted paper-derived figure, chart, table image, diagram, or image crop as one trusted atomic asset. Do not inspect overlaps among labels, arrows, nodes, curves, photographs, or other components baked inside that asset.",
  "scan_sequence": [
    "Inspect the main title, author, affiliation, logo, and header region.",
    "Inspect each major content panel or column separately, including its heading, body text, formulas, and callouts.",
    "Inspect every poster-level boundary between a figure and its caption, surrounding prose, neighboring figure, table, or panel.",
    "Inspect result cards, conclusion or takeaway callouts, badges, legends placed outside figures, and decorative overlays.",
    "Inspect the footer, references, QR codes, contact information, and lower edge regions."
  ],
  "fail_triggers": [
    "Two separately placed text blocks intersect so that characters cover one another or become materially harder to recognize.",
    "Poster-level text covers a meaningful figure or table region, including a data mark, label, legend, axis, image example, or necessary structural component.",
    "Two separately placed figures, tables, or content panels intersect and one obscures meaningful content in the other.",
    "A heading, caption, legend, formula, explanatory label, QR code, or contact item is visibly covered by another poster-level element.",
    "A logo, badge, decoration, border, rule, shadow, or color block lies above and obscures readable text or another meaningful visual element.",
    "A visible intersection makes a caption, heading, legend, or label appear to belong to the wrong content or leaves its ownership materially ambiguous."
  ],
  "explicit_non_failures": [
    "Two elements are close, cramped, or touching but do not spatially intersect.",
    "Text or a figure touches a panel boundary without being covered or losing meaningful content.",
    "A background fill, panel fill, highlight, label backing, or other clearly subordinate layer sits behind content while preserving recognizability.",
    "An intentional-looking border, underline, corner tag, badge backing, drop shadow, or decorative overlay does not obscure meaningful content.",
    "A connector, arrow, or rule crosses background space or another non-meaningful area without covering readable text or necessary visual content.",
    "The apparent overlap exists entirely inside a trusted paper-derived figure or other atomic visual asset."
  ],
  "verdict_guidance": "Mere proximity, insufficient spacing, or a small gap is not overlap. Return FAIL only when a directly visible poster-level intersection causes one of the listed harmful effects. For every FAIL, name both intersecting elements and their precise poster location."
}

PASS condition:
PASS when no separately placed poster-level element visibly intersects another in a way that obscures meaningful content, reduces recognizability, or creates clear ownership ambiguity. Apply all explicit non-failures.

FAIL condition:
FAIL when at least one listed fail trigger is directly visible. Identify both overlapping elements, the content affected, and the approximate location.

Applicability:
Always applicable.
\end{promptblock}

\begin{promptblock}[Prompt: AL05]
Question: Is every heading, caption, legend, bullet continuation, and explanatory label spatially grouped with the content it belongs to?

PASS condition:
Proximity and enclosure make the ownership of every auxiliary element unambiguous.

FAIL condition:
At least one auxiliary element is closer to, aligned with, or enclosed by the wrong content, or appears visibly orphaned.

Applicability:
Applicable when any of the named auxiliary elements are present; otherwise NOT_APPLICABLE.
\end{promptblock}

\begin{promptblock}[Prompt: DEN01]
Question: Does the poster have a reasonable visual information density?

Evaluation instructions:
{
  "definition": "Visual information density is the relationship between the visible amount and complexity of text, figures, charts, tables, formulas, modules, and decorative elements and the poster area that they occupy.",
  "scope": "Judge only visual information density and use of space. Do not judge whether the poster completely covers the source paper, whether the content is scientifically correct, or whether relationships inside a paper-derived figure are correct. For a complex figure, judge only whether its displayed size is appropriate for its visible amount of detail.",
  "pass_criteria": [
    "The amount of text, figures, charts, tables, and other modules broadly matches the available poster area; content has not been generally reduced to an impractically small display size merely to fit more material.",
    "Body-text paragraphs, text-and-figure pairs, and separate modules retain visibly distinguishable spacing; the poster does not contain a broad continuous region of pressed-together or densely stacked content.",
    "Complex figures, flow diagrams, tables, and image groups receive display areas commensurate with their visible numbers of labels, nodes, rows, columns, or subpanels; detail-rich assets are not presented at an obviously undersized scale.",
    "Differences in density across regions have an understandable layout purpose. A focal region may be denser and a supporting region sparser, but one region should not be severely crowded beside a large unused region with no visible design purpose.",
    "The poster is neither broadly filled with continuous small body text, dense graphics, and many fragmented modules nor visibly hollow because too little content leaves modules loose and substantial space purposelessly unused.",
    "In the normal whole-poster view, titles, main content, figures, and supporting information remain distinguishable; where local information is dense, element boundaries and grouping remain clear."
  ],
  "minor_variation_rule": "A few locally dense or sparse areas are acceptable when they do not materially impair element recognition, grouping, or the overall use of the poster area.",
  "fail_triggers": [
    "A broad continuous area is dominated by small body text or tightly packed modules.",
    "Several figures, charts, tables, or diagrams are obviously too small for their visible detail because of limited allocated space.",
    "Several modules are pressed together or lack visibly distinguishable separation.",
    "One region is severely crowded while a nearby region is conspicuously unused without an evident layout purpose.",
    "The poster is visibly overfilled or visibly underfilled as a whole.",
    "Local density makes element boundaries, grouping, or ownership difficult to identify."
  ]
}

PASS condition:
PASS only when the visible amount and complexity of content broadly match the occupied space, modules remain distinguishable, complex visuals receive adequate display area, regional density differences appear purposeful, and the poster is neither visibly overfilled nor underfilled. Apply the stated minor-variation rule.

FAIL condition:
FAIL when any listed fail trigger is clearly visible. Identify the affected region and the specific density problem in the evidence.

Applicability:
Always applicable. Judge the normal whole-poster view qualitatively and do not estimate an empty-area percentage.
\end{promptblock}

\begin{promptblock}[Prompt: WS01]
Question: Is the poster free from any large area of ineffective whitespace?

Evaluation instructions:
{
  "effective_whitespace_definition": "Effective whitespace is intentionally arranged empty space with an observable layout function. It helps readers understand relationships among content, establishes hierarchy, highlights important material, guides reading, relieves crowding, or creates a deliberate visual rhythm.",
  "ineffective_whitespace_definition": "Ineffective whitespace is a large empty region passively left by poor content placement. It has no observable grouping, emphasis, navigation, relief, or rhythm function and instead makes the poster look loose, disconnected, or visually imbalanced.",
  "effective_functions": [
    "Grouping: related elements are placed closer together while the whitespace separates them from a different module.",
    "Emphasis: extra space around an important title, key figure, or concluding message makes that element more prominent.",
    "Reading guidance: column gaps, section gaps, or paragraph gaps help establish a coherent reading path.",
    "Crowding relief: space between body text, figures, tables, numbers, or modules provides a useful visual pause.",
    "Rhythm: purposeful variation between denser and sparser regions creates an organized visual cadence rather than accidental emptiness."
  ],
  "scope": "Judge only poster-level whitespace between or within placed content modules. Do not inspect or penalize blank areas baked into a trusted paper-derived figure asset, and do not judge the source figure's aspect ratio. Where a figure is involved, judge only whether its poster-level displayed scale or placement leaves a large unusable surrounding gap. Do not infer that omitted paper content should have filled an empty region.",
  "decision_procedure": [
    "Identify any visibly large empty region relative to the poster and its neighboring modules; do not estimate a numerical area percentage.",
    "For each such region, look for an observable grouping, emphasis, reading-guidance, crowding-relief, or rhythm function.",
    "Treat a large region as ineffective only when no such function is observable and the region causes looseness, disconnection, unnecessary separation, wasted allocated space, or visual imbalance.",
    "Do not fail a poster merely because a blank region is large. A large intentional margin or open area is acceptable when its function is visually evident."
  ],
  "fail_examples": [
    "A figure or content module is displayed unnecessarily small while a large part of its allocated region remains unused.",
    "A module contains too little content and leaves a large internal hole with no visible grouping or emphasis purpose.",
    "Closely related neighboring content is separated by a disproportionately large empty gap without a reading-order function.",
    "One side or region is largely empty while another is crowded, producing a clear visual imbalance without an evident design purpose.",
    "A figure's poster-level displayed scale or placement creates an awkward large gap above, below, or beside it.",
    "A large empty region corresponds to no visible title, group boundary, focal element, reading path, or deliberate rhythm."
  ],
  "minor_variation_rule": "Small local holes, ordinary margins, and mildly uneven gaps are acceptable. FAIL requires at least one clearly large and clearly ineffective whitespace region."
}

PASS condition:
PASS when no clearly large ineffective whitespace region is visible. Large whitespace may pass when it observably supports grouping, emphasis, reading guidance, crowding relief, or visual rhythm.

FAIL condition:
FAIL when at least one clearly large empty region has no observable layout function and visibly causes looseness, disconnection, wasted allocated space, unnecessary separation, or visual imbalance. Identify the exact region and explain why it is ineffective.

Applicability:
Always applicable. If the poster contains no visibly large whitespace region, return PASS rather than NOT_APPLICABLE.
\end{promptblock}

\subsubsection{Output contract}
\label{app:prompt-output}

All visual-presentation calls use the same schema. It constrains the response to one criterion-level verdict with visible evidence, an approximate location, and confidence.

\begin{promptblock}[Output contract]
{
  "$schema": "https://json-schema.org/draft/2020-12/schema",
  "title": "P2P targeted atomic judgement response",
  "type": "object",
  "additionalProperties": false,
  "required": ["dimension", "results", "dimension_summary"],
  "properties": {
    "dimension": {
      "type": "string",
      "enum": [
        "aesthetic_element",
        "aesthetic_engagement",
        "aesthetic_layout"
      ]
    },
    "results": {
      "type": "array",
      "minItems": 1,
      "maxItems": 1,
      "items": {
        "type": "object",
        "additionalProperties": false,
        "required": ["id", "verdict", "evidence", "location", "confidence"],
        "properties": {
          "id": {"type": "string", "minLength": 3},
          "verdict": {
            "type": "string",
            "enum": ["PASS", "FAIL", "UNCERTAIN", "NOT_APPLICABLE"]
          },
          "evidence": {"type": "string", "minLength": 1},
          "location": {"type": "string", "minLength": 1},
          "confidence": {"type": "number", "minimum": 0.0, "maximum": 1.0}
        }
      }
    },
    "dimension_summary": {"type": "string"}
  }
}
\end{promptblock}

\subsection{Reference-Unit Extraction Prompt}
\label{app:prompt-extraction}

This prompt converts each author-designed official poster into exactly seven semantically distinct reference units. It follows the poster's own emphasis and defines the minimum evidence later required for coverage.

\begin{promptblock}[Prompt: reference-unit extraction]
You are constructing reference information units for an academic-poster information-coverage benchmark.

You receive exactly one attached image: an author-made OFFICIAL academic poster. Use only the visible poster. Do not inspect the paper, repository files, generated posters, or outside sources. Do not call tools.

Extract exactly SEVEN core information units that collectively represent the main information the poster's authors chose to communicate.

Selection principles:

1. Follow this poster's own emphasis. Different posters may emphasize a method, theoretical statements, equations, datasets, empirical findings, applications, or limitations. Do not force a fixed section taxonomy.
2. Prefer central messages over minor implementation details, isolated hyperparameters, acknowledgements, citations, QR codes, or decorative text.
3. Each unit must be independently judgeable in another poster. Avoid vague units such as “the method works well.”
4. Keep units semantically distinct. Do not split one claim into multiple near-duplicates merely to reach seven.
5. Conversely, do not combine unrelated claims into one unit. When some compression is necessary, combine only facts that form one coherent message.
6. Use the visible poster as the complete evidence boundary. Do not add facts remembered from the paper or inferred from scientific plausibility.
7. Balance motivation/problem, method/theory, and results only when the official poster itself emphasizes all of them.

Required fields for each unit:

- `unit_id`: use `<poster_id>-U1` through `<poster_id>-U7` in the order a reader would naturally understand the poster.
- `information_type`: a concise role such as `background`, `problem`, `method`, `theory`, `dataset`, `result`, `limitation`, or another poster-appropriate type.
- `topic`: a short label for the unit.
- `local_context`: two or three sentences defining terms and situating the claim so a later judge will not misinterpret it.
- `statement`: one concise, self-contained formulation of the core information.
- `coverage_requirement`: the minimum semantic content a generated poster must visibly communicate to count this unit as covered. Make this tolerant of paraphrases and compression. Do not require every number, example, dataset name, or qualifier unless it is central to the claim.
- `official_evidence`: a short description of the exact visible section, phrase, equation, diagram, plot, or table supporting the unit. Do not merely write “official poster.”

Quality guardrails:

- A statement may include an exact number only when the number is prominently emphasized and materially changes the message.
- A coverage requirement must not be stricter or broader than the statement.
- If a result table is dense, extract its clear comparative conclusion rather than several individual cells.
- Formulas and diagrams count as information, but describe the semantic message they convey.
- Correct obvious OCR spelling internally, but do not invent unreadable content.

Return only JSON conforming to the supplied schema.
\end{promptblock}

\subsection{Paper-Grounding Audit Prompt}
\label{app:prompt-audit}

This prompt checks every poster-derived reference unit against the corresponding source paper. It identifies unsupported, contradicted, or overly broad statements and requests a grounded revision when appropriate.

\begin{promptblock}[Prompt: paper-grounding audit]
You are auditing whether information units extracted from an author-made academic poster are grounded in the supplied original paper PDF.

You have one extracted paper text file. It contains explicit markers of the form `===== PDF PAGE N =====`. You also receive exactly seven poster-derived information units. Use only this paper text; do not inspect the poster, generated posters, repository materials, or external sources.

Your job is not to ask whether the poster said the claim. Your job is to determine whether the claim, as currently written, is supported by this exact paper version.

First verify paper identity from the title and first-page content:

- `MATCH`: the PDF is the paper expected for this benchmark ID.
- `TITLE_ALIAS_OR_RENAMED`: the scientific work matches, but the benchmark/file title is an older, shortened, or alternate title.
- `MISMATCH`: the PDF is a different scientific work.
- `UNCERTAIN`: identity cannot be established from the extracted text.

For each unit, assign exactly one verdict:

- `SUPPORTED`: every central part of the statement is explicitly supported by the paper or is a direct, unavoidable consequence of text/equations/results in the paper. Minor paraphrase is allowed.
- `SUPPORTED_WITH_NARROWING`: the central idea is present, but the current statement is broader, stronger, more specific, or more compound than the paper supports. Use this whenever one material clause of a compound statement is unsupported.
- `NOT_FOUND`: the paper does not provide adequate evidence for the statement. Do not treat a cited related-work claim as evidence that this paper established the claim.
- `CONTRADICTED`: the paper provides evidence that materially disagrees with the statement, including a different number, direction, scope, condition, dataset, or conclusion.
- `UNREADABLE`: text extraction is too incomplete to make a reliable judgment. Use sparingly.

Choose one issue type:

- `NONE`
- `OVERSTATEMENT`
- `EXTRA_DETAIL`
- `NUMERIC_MISMATCH`
- `CLAIM_MISMATCH`
- `POSTER_ONLY`
- `VERSION_MISMATCH`
- `COMPOUND_PARTIAL_SUPPORT`
- `UNVERIFIABLE`

Audit rules:

1. Judge semantic content, not word overlap.
2. Check exact numbers, dataset names, comparison directions, guarantees, assumptions, and qualifiers whenever they materially affect the claim.
3. For theoretical claims, distinguish assumptions, theorems, interpretations, and empirical observations.
4. For empirical claims, verify the named comparison and direction. Do not upgrade “competitive” to “best” or “some settings” to a universal result.
5. Background/motivation can be supported by the paper's own framing. A claim attributed to this work requires this work's own result, not only a bibliography citation.
6. Provide one to three short, exact evidence excerpts with PDF page number and section/figure/table context. Each quote must be short enough to locate easily.
7. If the verdict is `NOT_FOUND`, evidence should normally be empty. If `CONTRADICTED`, quote the conflicting paper evidence.
8. `recommended_statement` must be empty for `SUPPORTED`. For `SUPPORTED_WITH_NARROWING` or `CONTRADICTED`, provide a concise replacement fully grounded in the paper. For `NOT_FOUND`, provide a grounded salvage only if a clearly related claim exists; otherwise leave it empty.
9. Write `reasoning`, `overall_assessment`, and `recommended_statement` in Chinese. Preserve paper terminology and evidence quotes in their original language.
10. Return the seven units in the supplied order and output only JSON conforming to the schema.
\end{promptblock}

\subsection{Independent Adjudication Prompt}
\label{app:prompt-adjudication}

This second-pass prompt independently reviews only the units flagged by the first audit. It can confirm, revise, or dismiss the earlier decision before the reference set is frozen.

\begin{promptblock}[Prompt: independent adjudication]
You are the second-pass adjudicator for a paper-grounding audit of poster-derived information units.

You receive only units that a first-pass auditor flagged as not fully supported. Independently read the specified extracted paper text file and challenge the first-pass decision. Use only that paper. Do not inspect the official poster, generated posters, other repository materials, or external sources.

For each flagged unit, choose:

- `CONFIRM`: the first-pass verdict and material reasoning are correct.
- `REVISE`: the unit has a real grounding issue, but the verdict, issue type, evidence, or recommended wording should change.
- `DISMISS`: the original poster-derived unit is fully supported by this exact paper version; remove the first-pass flag.

Final verdict definitions:

- `SUPPORTED`: every central clause is supported. Use this with `DISMISS`.
- `SUPPORTED_WITH_NARROWING`: a meaningful core is supported, but a material clause is too broad, too strong, overly specific, or absent. In particular, do not use `NOT_FOUND` for a compound unit when a central core can be salvaged.
- `NOT_FOUND`: the paper provides no adequate support for the material claim; related-work mentions or a merely adjacent idea are insufficient.
- `CONTRADICTED`: the paper gives a materially different number, direction, condition, scope, dataset, entity name, or conclusion.

Adjudication rules:

1. Read the relevant paper pages yourself. Do not trust the first-pass rationale or evidence without checking.
2. Judge the original unit statement, not the first-pass recommended replacement.
3. Treat exact numbers and universal/superlative claims carefully, but do not call a harmless stylistic paraphrase a contradiction.
4. When a statement underclaims a stronger paper result without becoming false, normally use `SUPPORTED` or `SUPPORTED_WITH_NARROWING`, not `CONTRADICTED`.
5. When a dataset/method title changed across versions but the entity and all substantive properties are otherwise the same, use `CONTRADICTED` with `VERSION_MISMATCH` only if the old name would materially misidentify the current paper's entity.
6. Provide one to three short evidence excerpts with PDF page and section/table/figure location. For `NOT_FOUND`, evidence may be empty; explain what searches/sections fail to support it.
7. `recommended_statement` must be empty for `SUPPORTED`. Otherwise provide the shortest accurate replacement when salvageable; leave it empty only when the unit should be removed.
8. Write reasoning and recommended statements in Chinese. Preserve source evidence in its original language.
9. Return exactly the flagged unit IDs supplied, in the supplied order.
\end{promptblock}

\subsection{Candidate-Poster Coverage Prompt}
\label{app:prompt-coverage}

This prompt performs the final image-grounded coverage judgment. The evaluator receives one generated poster and its seven frozen reference units, then reports whether each unit's minimum semantic requirement is visibly covered.

\begin{promptblock}[Prompt: candidate-poster coverage]
You are evaluating information coverage in one generated academic poster.

You receive:
1. One attached image: the GENERATED poster to evaluate.
2. Seven reference information units extracted from the corresponding author-made OFFICIAL poster.

Use only the attached generated-poster image and the supplied reference units. This is an image-grounded coverage judgment, not a paper-factuality review.

For every reference unit, assign exactly one verdict:

- `COVERED`: The generated poster visibly communicates the unit's core meaning. Accept faithful paraphrases, synonyms, compressed wording, formulas, diagrams, plot labels, captions, and legible result tables. The same information may appear in a different section. Do not demand exact wording, every qualifier, or every example when the stated `coverage_requirement` is met.
- `NOT_COVERED`: The core meaning cannot be found, is too vague to identify, is only suggested by an unrelated keyword, or is replaced by a contradictory claim. Partial content counts as covered only when it satisfies the supplied minimum `coverage_requirement`.

Rules:
- Judge information only. Ignore aesthetics, layout quality, typography, and visual appeal, except that unreadable content cannot establish coverage.
- Do not penalize extra information.
- Do not infer a claim merely because it would be scientifically plausible.
- In `evidence`, quote a short visible phrase or name a specific visible visual/table/section. If absent, write `No supporting content found.`
- Keep `rationale` concise and explain semantic match or the missing core relation.
- Preserve the input unit order and unit IDs.
- Set `missing_count` equal to the number of `NOT_COVERED` verdicts.

Return only JSON conforming to the supplied schema.
\end{promptblock}

\end{document}